\documentclass{article}
\usepackage[utf8]{inputenc}

\PassOptionsToPackage{numbers, compress}{natbib}

\usepackage[preprint]{neurips_2026}

\usepackage[utf8]{inputenc} 
\usepackage[T1]{fontenc}    
\usepackage{alphalph}       
\usepackage{hyperref}       
\usepackage{url}            
\usepackage{booktabs}       
\usepackage{amsfonts}       
\usepackage[table]{xcolor}  
\usepackage{nicefrac}       
\usepackage{microtype}      
\usepackage{graphicx}
\usepackage{float}
\usepackage{multirow}
\usepackage{pifont}
\usepackage{amssymb}
\usepackage{chngcntr}
\counterwithout{table}{section}
\usepackage[most]{tcolorbox}
\usepackage{longtable}

\usepackage{makecell}
\usepackage{threeparttable}
\usepackage{array}
\newcolumntype{P}[1]{>{\centering\arraybackslash}p{#1}}

\newcommand{\eg}{\textit{e.g.}, }

\newcommand{\wrt}{\textit{w.r.t. }}

\newcommand{\cmark}{\ding{51}}

\title{DeepTumorVQA: A Hierarchical 3D CT Benchmark for Stage-Wise Evaluation of Medical VLMs and Tool-Augmented Agents}

\author{
\textbf{Yixiong Chen}$^{1}$,
\textbf{Wenjie Xiao}$^{1}$,
\textbf{Pedro R. A. S. Bassi}$^{1,2,4}$,
\textbf{Boyan Wang}$^{5}$,
\textbf{Liang He}$^{6}$, \\
\textbf{Xinze Zhou}$^{1}$,
\textbf{Sezgin Er}$^{3}$,
\textbf{Ibrahim Ethem Hamamci}$^{3}$,
\textbf{Zongwei Zhou}$^{1}$,
\textbf{Alan Yuille}$^{1}$ \\
$^{1}$Johns Hopkins University \quad
$^{2}$University of Bologna \quad
$^{3}$Istanbul Medipol University \quad \\
$^{4}$Center for Biomolecular Nanotechnologies, Istituto Italiano di Tecnologia \quad \\
$^{5}$The First Affiliated Hospital, Sun Yat-Sen University \quad
$^{6}$Tongji University\\
\texttt{ayuille1@jhu.edu}
}

\begin{document}

\maketitle

\begin{abstract}
Medical vision-language models (VLMs) and AI agents have made significant progress in learning how to analyze and reason about clinical images. However, existing medical visual question answering (VQA) benchmarks collapse their capabilities into a single accuracy score, obscuring \emph{where} and \emph{why} models fail.
We propose DeepTumorVQA, a hierarchical benchmark that follows the multi-stage evidence chain in tumor diagnosis and decomposes 3D CT reasoning into four stages: recognition, measurement, visual reasoning, and medical reasoning. Higher-level questions remain independently scorable, while their ground-truth evidence chains are defined over lower-level primitives. The benchmark is instantiated as 476K questions across 42 clinical subtypes on 9{,}262 3D CT volumes. In addition to a direct reasoning mode for VLMs, DeepTumorVQA provides tool-interaction environments for agent evaluation, where a model can call external tools, including segmentation models, measurement programs, and medical knowledge modules. Models can answer the question after observing tool outputs.
Evaluating over 30+ model configurations, we find that reliable quantitative measurement is the primary bottleneck, making tasks at later stages (visual reasoning, and medical reasoning) even harder for VLMs, while tool augmentation substantially mitigates this issue. When tools are available for VLMs, how to leverage medical knowledge and tools to reason about medical images becomes a new challenge. We further show that ground-truth step-by-step tool-use traces from DeepTumorVQA can supervise agents and reduce tool-use and reasoning failures.
This stage-wise progression from recognition to measurement to visual and medical reasoning provides a concrete roadmap for future medical VLM and AI agent studies. All data and code are released at \href{https://github.com/Schuture/DeepTumorVQA}{github.com/Schuture/DeepTumorVQA}. 

\end{abstract}

\section{Introduction}
\label{sect:intro}

Medical vision-language models (VLMs) have progressed from 2D image understanding~\citep{li2023llava,zhang2023pmc} to 3D volumetric reasoning~\citep{wu2023towards,bai2024m3d,blankemeier2024merlin,hamamci2024developing}, and more recently to tool-augmented AI agents that invoke segmentation, measurement, and knowledge tools for diagnostic reasoning~\citep{chen2025meissa,eltawil2025medrax,bassi2025radgpt}. This evolution has spurred evaluation benchmarks for both paradigms~\citep{hu2024omnimedvqa,schmidgall2025medagentbench,tang2025_3drad}.

However, existing benchmarks remain fragmented (Table~\ref{tab:benchmark_comparison}): they cover different parts of the evaluation problem rather than a unified diagnostic protocol. More importantly, most benchmarks still reduce medical image understanding to a single end-task score. This makes it difficult to determine whether a model fails because it cannot recognize a lesion, cannot measure quantitative evidence, cannot combine multiple visual findings, or cannot apply clinical knowledge. It also hides shortcut behavior: a model may answer a high-level diagnostic question correctly from dataset bias or language priors without grounding its answer in the required low-level visual evidence.

\begin{table}[t]
\centering
\scriptsize
\setlength{\tabcolsep}{2.5pt}
\caption{Comparison with concurrent medical AI benchmarks. Comp.\ = compositionally decomposed questions. DeepTumorVQA comprises 476K questions (428K train / 48K test). The benchmark evaluation uses a balanced subset of 10,000 questions from the test set, stratified across 42 subtypes.}
\label{tab:benchmark_comparison}
\begin{tabular}{lrr|ccccc}
\toprule
\textbf{Benchmark} & \textbf{Train} & \textbf{Bench} & \textbf{2D-view} & \textbf{3D-volume} & \textbf{Explicit reasoning} & \textbf{Agent} & \textbf{Compositional} \\
\midrule
\multicolumn{8}{l}{\textit{VQA Benchmarks}} \\
VQA-RAD~\citep{lau2018dataset} & 3.1K & 451 & \cmark & & & & \\
PMC-VQA~\citep{zhang2023pmc} & 150K & 2K & \cmark & & & & \\
OmniMedVQA~\citep{hu2024omnimedvqa} & -- & 128K & \cmark & & \cmark & & \\
MedSG-Bench~\citep{yu2025medsgbench} & 188K & 9.6K & \cmark & & \cmark & & \\
3D-RAD~\citep{tang2025_3drad} & 136K & 3.7K & & \cmark & \cmark & & \\
\midrule
\multicolumn{8}{l}{\textit{Agent Benchmarks}} \\
MedAgentBench~\citep{schmidgall2025medagentbench} & -- & 300 & & & \cmark & \cmark & \\
AgentClinic~\citep{schmidgall2024agentclinic} & -- & 200 & \cmark & & & \cmark & \\
MedChain~\citep{gao2024medchain} & -- & 12K & & & \cmark & \cmark & \\
ChestAgentBench~\citep{eltawil2025medrax} & -- & 2.5K & \cmark & & & \cmark & \\
\midrule
\textbf{DeepTumorVQA (Ours)} & \textbf{428K} & \textbf{10K} & \cmark & \cmark & \cmark & \cmark & \cmark \\
\bottomrule
\end{tabular}
\end{table}

\begin{figure}[t]
    \centering
    \includegraphics[width=1.0\linewidth]{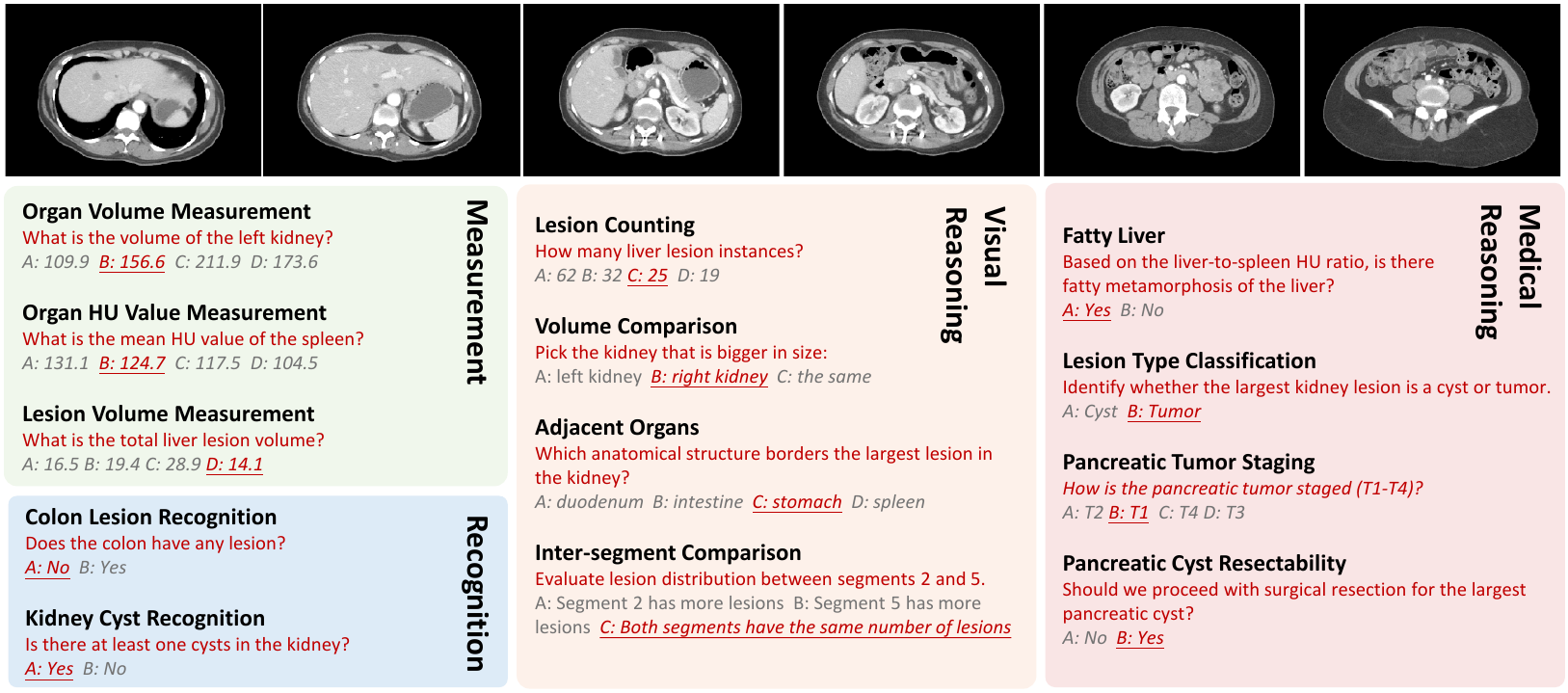}
    \caption{Overview of the DeepTumorVQA benchmark. The benchmark covers four diagnostic task types with 42 clinical subtypes, supports both 2D/3D inputs and both direct inference/agent reasoning, and evaluates along multiple dimensions including accuracy, latency, and agent trajectory quality.}
    \label{fig:task_dem}
\end{figure}

We present \textbf{DeepTumorVQA} (Figure~\ref{fig:task_dem}), a 3D CT diagnostic benchmark with 476K questions across \textbf{42 clinical subtypes}. Its core design is a four-level hierarchy organized by clinical evidence granularity: Recognition tests atomic visual entities such as organs and lesions; Measurement tests quantitative properties such as organ volume and HU values; Visual Reasoning composes these primitives across anatomy and spatial relations; and Medical Reasoning combines the composed evidence with clinical knowledge and guideline-like rules. By evaluating both final answers and their prerequisite evidence, DeepTumorVQA can distinguish grounded reasoning from high-level answers that may arise from shortcuts or language priors. The benchmark supports 2D slices, 3D volumes, and video inputs. It also provides three controlled agentic evaluation modes, where models call external evidence-producing tools such as segmentation models, measurement programs, and medical knowledge modules. These modes vary evidence availability and quality, allowing us to separate failures caused by visual perception, tool reliability, and final reasoning (Section~\ref{sect:agent_env}).

Benchmarking over 30 model configurations, DeepTumorVQA reveals a stage-wise error landscape that is hidden by aggregate accuracy. Across zero-shot VLM backbones, errors increase sharply from atomic recognition to quantitative measurement, indicating that reliable estimation of organ and lesion properties is a major bottleneck in our 3D CT diagnostic tasks. This bottleneck then propagates to visual and medical reasoning, where many failures can be traced to missing or incorrect quantitative evidence rather than to the final reasoning step alone. Tool augmentation improves access to this evidence, but it also exposes a different failure mode: models must select the right tool, sequence tool calls, interpret outputs, and combine them with medical knowledge. Thus, DeepTumorVQA does not only rank models; it explains why models with similar overall accuracy fail differently. We further show that step-by-step tool-use traces from DeepTumorVQA can supervise agents and reduce both tool-use and reasoning errors.

Our contributions are three-fold. First, we introduce DeepTumorVQA, a 3D CT benchmark with 476K questions, 42 clinical subtypes, and 9{,}262 CT volumes, organized as a four-level clinical evidence hierarchy from recognition to measurement, visual reasoning, and medical reasoning. Second, we provide a unified evaluation protocol for direct VLM inference and tool-augmented agents, including 2D, 3D, and video input formats, three controlled agent environments, and open-source evaluation infrastructure. Third, we benchmark over 30 model configurations and identify reliable quantitative measurement as a major bottleneck in 3D CT VQA, while showing how tool availability and trace supervision change the failure modes of medical VLMs and agents.

\section{Related Work}
\label{sect:related_work}

\paragraph{Medical Vision-Language Models.}
Medical VLMs have evolved along three tracks.
\emph{3D specialists} primarily differ in how they encode volumetric input: RadFM~\citep{wu2023towards} tokenizes 3D patches through a perceiver-based architecture trained on 16M image-text pairs; M3D-LaMed~\citep{bai2024m3d} applies a 3D vision transformer over 120K CT-text pairs; Merlin~\citep{blankemeier2024merlin} adopts a ResNet-based 3D encoder; CT-CHAT~\citep{hamamci2024developing} extends this line to full 3D CT report generation.
\emph{2D VLMs} are adapted via finetuning (LLaVA-Med~\citep{li2023llava}, HuatuoGPT-Vision~\citep{chen2024huatuogpt}, MedGemma~\citep{yang2024advancing}) or applied zero-shot (Qwen3-VL~\citep{qwen3vl2025}, Qwen3.5, InternVL3~\citep{internvl2024}, LLaVA-OneVision~\citep{liu2024llavanext}). Frontier model APIs (Gemini-3, GPT-5.4) offer strong reasoning but lack systematic 3D CT evaluation.
Despite rapid progress, no unified evaluation compares all three categories on identical 3D diagnostic tasks; DeepTumorVQA fills this gap.

\paragraph{Medical AI Agents.}
Medical AI agents span multi-agent and tool-augmented paradigms. MDAgents~\citep{kim2024mdagents} adaptively assembles specialist LLMs for clinical decision-making; MedAgentSim~\citep{almansoori2025medagentsim} simulates patient-doctor interactions for agent training. On the tool-augmented side, Meissa~\citep{chen2025meissa} distills agent trajectories from frontier models; MedRAX~\citep{eltawil2025medrax} coordinates eight tools for chest X-ray interpretation; RadGPT~\citep{bassi2025radgpt} uses tool-assisted annotation for 3D datasets; MMedAgent~\citep{li2024mmedagent} learns multi-modal tool selection; \citet{jiang2025incentivizing} incentivize tool-augmented thinking for medical image analysis. Most recently, CT-Agent~\citep{mao2025ct} introduces anatomically independent tools with dual-path token compression for 3D CT question answering. Despite this progress, few existing work provides systematic 3D agent benchmarking with tool ablations that isolate individual capability stages.

\paragraph{Medical VQA Benchmarks.}
Existing medical VQA benchmarks each leave a critical dimension untested (Table~\ref{tab:benchmark_comparison}). VQA-RAD~\citep{lau2018dataset} and PMC-VQA~\citep{zhang2023pmc} test 2D recognition; OmniMedVQA~\citep{hu2024omnimedvqa} scales to broad coverage but lacks agent evaluation; 3D-RAD~\citep{tang2025_3drad} adds 3D but not tool-augmented reasoning; CT-Bench~\citep{zhu2026ct} benchmarks multimodal lesion understanding but focuses on lesion-level tasks; 3DLAND~\citep{advand20263dland} provides large-scale 3D lesion annotations but targets detection rather than diagnostic VQA; MedSG-Bench~\citep{yu2025medsgbench} evaluates sequential grounding but not quantitative evidence. On the agent side, MedAgentBench~\citep{schmidgall2025medagentbench} tests EHR agents in text-only settings; AgentClinic~\citep{schmidgall2024agentclinic} simulates dialogues without volumetric data; MedChain~\citep{gao2024medchain} benchmarks clinical workflows but cannot isolate whether quantitative evidence is the bottleneck. Most prior medical VQA benchmarks score end-task accuracy. DeepTumorVQA instead provides intervention-based attribution, diagnosing \emph{why} models fail and at which stage. By providing ground-truth tool traces, controlled ablations, and multiple agent modes, it enables pinpointing bottlenecks rather than just ranking models.

\section{DeepTumorVQA Benchmark}
\label{sect:benchmark}

\subsection{Compositional Question Design}
\label{sect:question_design}

The central design principle is compositional decomposition: complex medical reasoning is broken into testable atomic capabilities, inspired by hierarchical benchmarks like CLEVR~\citep{johnson2017clevr}, CLEVR-Ref+~\citep{liu2019clevr}, and Super-CLEVR~\citep{li2023super} but grounded in clinical workflows (we discuss the compositional benchmarks in Appendix~\ref{sect:appendix_clevr}). Each Medical Reasoning question chains lower-level Recognition and Measurement primitives, with every subtype backed by clinical literature (Appendix~\ref{sect:task_definitions}).

\begin{figure}[t]
    \centering
    \includegraphics[width=1.0\linewidth]{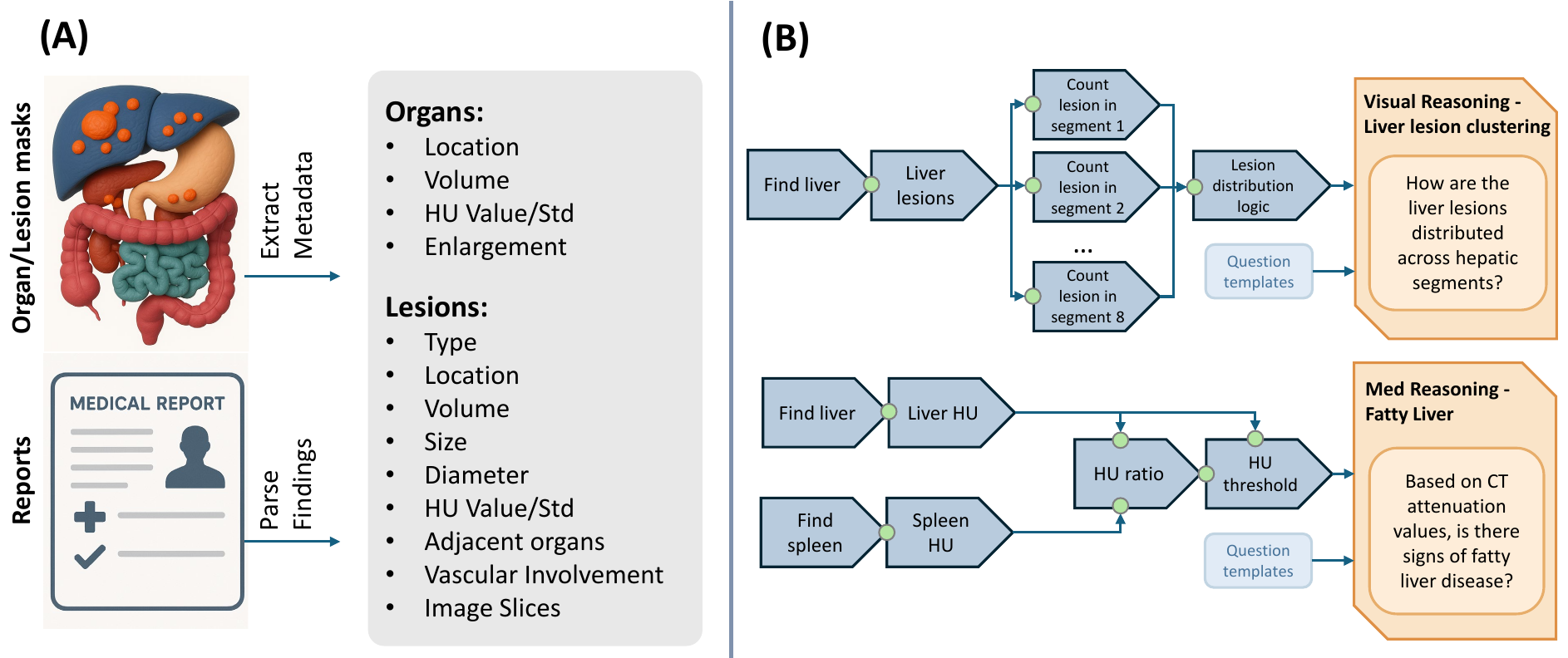}
    \caption{Compositional question design. \textbf{(A)} Structured metadata extracted from segmentation masks and radiology reports. \textbf{(B)} Modular logic programs generate questions across four types and 42 subtypes. Each medical reasoning subtype is composed from recognition and measurement primitives, with clinical literature grounding.}
    \label{fig:dataset_construction}
\end{figure}

We define four task types of increasing complexity (Figure~\ref{fig:dataset_construction}).
\textbf{Recognition} (9 subtypes): binary detection of organ- or lesion-level findings, including liver, kidney, pancreatic, and colon lesions, cysts, splenomegaly, and pancreatic tumor categories such as PDAC and PNET.
\textbf{Measurement} (5 subtypes): organ volume, organ Hounsfield unit (HU), lesion volume, HU ratio, and tumor burden percentage from 3D volumes.
\textbf{Visual Reasoning} (16 subtypes): compositional tasks combining recognition and measurement with spatial comparison (counting, localization, bilateral asymmetry, multi-organ burden comparison).
\textbf{Medical Reasoning} (12 subtypes): compositional tasks requiring clinical knowledge integration, including fatty liver (L/S HU ratio~\citep{zeb2012computed}), hepatic steatosis grading~\citep{kodama2007comparison}, pancreatic steatosis~\citep{guneyli2022computed}, splenomegaly grading~\citep{bezerra2005spleen}, PDAC vs.\ PNET~\citep{nccn2024pancreatic}, portal hypertension~\citep{harbin1980portal}, renal mass characterization~\citep{silverman2019bosniak}, kidney lesion type~\citep{agochukwu2017renal}, pseudocyst determination~\citep{allen2011pseudocyst}, etc. Each subtype has a deterministic ``program'' chaining recognition, measurement, and knowledge lookup (metadata extraction details in Appendix~\ref{sect:metadata_extraction}).

Recognition includes both \emph{morphological} subtypes (lesion presence from segmentation) and one \emph{threshold-based} subtype (splenomegaly: volume $>$314.5\,cm$^3$), reflecting how clinical diagnosis routinely blurs categorical and quantitative assessment (Appendix~\ref{sect:recog_meas_boundary}).

This compositional structure provides interpretable diagnostics of model capabilities, defines ground-truth tool traces for agent evaluation (Section~\ref{sect:agent_env}), and supports controlled compositional splits over clinically meaningful factors (Section~\ref{sect:eval}).

\paragraph{Question-answer pair generation.}
For every subtype, we generate the question from 10 pre-defined templates, and the ground-truth answer from its program. The correct answer is among several distractors in the multiple-choice (MC) question. The distractors are sampled uniformly from [0.7$\times$GT, 1.3$\times$GT] with minimum pairwise separation of 3--10\% of GT to prevent trivially close options for continuous-valued subtypes (volume, HU). For ordinal subtypes (grading, staging), all clinically defined grades serve as options. For binary subtypes (recognition), options are Yes/No. All models face identical option sets.

\subsection{Data and Annotations}
\label{sect:data}

The benchmark is built on 9,262 abdominal CT volumes from 17 public datasets (Table~\ref{tab:data_overview} in Appendix \ref{sect:data_sources}), encompassing 88 medical centers. Annotations derive from 23 radiologists (detailed in Appendix \ref{sect:annotators}). They annotated 11{,}319 lesions in 3D with multi-round consensus. QA generation is fully deterministic given validated metadata fields (organ volumes, HU values, lesion counts) computed from consensus segmentation masks covering 43 anatomical structures per volume; we verify annotation quality on external datasets in Appendix~\ref{sect:seg_quality}.

\begin{figure}[t]
    \centering
    \includegraphics[
        width=1.0\linewidth,
        trim={0.0cm 0.2cm 0.0cm 0.2cm},
        clip
    ]{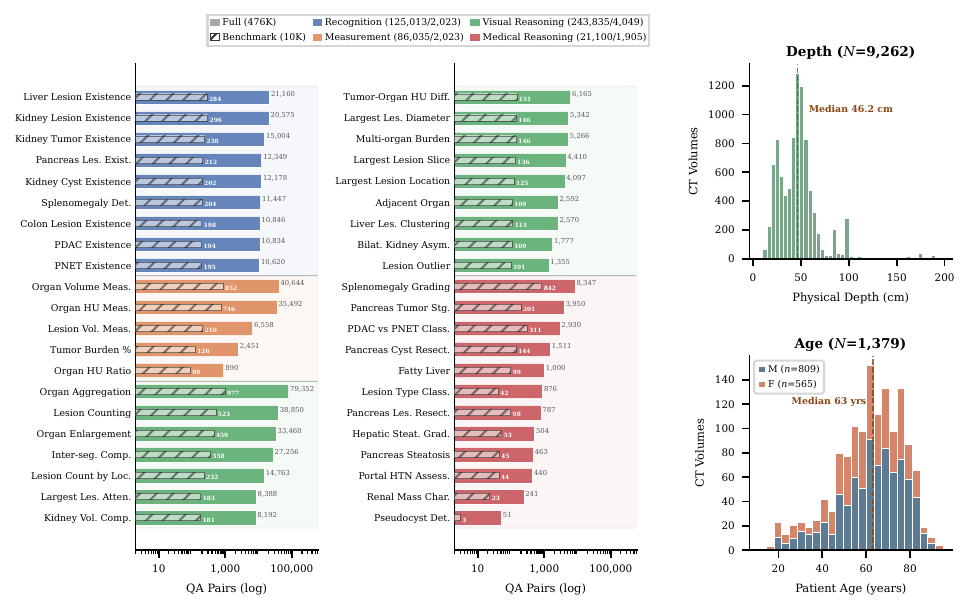}
    \caption{Statistics of DeepTumorVQA. Left: distribution of QA pairs across four main types and 42 subtypes. Right: distribution of CT volumes \wrt physical depth and patient demographics.}
    \label{fig:stat}
    \vspace{-0.3cm}
\end{figure}

The dataset contains 428K training and 48K testing QA pairs across 42 subtypes (Figure~\ref{fig:stat}). Train/test splits are patient-level with zero overlap. The benchmark evaluation subset is a balanced sample of 10,000 questions from the test set, spanning 991 CT volumes.

\subsection{Multi-Modal Input and Agent Environment}
\label{sect:agent_env}

DeepTumorVQA provides three input modalities: (1)~full 3D CT NIfTI files and preprocessed arrays at multiple resolutions, compatible with 3D specialists; (2)~2D organ-specific PNG slices and whole-volume slices; (3)~MP4 videos of axial slices within each organ's extent. All modalities are pre-computed for reproducible evaluation.

\paragraph{Agent tools.} We implement four diagnostic tools in a ReAct-style~\citep{yao2022react} agent loop (up to 8 steps): \texttt{segment\_organ(target)} returns segmentation statistics for 43 targets; \texttt{measure(target, type)} returns measurements (volume, mean HU, diameter, count); \texttt{lookup\_medical\_knowledge(query)} returns clinical criteria from a curated 27-entry knowledge base; \texttt{crop\_region(organ)} returns organ-focused multi-slice crops (vision mode only). Tool outputs are sourced from expert annotations (oracle), TotalSegmentator auto-segmentation (predicted), or visual crops (vision), depending on the evaluation mode. Full tool schemas are in Appendix~\ref{sect:tool_definitions}.

\paragraph{Evaluation modes.} In oracle mode, tools return ground-truth values from expert segmentation masks (upper bound). In predicted mode, tools return values computed from TotalSegmentator~\cite{wasserthal2023totalsegmentator} auto-segmentation (realistic deployment). In vision mode, models access only crop and knowledge tools (no quantitative measurements). For each subtype, deterministic ground-truth tool traces define the expected sequence (\eg fatty liver: \texttt{lookup(fatty liver)} $\to$ \texttt{measure(liver, HU)} $\to$ \texttt{measure(spleen, HU)}); all 42 traces are listed in Appendix~\ref{sect:gt_traces}.

\paragraph{Metrics.}
We evaluate: (1)~answer accuracy per subtype, type, and overall (10K MC questions; 95\% binomial CIs are $\pm$1.0\% at $N$=10,000, so all differences $>$2pp are statistically significant; see Appendix~\ref{sect:statistical_robustness}); (2)~end-to-end latency (seconds/sample, including all tool calls); (3)~trajectory quality, comprising tool set Jaccard ($|GT \cap Actual|/|GT \cup Actual|$), parameter accuracy (regex-based flexible matching), average steps, and valid prediction rate (fraction of samples yielding a parseable answer; Appendix~\ref{sect:trajectory_details} provides per-subtype breakdowns).

\subsection{Benchmark as Diagnostic Framework}
\label{sect:diagnostic_framework}

The four task types are not only dataset categories but also diagnostic probes. Recognition tests finding-level perception; Measurement adds quantitative evidence extraction; Visual Reasoning composes image-derived primitives; and Medical Reasoning combines these primitives with clinical criteria. This hierarchy allows us to interpret stage-wise performance drops as evidence of where a model's capability breaks, especially when combined with tool interventions that substitute or degrade specific evidence sources.

\begin{table}[t]
\centering
\scriptsize
\setlength{\tabcolsep}{2.5pt}
\caption{Diagnostic framework: each task type probes a specific model capability and is built on lower-level prerequisites, enabling failure attribution.}
\label{tab:diagnostic_framework}
\begin{tabular}{l|p{3.1cm}|p{7.1cm}}
\toprule
\textbf{Task Type} & \textbf{Prerequisite} & \textbf{Capability Probed} \\
\midrule
Recognition (9 sub.) & --- & Detect the presence of target organs or lesions in the image. \\
\hline
Measurement (5 sub.) & Recognition & Localize the target organ/lesion and read out its quantitative attributes (volume, HU, count, ratio). \\
\hline
Visual Reasoning (16 sub.) & Recognition + Measurement & Compose recognition and measurement primitives with spatial, counting, or cross-organ comparison operations over the image. \\
\hline
Medical Reasoning (12 sub.) & Recognition + Measurement + medical knowledge & Integrate measured image evidence with clinical knowledge (thresholds, grading criteria) to reach a diagnostic conclusion. \\
\bottomrule
\end{tabular}
\end{table}

This hierarchy maps directly onto our experimental interventions. Direct inference evaluates the full evidence chain end-to-end. Oracle agents replace image-derived measurements with ground-truth tool outputs, isolating reasoning over correct evidence. Predicted agents use automatic tool outputs, testing robustness to imperfect evidence. Vision-mode agents remove structured measurements and require models to recover evidence from crops and knowledge lookup. Measurement-only ablations remove images entirely and test whether models can reason from numeric evidence alone. Together, these settings turn DeepTumorVQA from a task taxonomy into an intervention-based diagnostic framework.

\section{Experiments}
\label{sect:eval}

We evaluate DeepTumorVQA along five questions: how direct VLMs fail across stages (\S\ref{sect:vlm_results}); whether tools break the measurement bottleneck (\S\ref{sect:agent_results}); whether measurement interventions explain the gains (\S\ref{sect:robustness}); how failures shift from measurement to tool use and reasoning (\S\ref{sect:error_analysis}); and whether models generalize under compositional splits (\S\ref{sect:case_study}).

\subsection{Direct Inference: The VLM Landscape}
\label{sect:vlm_results}

We evaluate \textbf{25 model configurations}: 3D-finetuned specialists (M3D-Phi3, RadFM, M3D-Llama2, Merlin), pretrained baselines, 2D VLMs (Qwen3-VL~\citep{qwen3vl2025}, Qwen3.5, Meissa~\citep{chen2025meissa}, LLaVA-OneVision~\citep{liu2024llavanext}, LLaVA-Med~\citep{li2023llava}, InternVL3~\citep{internvl2024}, HuatuoGPT~\citep{chen2024huatuogpt}, MedGemma~\citep{google2025medgemma}), 2D LoRA-finetuned models, video variants, and Gemini-3-Flash API. All answer 10,000 MC questions.

\begin{table*}[t]
\centering
\scriptsize
\setlength{\tabcolsep}{2pt}
\caption{Evaluation results on DeepTumorVQA. FT = LoRA-finetuned on 428K training pairs; WV = whole-volume slices. Human evaluations are on a 200 sample subset.}
\label{tab:vlm_results}
\begin{tabular}{ll|ccl|ccccc|c}
\toprule
\textbf{Cat.} & \textbf{Model} & \textbf{Params} & \textbf{Input} & \textbf{Notes} & \textbf{Overall} & \textbf{Recog.} & \textbf{Meas.} & \textbf{Vis.R.} & \textbf{Med.R.} & \textbf{s/samp} \\
\midrule
\multirow{2}{*}{\rotatebox{90}{\scriptsize Hum.}}
& Junior radiologist & -- & 3D vol & multi-choice & 45.0\% & 46.7\% & 37.8\% & 49.3\% & \textbf{44.0\%} & 145 \\
& Senior radiologist & -- & 3D vol & free-form & \textbf{54.5\%} & \textbf{70.0\%} & \textbf{40.0\%} & \textbf{64.0\%} & \textbf{44.0\%} & 85 \\
\midrule
\multirow{4}{*}{\rotatebox{90}{\scriptsize 3D FT}}
& M3D-Llama2~\cite{bai2024m3d} & 7B & 3D vol & & 54.0\% & 48.4\% & 70.1\% & 57.1\% & \textbf{36.5\%} & 0.33 \\
& M3D-Phi3~\cite{bai2024m3d} & 4B & 3D vol & & \textbf{59.8\%} & \textbf{66.2\%} & \textbf{70.3\%} & 65.8\% & 29.1\% & 0.26 \\
& Merlin~\cite{blankemeier2024merlin} & 7B & 3D vol & ResNet enc. & 34.0\% & 46.3\% & 24.6\% & 33.3\% & 32.3\% & 0.17 \\
& RadFM~\cite{wu2023towards} & 14B & 3D vol & Largest 3D & 58.9\% & 62.3\% & 70.1\% & \textbf{66.3\%} & 27.7\% & 1.47 \\
\midrule
\multirow{3}{*}{\rotatebox{90}{\scriptsize 3D ZS}}
& M3D-Llama2 pre.~\cite{bai2024m3d} & 7B & 3D vol & No FT & 19.0\% & 10.3\% & 23.3\% & 23.7\% & 13.6\% & 0.33 \\
& M3D-Phi3 pre.~\cite{bai2024m3d} & 4B & 3D vol & No FT & 18.9\% & 4.6\% & \textbf{25.0\%} & 25.2\% & 14.2\% & 0.26 \\
& RadFM pre.~\cite{wu2023towards} & 14B & 3D vol & No FT & \textbf{27.6\%} & \textbf{39.1\%} & 21.7\% & \textbf{25.4\%} & \textbf{26.3\%} & $\sim$2 \\
\midrule
\multirow{5}{*}{\rotatebox{90}{\scriptsize 2D FT}}
& Meissa FT~\cite{chen2025meissa} & 4B & 2D & LoRA, organ & \textbf{66.3\%} & 70.3\% & 66.7\% & 65.2\% & 64.0\% & 0.40 \\
& Qwen3-VL FT~\cite{qwen3vl2025} & 4B & 2D & LoRA, organ & 64.6\% & 69.1\% & 64.2\% & 64.3\% & 61.1\% & 0.40 \\
& Meissa FT (WV infer)~\cite{chen2025meissa} & 4B & 2D & LoRA, organ$\to$WV & \textbf{66.3\%} & 68.8\% & 67.6\% & \textbf{65.3\%} & \textbf{64.3\%} & 0.40 \\
& Qwen3-VL FT (WV infer)~\cite{qwen3vl2025} & 4B & 2D & LoRA, organ$\to$WV & 64.6\% & 69.7\% & 64.8\% & 63.6\% & 61.4\% & 0.40 \\
& Meissa FT (WV train+infer)~\cite{chen2025meissa} & 4B & 2D & LoRA, WV$\to$WV & 66.1\% & \textbf{71.0\%} & \textbf{67.7\%} & 64.3\% & 63.0\% & 0.40 \\
\midrule
\multirow{11}{*}{\rotatebox{90}{\scriptsize 2D Zero-shot}}
& HuatuoGPT~\cite{chen2024huatuogpt} & 7B & 2D & Medical & 36.0\% & 50.1\% & 28.2\% & 36.8\% & 27.7\% & 0.51 \\
& InternVL3-8B~\cite{internvl2024} & 8B & 2D & & 34.4\% & 52.5\% & \textbf{32.0\%} & 31.8\% & 23.1\% & 0.51 \\
& LLaVA-Med~\cite{li2023llava} & 7B & 2D & Medical & 32.3\% & 50.0\% & 23.4\% & 28.7\% & 30.3\% & 0.45 \\
& LLaVA-OV~\cite{liu2024llavanext} & 7B & 2D & & 38.9\% & 51.3\% & 30.4\% & 36.6\% & \textbf{39.4\%} & 1.22 \\
& MedGemma 2D~\cite{google2025medgemma} & 4B & 2D & Medical & 30.8\% & 48.3\% & 24.6\% & 26.1\% & 29.0\% & 1.18 \\
& MedGemma 3D~\cite{google2025medgemma} & 4B & 2D 85-sl. & 85-slice RGB & 29.5\% & 38.3\% & 23.7\% & 28.3\% & 29.1\% & $\sim$28 \\
& Meissa-4B~\cite{chen2025meissa} & 4B & 2D & Medical & 38.2\% & 51.5\% & 30.8\% & \textbf{39.8\%} & 28.5\% & 0.40 \\
& Qwen3-VL-4B~\cite{qwen3vl2025} & 4B & 2D & & 37.8\% & 56.6\% & 29.1\% & 38.8\% & 25.0\% & 0.40 \\
& Qwen3-VL-8B~\cite{qwen3vl2025} & 8B & 2D & & 34.4\% & 50.7\% & 27.7\% & 35.6\% & 21.8\% & 0.40 \\
& Qwen3.5-9B~\cite{qwen2025qwen35} & 9B & 2D & & 38.5\% & \textbf{58.4\%} & 29.3\% & 38.4\% & 27.4\% & 0.68 \\
& Qwen3.5-35B~\cite{qwen2025qwen35} & 35B(3B) & 2D & MoE & \textbf{39.9\%} & 53.0\% & 31.4\% & 38.2\% & 38.7\% & 1.58 \\
\midrule
\multirow{3}{*}{\rotatebox{90}{\scriptsize Video}}
& Meissa-4B (video)~\cite{chen2025meissa} & 4B & Video & & 37.9\% & 45.9\% & 30.3\% & \textbf{41.3\%} & \textbf{30.0\%} & $\sim$7 \\
& Qwen3-VL-4B (vid.)~\cite{qwen3vl2025} & 4B & Video & & \textbf{38.9\%} & \textbf{58.3\%} & \textbf{31.1\%} & 39.2\% & 25.7\% & 0.40 \\
& Qwen3-VL-8B (vid.)~\cite{qwen3vl2025} & 8B & Video & & 36.2\% & 55.1\% & 28.5\% & 36.6\% & 23.4\% & 0.40 \\
\midrule
\multirow{1}{*}{\rotatebox{90}{\scriptsize API}}
& Gemini-3-Flash~\cite{google2025gemini3} & -- & 2D & API & 38.8\% & 42.6\% & 35.7\% & 40.7\% & 33.9\% & $\sim$6 \\
\bottomrule
\end{tabular}
\vspace{-0.3cm}
\end{table*}

Table~\ref{tab:vlm_results} reveals a striking asymmetry. All zero-shot 2D models perform best on Recognition (50--58\%) but collapse on four-option Measurement (24--32\%), near chance. This measurement ceiling cascades into Visual and Medical Reasoning, because most clinical diagnoses require comparing measured values against thresholds.
Zero-shot 3D models hover near random (M3D-Phi3: 18.9\%), confirming that general volumetric pretraining alone does not transfer to clinical VQA.

Finetuning reverses the ranking. The best 3D specialist M3D-Phi3 jumps to 59.8\%, but Meissa reaches 66.3\% with only 2D input. Neither larger models, video, nor multi-slice pseudo-3D bridges this gap without task supervision. Whole-volume train/infer combinations (WV rows) change accuracy by at most 0.7pp, ruling out the localization prior (Appendix~\ref{sect:input_modality}); free-form evaluation on the full benchmark nearly preserves every ranking, ruling out MC as a format confound (Appendix~\ref{sect:freeform_eval}). Two practicing radiologists (7 and 13 years experience, independent from our annotators) evaluated on a 200-question stratified subset reach 45.0\% and 54.5\% respectively, both below 2D-FT (66.3\%); experience helps Recognition ($+$23pp) and Visual Reasoning ($+$15pp) but not Measurement, where eyeballing precise quantities from raw CT is the human ceiling (Appendix~\ref{sect:human_eval}). Per-subtype and demographic breakdowns are in Appendices~\ref{sect:per_subtype},~\ref{sect:demographic_breakdown}.

\subsection{Agent-Based Reasoning}
\label{sect:agent_results}

We evaluate Qwen3.5-9B, Meissa-4B, and Gemini-3-Flash in oracle/vision modes, plus Meissa-4B after SFT on ground-truth tool traces. A \textit{predicted} mode using TotalSegmentator~\cite{wasserthal2023totalsegmentator} auto-segmentation bridges \textit{oracle} mode with GT masks and \textit{vision} mode with no segmentation.

\begin{table*}[t]
\centering
\scriptsize
\setlength{\tabcolsep}{2pt}
\caption{Agent evaluation results. $\Delta$ = change vs.\ direct inference (pp). $\dagger$Trained on ground-truth traces. Oracle = GT masks; Predicted = TotalSegmentator auto-segmentation; Vision = crops only. Best per-column in \textbf{bold} (for Steps and s/samp, lower is better).}
\label{tab:agent_results}
\begin{tabular}{ll|cc|ccccc|c|c|cccc}
\toprule
& & & & \multicolumn{5}{c|}{\textbf{Answer Accuracy}} & & & \multicolumn{4}{c}{\textbf{Trajectory Quality}} \\
\textbf{Mode} & \textbf{Model} & \textbf{Params} & \textbf{Input} & \textbf{Overall} & \textbf{Recog.} & \textbf{Meas.} & \textbf{Vis.R.} & \textbf{Med.R.} & \textbf{$\Delta$} & \textbf{Steps} & \textbf{Jacc.} & \textbf{Param} & \textbf{Valid} & \textbf{s/samp} \\
\midrule
\multirow{4}{*}{Oracle}
& Gemini-3-Flash~\cite{google2025gemini3} & -- & 2D+tools & 61.9\% & 66.4\% & 60.0\% & \textbf{58.0\%} & \textbf{67.8\%} & +23.1 & 7.2 & 0.88 & \textbf{0.95} & 97.8\% & $\sim$43 \\
& Meissa-4B~\cite{chen2025meissa} & 4B & 2D+tools & 46.0\% & 44.9\% & 64.4\% & 35.2\% & 50.7\% & +7.8 & 5.5 & 0.83 & 0.80 & 95.1\% & $\sim$13 \\
& Meissa SFT$^\dagger$~\cite{chen2025meissa} & 4B & 2D+tools & \textbf{63.8\%} & 68.2\% & \textbf{78.3\%} & 55.6\% & 61.5\% & \textbf{+25.6} & 3.8 & \textbf{0.96} & 0.85 & 99.9\% & $\sim$\textbf{5} \\
& Qwen3.5-9B~\cite{qwen2025qwen35} & 9B & 2D+tools & 48.5\% & 50.1\% & 64.8\% & 38.0\% & 52.3\% & +10.0 & 4.2 & 0.75 & 0.56 & 88.1\% & $\sim$17 \\
\midrule
\multirow{2}{*}{Predicted}
& Meissa SFT$^\dagger$~\cite{chen2025meissa} & 4B & 2D+tools & 60.9\% & \textbf{69.9\%} & 60.8\% & 55.5\% & 63.2\% & +22.7 & 3.7 & 0.95 & 0.84 & \textbf{100\%} & $\sim$6 \\
& Qwen3.5-9B~\cite{qwen2025qwen35} & 9B & 2D+tools & 43.9\% & 48.9\% & 46.9\% & 31.6\% & 61.3\% & +5.4 & 7.1 & 0.80 & 0.78 & 82.0\% & $\sim$55 \\
\midrule
\multirow{3}{*}{Vision}
& Gemini-3-Flash~\cite{google2025gemini3} & -- & 2D+tools & 35.3\% & 49.6\% & 26.5\% & 33.3\% & 33.7\% & $-$3.5 & 3.1 & 0.07 & 0.07 & 99.2\% & $\sim$17 \\
& Meissa-4B~\cite{chen2025meissa} & 4B & 2D+tools & 31.5\% & 41.1\% & 26.2\% & 32.6\% & 24.5\% & $-$6.7 & \textbf{2.3} & 0.84 & 0.85 & 99.8\% & $\sim$9 \\
& Qwen3.5-9B~\cite{qwen2025qwen35} & 9B & 2D+tools & 40.3\% & 59.7\% & 29.5\% & 38.6\% & 35.0\% & +1.8 & 2.4 & 0.92 & 0.93 & 99.9\% & $\sim$25 \\
\bottomrule
\end{tabular}
\end{table*}

Oracle tools transform the error landscape: Measurement jumps from 29--31\% to 64--65\%, Medical Reasoning from 27--28\% to 51--54\%. VLMs already possess substantial clinical reasoning, locked behind a measurement bottleneck that quantitative tools unlock; Gemini-3-Flash reaches 61.9\% with oracle tools, confirming that stronger backbones extract more from tool-provided evidence.

Predicted mode with real segmentation retains most gains: Meissa SFT drops only 2.9pp, keeping 93\% of the oracle-to-direct gap; on 12 organ-centric subtypes retention is 85\%, while lesion-centric ones may reflect partial fallback (Appendix~\ref{sect:predicted_matrix}). Vision-only tools backfire: Meissa-4B vision falls below direct inference (31.5\% vs. 38.2\%), as models hallucinate quantitative estimates from crops.

Training Meissa-4B on 20K tool traces synthesized with the same logic as data annotation yields 63.8\% (+17.8pp over zero-shot oracle, +25.6pp over direct), with Measurement reaching 78.3\% and surpassing 3D specialists (70\%). Trajectory supervision also teaches efficient workflows: average steps drop from 5.5 to 3.8, and the 8-step hit rate from 23.2\% to 0.3\%.

\subsection{Causal Analysis with Agents: Measurement is the Bottleneck}
\label{sect:robustness}

The preceding sections show tools help, but correlation is not causation. Table~\ref{tab:ablation_results} presents systematic ablations identifying reliable quantitative measurement as both necessary and, for capable backbones, sufficient within the benchmark's program-defined task space.

\begin{table}[t]
\centering
\scriptsize
\setlength{\tabcolsep}{2.5pt}
\caption{Causal analysis with agents. Measurement-only: text-only inference with pre-computed measurements (no image). $\Delta$ vs.\ oracle agent (Qwen3.5-9B: 48.5\%).}
\label{tab:ablation_results}
\begin{tabular}{l|ccccc|c}
\toprule
\textbf{Condition} & \textbf{Overall} & \textbf{Recog.} & \textbf{Meas.} & \textbf{Vis.R.} & \textbf{Med.R.} & \textbf{$\Delta$} \\
\midrule
\multicolumn{7}{l}{\emph{Baselines}} \\
Direct (image only) & 38.5\% & 58.4\% & 29.3\% & 38.4\% & 27.4\% & -- \\
Oracle agent (img+tools) & 48.5\% & 50.1\% & 64.8\% & 38.0\% & 52.3\% & +10.0 \\
\midrule
\multicolumn{7}{l}{\emph{Measurement is necessary}} \\
No measure tool & 32.3\% & 57.1\% & 23.5\% & 25.7\% & 28.2\% & $-$16.2 \\
\midrule
\multicolumn{7}{l}{\emph{For capable backbones, measurements alone exceed oracle agent}} \\
Meas-only Qwen3.5-9B & 54.4\% & 68.8\% & 63.3\% & 40.6\% & 59.1\% & +5.9 \\
{\color{gray}Meas-only Meissa-4B} & {\color{gray}33.0\%} & {\color{gray}53.4\%} & {\color{gray}50.9\%} & {\color{gray}17.8\%} & {\color{gray}24.8\%} & {\color{gray}$-$15.5} \\
\midrule
\multicolumn{7}{l}{\emph{Measurement precision matters}} \\
Noise $\sigma$=10\% & 38.3\% & 45.9\% & 39.6\% & 30.9\% & 45.5\% & $-$10.2 \\
Noise $\sigma$=50\% & 35.8\% & 46.2\% & 22.1\% & 31.1\% & 47.6\% & $-$12.7 \\
\bottomrule
\end{tabular}
\end{table}

\textbf{Measurement is necessary.} Removing the measurement tool causes $-$16.2pp overall and $-$24.1pp on Medical Reasoning. Recognition improves without measurement from 50.1\% to 57.1\%, as the model falls back to image-based recognition (sensitivity/specificity breakdown in Appendix~\ref{sect:sens_spec}).

\textbf{For capable backbones, measurements alone exceed the oracle agent.} Measurements alone without image yield 54.4\% for Qwen3.5-9B, surpassing the oracle agent; Medical Reasoning reaches 59.1\% vs.\ 27.4\% from image. This is backbone-dependent and task-space-bounded rather than a general clinical claim (Appendix~\ref{sect:limitations_detailed}).


\textbf{Precision matters.} Even 10\% noise drops measurement accuracy from 65\% to 40\%, with damage saturating by 50\% noise. Distractor spacing does not explain this sensitivity ($r$=0.25; Appendix~\ref{sect:distractor_analysis}).


\subsection{Error Analysis: From Scattered to Concentrated}
\label{sect:error_analysis}

\begin{figure}[t]
\centering
\includegraphics[
        width=1.0\linewidth,
        trim={0.0cm 0.2cm 0.0cm 0.2cm},
        clip
    ]{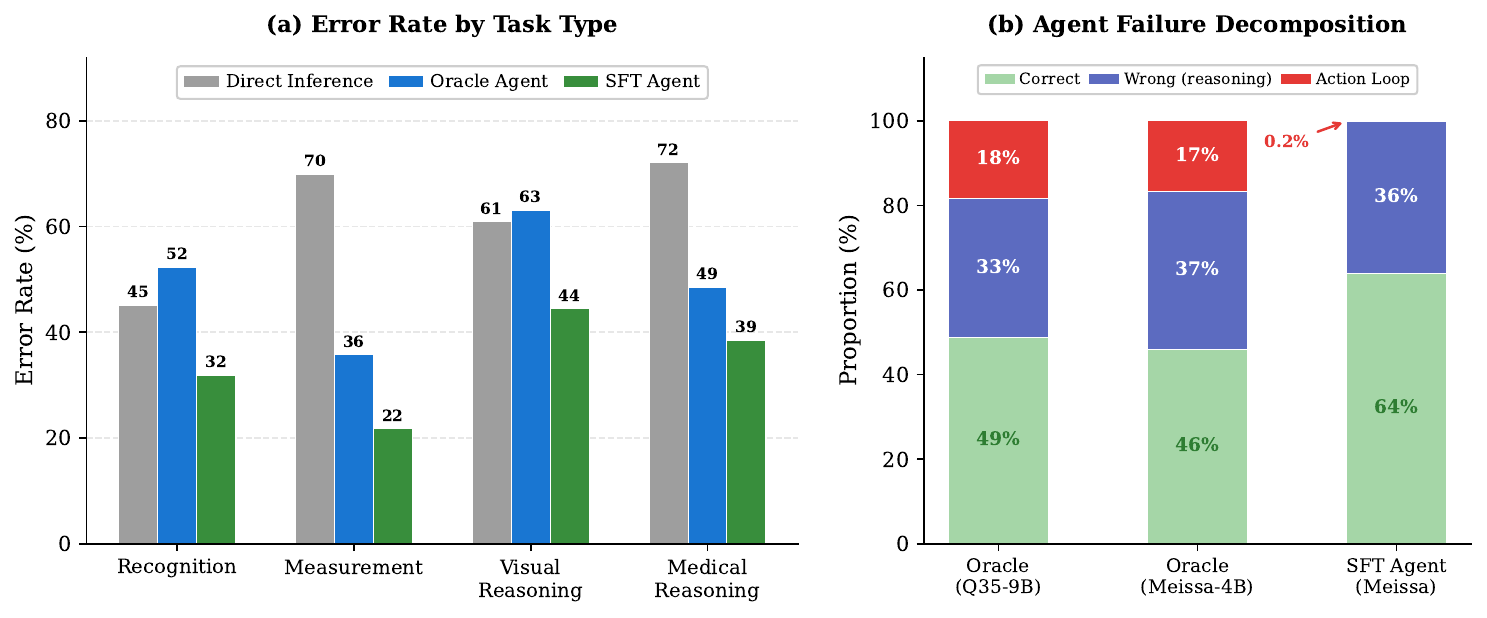}
\caption{\textbf{Error landscape transformation.} (a)~Error rate by task type across three paradigms. Oracle tools halve Measurement errors (70\%$\to$36\%) and sharply reduce Medical Reasoning errors (72\%$\to$49\%); SFT further lowers all categories. (b)~Agent failure decomposition: action loops (step-limit hits, red) account for 17--18\% of all questions in oracle agents, concentrated in multi-step Visual Reasoning; SFT virtually eliminates this tool-use failure mode (0.2\%).}
\label{fig:error_analysis}
\end{figure}

Tools reshape where models fail (Figure~\ref{fig:error_analysis}a). Without tools, errors are diffuse. Oracle tools concentrate gains on quantitative tasks: Measurement errors drop from 70\% to 36\%, Medical Reasoning from 72\% to 49\%, while Recognition sees little improvement.

A new failure mode emerges in tool use: \textbf{action loops}. Oracle agents hit the step limit on 17--18\% of questions (Figure~\ref{fig:error_analysis}b), with Visual Reasoning most affected due to multi-step tool chains. The knowledge tool is invoked in only 1.3\% of tool calls despite being needed for 19\% of questions.

Agent SFT resolves these tool-use failures: action loops drop from 18\% to 0.2\%, accuracy rises from 46\% to 64\%. Residual errors are reasoning failures where correct tool outputs lead to wrong conclusions, a harder frontier (Appendices~\ref{sect:appendix_error} and~\ref{sect:failure_cases}). The framework thus reveals successive frontiers: measurement before tools, tool use with tools (action loops), and reasoning over correct evidence after SFT (Table~\ref{tab:error_attribution} in Appendix).

\subsection{Compositional Generalization of Models and Agents}
\label{sect:case_study}

A CLEVR-style split on lesions binned by volume with four held-out (organ, size) pairs (OOD $n$=2,517; IID $n$=1,437) yields a small IID--OOD gap ($-$5.6pp to $+$3.9pp), suggesting that the main trends are not driven by memorizing observed organ-size combinations. Agents show negative gaps (mean $-$4.1pp) since tool-based measurements are distribution-independent; Meissa SFT maintains this trend ($-$4.4pp), supporting compositional generalization (Appendix~\ref{sect:compositional_full}).

\section{Discussion}
\label{sect:discussion}

\paragraph{Design Principles.}
Three principles emerge: (1)~\emph{quantitative, not visual, tools matter}: measurements deliver $+$7.8pp while crop/zoom costs $-$6.7pp; (2)~\emph{trace SFT resolves tool use, exposing reasoning}: $+$18pp with action loops dropping from 18\% to 0.2\%, leaving reasoning over correct evidence as the remaining frontier; (3)~\emph{imperfect tools still help}: realistic auto-segmentation retains 93\% of oracle gains, with non-uniform noise sensitivity (Appendix~\ref{sect:robustness_appendix}).

\paragraph{Why 2D Finetuning Surpasses 3D Specialists.}
Task supervision dominates representation: 2D LoRA (66.3\%) beats the best 3D specialist (59.8\%) because it aligns the model with the benchmark's evidence primitives and answer space; whole-volume controls (\S\ref{sect:vlm_results}) rule out the localization prior.
\textbf{Limitations}: sufficiency holds within structured programs; MC preserves rankings vs.\ tolerance scoring ($r$=0.999); annotation quality bounds ground truth (Appendices~\ref{sect:limitations_detailed},~\ref{sect:tolerance_eval},~\ref{sect:seg_quality}).

\section{Conclusion}
\label{sect:conclusion}

We introduced DeepTumorVQA, a 3D CT benchmark that turns tumor VQA from single-score answering into stage-wise diagnostic evaluation. Its core contribution is a clinically grounded evidence hierarchy that links recognition, measurement, visual reasoning, and medical reasoning through deterministic evidence primitives. This design supports the same diagnostic tasks across direct VLM inference and tool-augmented agents, while enabling controlled interventions over evidence availability, tool quality, and agent trajectories. With 476K questions across 42 clinical subtypes and 9{,}262 CT volumes, DeepTumorVQA provides a testbed not only for ranking models, but also for identifying where and why they fail.
We hope DeepTumorVQA encourages future medical VLM and agent research to move beyond aggregate accuracy toward evidence-grounded, tool-aware, and failure-attributable evaluation.

\begin{ack}
This work was supported by the Lustgarten Foundation for Pancreatic Cancer Research and the Patrick J. McGovern Foundation Award.
\end{ack}

\bibliographystyle{plainnat}
\bibliography{TumorVQA}

@article{yang2024advancing,
  title={Advancing multimodal medical capabilities of Gemini},
  author={Yang, Lin and Xu, Shawn and Sellergren, Andrew and Kohlberger, Timo and Zhou, Yuchen and Ktena, Ira and Kiraly, Atilla and Ahmed, Faruk and Hormozdiari, Farhad and Jaroensri, Tiam and others},
  journal={arXiv preprint arXiv:2405.03162},
  year={2024}
}

@article{bai2024m3d,
  title={M3d: Advancing 3d medical image analysis with multi-modal large language models},
  author={Bai, Fan and Du, Yuxin and Huang, Tiejun and Meng, Max Q-H and Zhao, Bo},
  journal={arXiv preprint arXiv:2404.00578},
  year={2024}
}

@inproceedings{hu2024omnimedvqa,
  title={Omnimedvqa: A new large-scale comprehensive evaluation benchmark for medical lvlm},
  author={Hu, Yutao and Li, Tianbin and Lu, Quanfeng and Shao, Wenqi and He, Junjun and Qiao, Yu and Luo, Ping},
  booktitle={Proceedings of the IEEE/CVF Conference on Computer Vision and Pattern Recognition},
  pages={22170--22183},
  year={2024}
}

@article{lau2018dataset,
  title={A dataset of clinically generated visual questions and answers about radiology images},
  author={Lau, Jason J and Gayen, Soumya and Ben Abacha, Asma and Demner-Fushman, Dina},
  journal={Scientific data},
  volume={5},
  number={1},
  pages={1--10},
  year={2018},
  publisher={Nature Publishing Group}
}

@article{zhang2023pmc,
  title={Pmc-vqa: Visual instruction tuning for medical visual question answering},
  author={Zhang, Xiaoman and Wu, Chaoyi and Zhao, Ziheng and Lin, Weixiong and Zhang, Ya and Wang, Yanfeng and Xie, Weidi},
  journal={arXiv preprint arXiv:2305.10415},
  year={2023}
}

@article{wu2023towards,
  title={Towards generalist foundation model for radiology by leveraging web-scale 2D\&3D medical data},
  author={Wu, Chaoyi and Zhang, Xiaoman and Zhang, Ya and Wang, Yanfeng and Xie, Weidi},
  journal={arXiv preprint arXiv:2308.02463},
  year={2023}
}

@article{blankemeier2024merlin,
  title={Merlin: A vision language foundation model for 3d computed tomography},
  author={Blankemeier, Louis and Cohen, Joseph Paul and Kumar, Ashwin and Van Veen, Dave and Gardezi, Syed Jamal Safdar and Paschali, Magdalini and Chen, Zhihong and Delbrouck, Jean-Benoit and Reis, Eduardo and Truyts, Cesar and others},
  journal={Research Square},
  pages={rs--3},
  year={2024}
}

@article{ji2022amos,
  title={Amos: A large-scale abdominal multi-organ benchmark for versatile medical image segmentation},
  author={Ji, Yuanfeng and Bai, Haotian and Ge, Chongjian and Yang, Jie and Zhu, Ye and Zhang, Ruimao and Li, Zhen and Zhang, Lingyan and Ma, Wanling and Wan, Xiang and others},
  journal={Advances in Neural Information Processing Systems},
  volume={35},
  pages={36722--36732},
  year={2022}
}

@article{hamamci2024developing,
  title={Developing Generalist Foundation Models from a Multimodal Dataset for 3D Computed Tomography},
  author={Hamamci, Ibrahim Ethem and Er, Sezgin and Almas, Furkan and Simsek, Ayse Gulnihan and Esirgun, Sevval Nil and Dogan, Irem and Dasdelen, Muhammed Furkan and Durugol, Omer Faruk and Wittmann, Bastian and Amiranashvili, Tamaz and others},
  journal={arXiv preprint arXiv:2403.17834},
  year={2024}
}

@article{li2023llava,
  title={Llava-med: Training a large language-and-vision assistant for biomedicine in one day},
  author={Li, Chunyuan and Wong, Cliff and Zhang, Sheng and Usuyama, Naoto and Liu, Haotian and Yang, Jianwei and Naumann, Tristan and Poon, Hoifung and Gao, Jianfeng},
  journal={Advances in Neural Information Processing Systems},
  volume={36},
  pages={28541--28564},
  year={2023}
}

@inproceedings{johnson2017clevr,
  title={Clevr: A diagnostic dataset for compositional language and elementary visual reasoning},
  author={Johnson, Justin and Hariharan, Bharath and Van Der Maaten, Laurens and Fei-Fei, Li and Lawrence Zitnick, C and Girshick, Ross},
  booktitle={Proceedings of the IEEE conference on computer vision and pattern recognition},
  pages={2901--2910},
  year={2017}
}

@inproceedings{roth2015deeporgan,
  title={Deeporgan: Multi-level deep convolutional networks for automated pancreas segmentation},
  author={Roth, Holger R and Lu, Le and Farag, Amal and Shin, Hoo-Chang and Liu, Jiamin and Turkbey, Evrim B and Summers, Ronald M},
  booktitle={International conference on medical image computing and computer-assisted intervention},
  pages={556--564},
  year={2015},
  organization={Springer}
}

@inproceedings{landman2015miccai,
  title={Miccai multi-atlas labeling beyond the cranial vault--workshop and challenge},
  author={Landman, Bennett and Xu, Zhoubing and Igelsias, J and Styner, Martin and Langerak, T and Klein, Arno},
  booktitle={Proc. MICCAI Multi-Atlas Labeling Beyond Cranial Vault---Workshop Challenge},
  volume={5},
  pages={12},
  year={2015}
}

@article{ma2021abdomenct,
  title={Abdomenct-1k: Is abdominal organ segmentation a solved problem},
  author={Ma, Jun and Zhang, Yao and Gu, Song and Zhu, Cheng and Ge, Cheng and Zhang, Yichi and An, Xingle and Wang, Congcong and Wang, Qiyuan and Liu, Xin and others},
  journal={IEEE Transactions on Pattern Analysis and Machine Intelligence},
  year={2021},
  publisher={IEEE}
}

@inproceedings{valindria2018multi,
  title={Multi-modal learning from unpaired images: Application to multi-organ segmentation in CT and MRI},
  author={Valindria, Vanya V and Pawlowski, Nick and Rajchl, Martin and Lavdas, Ioannis and Aboagye, Eric O and Rockall, Andrea G and Rueckert, Daniel and Glocker, Ben},
  booktitle={2018 IEEE winter conference on applications of computer vision (WACV)},
  pages={547--556},
  year={2018},
  organization={IEEE}
}

@misc{rsna-2023-abdominal-trauma-detection,
    author = {Colak, Errol and Lin, Hui-Ming and Ball, Robyn and Davis, Melissa and Flanders, Adam and Jalal, Sabeena and Magudia, Kirti and Marinelli, Brett and Nicolaou, Savvas and Prevedello, Luciano and Rudie, Jeff and Shih, George and Vazirabad, Maryam and Mongan, John},
    title = {RSNA 2023 Abdominal Trauma Detection},
    publisher = {Kaggle},
    year = {2023},
    url = {https://kaggle.com/competitions/rsna-2023-abdominal-trauma-detection}
}

@article{bilic2019liver,
  title={The liver tumor segmentation benchmark (lits)},
  author={Bilic, Patrick and Christ, Patrick Ferdinand and Vorontsov, Eugene and Chlebus, Grzegorz and Chen, Hao and Dou, Qi and Fu, Chi-Wing and Han, Xiao and Heng, Pheng-Ann and Hesser, J{\"u}rgen and others},
  journal={arXiv preprint arXiv:1901.04056},
  year={2019}
}

@article{antonelli2021medical,
  title={The Medical Segmentation Decathlon},
  author={Antonelli, Michela and Reinke, Annika and Bakas, Spyridon and Farahani, Keyvan and Landman, Bennett A and Litjens, Geert and Menze, Bjoern and Ronneberger, Olaf and Summers, Ronald M and van Ginneken, Bram and others},
  journal={arXiv preprint arXiv:2106.05735},
  year={2021}
}

@misc{heller2020international,
  title={An international challenge to use artificial intelligence to define the state-of-the-art in kidney and kidney tumor segmentation in CT imaging.},
  author={Heller, Nicholas and McSweeney, Sean and Peterson, Matthew Thomas and Peterson, Sarah and Rickman, Jack and Stai, Bethany and Tejpaul, Resha and Oestreich, Makinna and Blake, Paul and Rosenberg, Joel and others},
  year={2020},
  publisher={American Society of Clinical Oncology}
}

@article{luo2021word,
  title={WORD: Revisiting Organs Segmentation in the Whole Abdominal Region},
  author={Luo, Xiangde and Liao, Wenjun and Xiao, Jianghong and Song, Tao and Zhang, Xiaofan and Li, Kang and Wang, Guotai and Zhang, Shaoting},
  journal={arXiv preprint arXiv:2111.02403},
  year={2021}
}

@article{rister2020ct,
  title={CT-ORG, a new dataset for multiple organ segmentation in computed tomography},
  author={Rister, Blaine and Yi, Darvin and Shivakumar, Kaushik and Nobashi, Tomomi and Rubin, Daniel L},
  journal={Scientific Data},
  volume={7},
  number={1},
  pages={1--9},
  year={2020},
  publisher={Nature Publishing Group}
}

@article{ma2022fast,
  title={Fast and low-GPU-memory abdomen CT organ segmentation: the flare challenge},
  author={Ma, Jun and Zhang, Yao and Gu, Song and An, Xingle and Wang, Zhihe and Ge, Cheng and Wang, Congcong and Zhang, Fan and Wang, Yu and Xu, Yinan and others},
  journal={Medical Image Analysis},
  volume={82},
  pages={102616},
  year={2022},
  publisher={Elsevier}
}

@article{zeb2012computed,
  title={Computed tomography scans in the evaluation of fatty liver disease in a population based study: the multi-ethnic study of atherosclerosis},
  author={Zeb, Irfan and Li, Dong and Nasir, Khurram and Katz, Ronit and Larijani, Vahid N and Budoff, Matthew J},
  journal={Academic radiology},
  volume={19},
  number={7},
  pages={811--818},
  year={2012},
  publisher={Elsevier}
}

@article{guneyli2022computed,
  title={Computed tomography evaluation of pancreatic steatosis: correlation with COVID-19 prognosis},
  author={Guneyli, Serkan and Dogan, Hakan and Esengur, Omer Tarik and Hassoy, Hur},
  journal={Future Virology},
  volume={17},
  number={4},
  pages={231--237},
  year={2022},
  publisher={Taylor \& Francis}
}

@article{bassi2025radgpt,
  title={RadGPT: Constructing 3D Image-Text Tumor Datasets},
  author={Bassi, Pedro RAS and Yavuz, Mehmet Can and Wang, Kang and Chen, Xiaoxi and Li, Wenxuan and Decherchi, Sergio and Cavalli, Andrea and Yang, Yang and Yuille, Alan and Zhou, Zongwei},
  journal={arXiv preprint arXiv:2501.04678},
  year={2025}
}

@article{kodama2007comparison,
  title={Comparison of CT methods for determining the fat content of the liver},
  author={Kodama, Yuji and Ng, Connie S and Wu, Tsung-Teh and Ayers, Gregory D and Curley, Steven A and Abdalla, Eddie K and Vauthey, Jean-Nicolas and Charnsangavej, Chusilp},
  journal={American Journal of Roentgenology},
  volume={188},
  number={5},
  pages={1307--1312},
  year={2007},
  publisher={Am Roentgen Ray Soc}
}

@article{bezerra2005spleen,
  title={Determination of splenomegaly by CT: is there a place for a single measurement?},
  author={Bezerra, Arlindo Sprenger and D'Ippolito, Giuseppe and Frias, Luzia Akiko and Ozaki, Mauro and Szejnfeld, Jacob and Fuchs, Marcelo},
  journal={American Journal of Roentgenology},
  volume={184},
  number={5},
  pages={1510--1513},
  year={2005},
  publisher={Am Roentgen Ray Soc}
}

@misc{nccn2024pancreatic,
  title={NCCN Clinical Practice Guidelines in Oncology: Pancreatic Adenocarcinoma},
  author={{National Comprehensive Cancer Network}},
  year={2024},
  url={https://www.nccn.org/guidelines/guidelines-detail?category=1&id=1455}
}

@article{silverman2019bosniak,
  title={Bosniak classification of cystic renal masses, version 2019: an update proposal and needs assessment},
  author={Silverman, Stuart G and Pedrosa, Ivan and Ellis, James H and Hindman, Nicole M and Schieda, Nicola and Smith, Andrew D and Remer, Erick M and Shinagare, Atul B and Curci, Natalie E and Manganaro, Lucia and others},
  journal={Radiology},
  volume={292},
  number={2},
  pages={475--488},
  year={2019},
  publisher={Radiological Society of North America}
}

@article{allen2011pseudocyst,
  title={Cystic lesions of the pancreas},
  author={Allen, Peter J and D'Angelica, Michael and Gonen, Mithat and Jaques, David P and Coit, Daniel G and Jarnagin, William R and DeMatteo, Ronald and Fong, Yuman and Blumgart, Leslie H and Brennan, Murray F},
  journal={Annals of Surgery},
  volume={254},
  pages={680--686},
  year={2011}
}

@article{harbin1980portal,
  title={Diagnosis of cirrhosis based on regional changes in hepatic morphology: a radiological and pathological analysis},
  author={Harbin, William P and Robert, Nicholas J and Ferrucci Jr, Joseph T},
  journal={Radiology},
  volume={135},
  number={2},
  pages={273--283},
  year={1980},
  publisher={Radiological Society of North America}
}

@misc{ajcc2017cancer,
  title={AJCC Cancer Staging Manual},
  author={{American Joint Committee on Cancer}},
  edition={8th},
  year={2017},
  publisher={Springer}
}

@article{tanaka2012revisions,
  title={Revisions of international consensus Fukuoka guidelines for the management of IPMN of the pancreas},
  author={Tanaka, Masao and Fern{\'a}ndez-del Castillo, Carlos and Adsay, Volkan and Chari, Suresh and Falconi, Massimo and Jang, Jin-Young and Kimura, Wataru and Levy, Philip and Pitman, Martha B and Schmidt, C Max and others},
  journal={Pancreatology},
  volume={12},
  number={3},
  pages={183--197},
  year={2012},
  publisher={Elsevier}
}

@article{schmidgall2025medagentbench,
  title={MedAgentBench: Dataset for Benchmarking LLMs as Agents in Medical Applications},
  author={Schmidgall, Samuel and Kim, Baekjin and Goh, Hyunjin and Jiang, Albert Q. and Grabowski, Alex and Chauhan, Aarav and others},
  journal={NEJM AI},
  year={2025},
  publisher={Massachusetts Medical Society}
}

@article{schmidgall2024agentclinic,
  title={AgentClinic: A Multimodal Agent Benchmark to Evaluate AI in Simulated Clinical Environments},
  author={Schmidgall, Samuel and Ziaei, Rojin and Harris, Carl and Reis, Eduardo Pontes and Yarlagadda, Sheshank and others},
  journal={arXiv preprint arXiv:2405.07960},
  year={2024}
}

@inproceedings{gao2024medchain,
  title={MedChain: Bridging the Gap Between LLM Agents and Clinical Practice through Interactive Sequential Benchmarking},
  author={Gao, Jie and Chen, Yuanyuan and Wang, Guozheng and Li, Hongyi and Yang, Jianye and others},
  booktitle={Advances in Neural Information Processing Systems},
  year={2025}
}

@inproceedings{eltawil2025medrax,
  title={MedRAX: Medical Reasoning Agent for Chest X-ray},
  author={Eltawil, Adibvafa and Cheema, Fares and Shin, Yong Suk Paul and Rubin, Jonathan and Chen, Sheyang and others},
  booktitle={Proceedings of the International Conference on Machine Learning},
  year={2025}
}

@inproceedings{yu2025medsgbench,
  title={MedSG-Bench: Benchmarking Medical Image Sequences Grounding Capability of Multimodal Large Language Models},
  author={Yu, Jingkun and Ge, Yixiao and Wang, Rui and Ying, Shan and others},
  booktitle={Advances in Neural Information Processing Systems},
  year={2025}
}

@inproceedings{tang2025_3drad,
  title={3D-RAD: A Large-Scale 3D Radiology Med-VQA Dataset},
  author={Tang, Xiaoxiao and others},
  booktitle={Advances in Neural Information Processing Systems},
  year={2025}
}

@article{chen2025meissa,
  title={Meissa: Multi-modal Medical Agentic Intelligence},
  author={Chen, Yixiong and Bassi, Pedro R. A. S. and Zhou, Xinze and Xiao, Wenjie and Zhou, Zongwei and Yuille, Alan},
  journal={arXiv preprint arXiv:2603.09018},
  year={2026}
}

@inproceedings{hu2021lora,
  title={{LoRA}: Low-Rank Adaptation of Large Language Models},
  author={Hu, Edward J and Shen, Yelong and Wallis, Phillip and Allen-Zhu, Zeyuan and Li, Yuanzhi and Wang, Shean and Wang, Lu and Chen, Weizhu},
  booktitle={International Conference on Learning Representations},
  year={2022}
}

@article{wasserthal2023totalsegmentator,
  title={TotalSegmentator: Robust Segmentation of 104 Anatomic Structures in CT Images},
  author={Wasserthal, Jakob and Breit, Hanns-Christian and Meyer, Manfred T and Pradella, Maurice and Hinck, Daniel and Saez, Alexander W and Semler, Tobias and Stritzel, Florian and Segeroth, Martin and Josi, Joshy},
  journal={Radiology: Artificial Intelligence},
  volume={5},
  number={5},
  pages={e230024},
  year={2023}
}

@inproceedings{yao2022react,
  title={ReAct: Synergizing Reasoning and Acting in Language Models},
  author={Yao, Shunyu and Zhao, Jeffrey and Yu, Dian and Du, Nan and Shafran, Izhak and Narasimhan, Karthik and Cao, Yuan},
  booktitle={International Conference on Learning Representations},
  year={2023}
}

@article{qwen3vl2025,
  title={Qwen2.5-VL Technical Report},
  author={Bai, Shuai and Chen, Keqin and Liu, Xuejing and Wang, Jialin and Ge, Wenbin and Song, Sibo and Dang, Kai and Wang, Peng and Wang, Shijie and Tang, Jun and others},
  journal={arXiv preprint arXiv:2502.13923},
  year={2025}
}

@article{chen2024huatuogpt,
  title={HuatuoGPT-Vision, Towards Injecting Medical Visual Knowledge into Multimodal LLMs at Scale},
  author={Chen, Junying and Gui, Ruyi and Chi, Anningzhe and Peng, Anqi and Zhang, Zhongzhi and Wan, Xiang and others},
  journal={arXiv preprint arXiv:2406.19280},
  year={2024}
}

@misc{google2025medgemma,
  title={MedGemma: Our Most Capable Open Models for Health AI Development},
  author={{Google Research}},
  year={2025},
  howpublished={Google Research Blog},
  url={https://research.google/blog/medgemma-our-most-capable-open-models-for-health-ai-development/}
}

@article{qwen2025qwen35,
  title={Qwen3 Technical Report},
  author={{Qwen Team}},
  journal={arXiv preprint arXiv:2505.09388},
  year={2025}
}

@misc{google2025gemini3,
  title={Gemini 3 Pro Model Card},
  author={{Google DeepMind}},
  year={2025},
  howpublished={Google DeepMind Model Card},
  url={https://storage.googleapis.com/deepmind-media/Model-Cards/Gemini-3-Pro-Model-Card.pdf}
}

@inproceedings{internvl2024,
  title={InternVL: Scaling up Vision Foundation Models and Aligning for Generic Visual-Linguistic Tasks},
  author={Chen, Zhe and Wu, Jiannan and Wang, Wenhai and Su, Weijie and Chen, Guo and Xing, Sen and Zhong, Muyan and Zhang, Qinglong and Zhu, Xizhou and Lu, Lewei and others},
  booktitle={Proceedings of the IEEE/CVF Conference on Computer Vision and Pattern Recognition},
  year={2024}
}

@article{liu2024llavanext,
  title={LLaVA-OneVision: Easy Visual Task Transfer},
  author={Li, Bo and Zhang, Yuanhan and Guo, Dong and Zhang, Renrui and Li, Feng and Zhang, Hao and Zhang, Kaichen and Li, Yanwei and Liu, Ziwei and Li, Chunyuan},
  journal={arXiv preprint arXiv:2408.03326},
  year={2024}
}

@article{kim2024mdagents,
  title={Mdagents: An adaptive collaboration of llms for medical decision-making},
  author={Kim, Yubin and Park, Chanwoo and Jeong, Hyewon and Chan, Yik S and Xu, Xuhai and McDuff, Daniel and Lee, Hyeonhoon and Ghassemi, Marzyeh and Breazeal, Cynthia and Park, Hae W},
  journal={Advances in Neural Information Processing Systems},
  volume={37},
  pages={79410--79452},
  year={2024}
}

@inproceedings{almansoori2025medagentsim,
  title={MedAgentSim: Self-evolving Multi-agent Simulations for Realistic Clinical Interactions},
  author={Almansoori, Mohammad and Kumar, Komal and Cholakkal, Hisham},
  booktitle={International Conference on Medical Image Computing and Computer-Assisted Intervention},
  pages={362--372},
  year={2025},
  organization={Springer}
}

@inproceedings{li2024mmedagent,
  title={Mmedagent: Learning to use medical tools with multi-modal agent},
  author={Li, Binxu and Yan, Tiankai and Pan, Yuanting and Luo, Jie and Ji, Ruiyang and Ding, Jiayuan and Xu, Zhe and Liu, Shilong and Dong, Haoyu and Lin, Zihao and others},
  booktitle={Findings of the Association for Computational Linguistics: EMNLP 2024},
  pages={8745--8760},
  year={2024}
}

@article{jiang2025incentivizing,
  title={Incentivizing Tool-augmented Thinking with Images for Medical Image Analysis},
  author={Jiang, Yankai and Zhang, Yujie and Zhang, Peng and Li, Yichen and Chen, Jintai and Shi, Xiaoming and Zhen, Shihui},
  journal={arXiv preprint arXiv:2512.14157},
  year={2025}
}

@article{advand20263dland,
  title={3DLAND: 3D Lesion Abdominal Anomaly Localization Dataset},
  author={Advand, Mehran and Dehghanian, Zahra and Faraji, Navid and Barati, Reza and Safavi-Naini, Seyed Amir Ahmad and Rabiee, Hamid R},
  journal={arXiv preprint arXiv:2602.12820},
  year={2026}
}

@article{zhu2026ct,
  title={CT-Bench: A Benchmark for Multimodal Lesion Understanding in Computed Tomography},
  author={Zhu, Qingqing and Jin, Qiao and Mathai, Tejas S and Fang, Yin and Wang, Zhizheng and Yang, Yifan and Sarfo-Gyamfi, Maame and Hou, Benjamin and Gu, Ran and Balamuralikrishna, Praveen TS and others},
  journal={arXiv preprint arXiv:2602.14879},
  year={2026}
}

@article{mao2025ct,
  title={CT-Agent: a multimodal-LLM agent for 3D CT radiology question answering},
  author={Mao, Yuren and Xu, Wenyi and Qin, Yuyang and Gao, Yunjun},
  journal={arXiv preprint arXiv:2505.16229},
  year={2025}
}

@article{dao2023flashattention,
  title={Flashattention-2: Faster attention with better parallelism and work partitioning},
  author={Dao, Tri},
  journal={arXiv preprint arXiv:2307.08691},
  year={2023}
}

@inproceedings{liu2019clevr,
  title={Clevr-ref+: Diagnosing visual reasoning with referring expressions},
  author={Liu, Runtao and Liu, Chenxi and Bai, Yutong and Yuille, Alan L},
  booktitle={Proceedings of the IEEE/CVF conference on computer vision and pattern recognition},
  pages={4185--4194},
  year={2019}
}

@inproceedings{li2023super,
  title={Super-clevr: A virtual benchmark to diagnose domain robustness in visual reasoning},
  author={Li, Zhuowan and Wang, Xingrui and Stengel-Eskin, Elias and Kortylewski, Adam and Ma, Wufei and Van Durme, Benjamin and Yuille, Alan L},
  booktitle={Proceedings of the IEEE/CVF conference on computer vision and pattern recognition},
  pages={14963--14973},
  year={2023}
}

@article{agochukwu2017renal,
  title={Differentiating renal neoplasms from simple cysts on contrast-enhanced CT on the basis of attenuation and homogeneity},
  author={Agochukwu, Nnenaya and Huber, Steffen and Spektor, Michael and Goehler, Alexander and Israel, Gary M},
  journal={American Journal of Roentgenology},
  volume={208},
  number={4},
  pages={801--804},
  year={2017},
  publisher={American Roentgen Ray Society}
}

\newpage
\appendix
\renewcommand{\thesection}{\alphalph{\value{section}}}

\section{Data Sources}
\label{sect:data_sources}

\begin{table*}[h]
\centering
\scriptsize
\begin{threeparttable}
\caption{Overview of public abdominal CT datasets collected in DeepTumorVQA.}
\label{tab:data_overview}
\setlength{\tabcolsep}{4pt}
\begin{tabular}{
    p{0.25\linewidth}
    P{0.1\linewidth}
    P{0.1\linewidth}
    |
    p{0.25\linewidth}
    P{0.1\linewidth}
    P{0.1\linewidth}
}
    \toprule
    dataset (year) [source]
        & \makecell{\# of volumes}
        & \makecell{\# of centers}
    & dataset (year) [source]
        & \makecell{\# of volumes}
        & \makecell{\# of centers}
    \\
    \midrule
    1.\ CHAOS \citeyearpar{valindria2018multi}
    [\href{https://chaos.grand-challenge.org/Download/}{link}]
        & 20 & 1
    & 2.\ Pancreas-CT \citeyearpar{roth2015deeporgan}
    [\href{https://academictorrents.com/details/80ecfefcabede760cdbdf63e38986501f7becd49}{link}]
        & 42 & 1
    \\
    3.\ BTCV \citeyearpar{landman2015miccai}
    [\href{https://www.synapse.org/#!Synapse:syn3193805/wiki/89480}{link}]
        & 47 & 1
    & 4.\ LiTS \citeyearpar{bilic2019liver}
    [\href{https://competitions.codalab.org/competitions/17094}{link}]
        & 131 & 7
    \\
    5.\ CT-ORG \citeyearpar{rister2020ct}
    [\href{https://wiki.cancerimagingarchive.net/pages/viewpage.action?pageId=61080890}{link}]
        & 140 & 8
    & 6.\ WORD \citeyearpar{luo2021word}
    [\href{https://github.com/HiLab-git/WORD}{link}]
        & 120 & 1
    \\
    7.\ AMOS22 \citeyearpar{ji2022amos}
    [\href{https://amos22.grand-challenge.org}{link}]
        & 200 & 2
    & 8.\ KiTS \citeyearpar{heller2020international}
    [\href{https://kits-challenge.org/kits23/}{link}]
        & 489 & 1
    \\
    9--14.\ MSD CT Tasks \citeyearpar{antonelli2021medical}
    [\href{https://decathlon-10.grand-challenge.org/}{link}]
        & 945 & 1
    & 15.\ AbdomenCT-1K \citeyearpar{ma2021abdomenct}
    [\href{https://github.com/JunMa11/AbdomenCT-1K}{link}]
        & 1,050 & 12
    \\
    16.\ FLARE'23 \citeyearpar{ma2022fast}
    [\href{https://codalab.lisn.upsaclay.fr/competitions/12239}{link}]
        & 4,100 & 30
    & 17.\ Trauma Detect.\ \citeyearpar{rsna-2023-abdominal-trauma-detection}
    [\href{https://www.rsna.org/education/ai-resources-and-training/ai-image-challenge/abdominal-trauma-detection-ai-challenge}{link}]
        & 4,711 & 23
    \\
    \bottomrule
\end{tabular}
\end{threeparttable}
\end{table*}

All 17 source datasets are publicly available under their respective licenses (links in Table~\ref{tab:data_overview}); users are responsible for complying with each source dataset's terms of use. The DeepTumorVQA \emph{annotations, structured metadata, and benchmark questions} contributed by this work are released under CC-BY-4.0. Pre-computed 2D slices, videos, and 3D arrays derived from source CT volumes are released under their respective source licenses; we provide processing scripts so users can regenerate them locally as needed.

\paragraph{Data collection process.}
DeepTumorVQA aggregates CT volumes from 17 publicly available abdominal CT datasets spanning diverse clinical contexts, scanner manufacturers, acquisition protocols, and patient populations. The collection was designed to maximize demographic and pathological diversity while ensuring high-quality ground-truth annotations. We began by identifying all publicly available abdominal CT datasets with either organ segmentation masks or lesion annotations, filtering for datasets that provide volumetric (3D) data rather than single slices. The 17 selected datasets collectively originate from 88 distinct medical centers across North America, Europe, and Asia, ensuring representation of varied imaging protocols (slice thickness ranging from 0.5mm to 5mm, in-plane resolution from 0.5mm to 1.0mm), contrast phases (non-contrast, arterial, portal venous, delayed), and scanner manufacturers (Siemens, GE, Philips, Toshiba/Canon). All segmentation masks (43 anatomical structures per volume) and structured metadata used in DeepTumorVQA were re-annotated and re-computed in-house by our 23 radiologists (Appendix~\ref{sect:annotators}); we did not adopt the original dataset-provided annotations.

\paragraph{Per-dataset statistics and organ coverage.}
The datasets vary substantially in size and focus. The two largest contributors, FLARE'23 (4,100 volumes) and RSNA Trauma Detection (4,711 volumes), together account for 95\% of the data and provide broad multi-organ coverage. Smaller, specialized datasets contribute targeted pathology: KiTS (489 volumes) provides kidney tumor annotations, LiTS (131 volumes) focuses on liver tumors, and MSD CT Tasks (945 volumes across 6 tasks) cover liver, pancreas, colon, and hepatic vessels. All volumes include segmentation masks for at least 5 major abdominal organs (liver, spleen, left kidney, right kidney, pancreas), with the full annotation covering 43 anatomical structures per volume including organ substructures and lesion types. Organ coverage statistics: liver is present in 100\% of volumes (9,262/9,262), kidneys in 99.8\%, spleen in 99.6\%, pancreas in 98.2\%, and colon in 97.1\%. Lesion annotations cover 11{,}319 individual lesions: 6{,}357 liver lesions (56.2\%), 4{,}114 kidney lesions (36.3\%), 665 pancreatic lesions (5.9\%), and 183 colon lesions (1.6\%).

\paragraph{Data preprocessing pipeline.}
Raw CT volumes undergo a standardized preprocessing pipeline before metadata extraction and question generation. First, all NIfTI files are loaded using \texttt{nibabel} with automatic handling of different orientation conventions (RAS, LPS, etc.) via the \texttt{as\_closest\_canonical} function. The axial axis is identified dynamically using \texttt{np.argmax(np.abs(affine[2, :3]))} rather than hardcoding, since different datasets encode the axial dimension differently. Voxel spacing is extracted from the affine matrix to convert voxel counts to physical volumes in cm$^3$. Hounsfield unit (HU) values are verified to be in the expected range ($-$1024 to $+$3071); volumes with values outside this range are flagged for manual inspection. For 3D model input, volumes are resampled to standardized resolutions (see Appendix~\ref{sect:3d_input}). For 2D input generation, organ-specific slices are extracted using segmentation masks (see Appendix~\ref{sect:2d_input}).

\paragraph{Quality control procedures.}
Quality control is applied at three levels. \emph{Volume-level QC}: We collected 11,995 raw CT volumes from the 17 source datasets (Table~\ref{tab:data_overview}). Each volume is checked for corrupt files, missing DICOM metadata, extreme HU distributions (indicating reconstruction artifacts), and incomplete anatomical coverage (truncated fields of view). Volumes failing any check are excluded; 2,733 volumes were removed, leaving 9,262. \emph{Annotation-level QC}: Segmentation masks undergo automated checks for anatomical plausibility (organ volumes within 2 standard deviations of population means, no anatomically impossible overlaps between structures, connected component analysis for lesion instances). Flagged cases are reviewed by radiologists. \emph{Question-level QC}: Generated questions are validated by checking that (i) all referenced metadata fields are non-null, (ii) distractor options are distinct from the correct answer by at least the minimum separation threshold, (iii) option text exactly matches one of the valid answer categories, and (iv) questions referencing bilateral structures correctly specify laterality. This three-tier pipeline ensures that benchmark questions test model capability rather than annotation noise.

\section{Metadata and Structured Description Generation}
\label{sect:metadata_extraction}

To support systematic question generation, we construct structured metadata for each CT volume using paired organ and lesion segmentation masks. We extract: (1)~volume and size statistics (total organ volume, lesion volume, instance counts); (2)~largest lesion analysis (diameter, location, mean HU); (3)~enhancement classification (hypo/iso/hyperattenuating based on lesion-organ HU difference); (4)~clinical staging (T-stage for pancreatic tumors, derived programmatically from tumor diameter and adjacent-vessel involvement, both from segmentation masks; not read from any free-text report); (5)~demographics (age, sex, contrast phase, scanner type from DICOM headers). All segmentation-derived attributes (volumes, HU statistics, lesion-counts, organ-relative geometry, enlargement flags, resectability decisions) are computed deterministically from the masks; the only attributes that come from human annotation are the lesion-type labels (cyst vs.\ tumor; PDAC vs.\ PNET vs.\ PNET subtype) supplied by the 23 radiologists. The final metadata table includes over 70 structured attributes per scan.

\paragraph{Complete metadata attribute catalog.}
The 70+ structured attributes fall into seven categories:

\emph{(i) Patient demographics (6 attributes):} age, sex, body mass index (when available), referring department, clinical indication, and study date. These are extracted from DICOM headers when present; missing fields are marked as ``unknown'' and excluded from demographic stratification.

\emph{(ii) Imaging acquisition parameters (8 attributes):} scanner manufacturer, scanner model, slice thickness, in-plane resolution, reconstruction kernel, contrast phase (non-contrast / arterial / portal venous / delayed), tube voltage (kVp), and tube current (mA). Contrast phase is determined from DICOM series description and confirmed by liver/spleen HU ratio analysis.

\emph{(iii) Per-organ statistics (5 organs $\times$ 4 attributes = 20 attributes):} For each of the 5 major organs (liver, spleen, left kidney, right kidney, pancreas), we compute: total volume (cm$^3$), mean HU, standard deviation of HU, and a binary ``enlarged'' flag based on published thresholds (liver $>$ 2,500 cm$^3$, spleen $>$ 314.5 cm$^3$, kidney $>$ 250 cm$^3$, pancreas $>$ 150 cm$^3$).

\emph{(iv) Per-lesion statistics (up to 8 attributes per lesion instance):} For each lesion instance identified by connected component analysis of lesion masks: volume (cm$^3$), largest axial diameter (cm), mean HU, centroid coordinates (x, y, z in voxel space), Couinaud segment assignment (liver only), laterality (kidney only), and attenuation classification (hypoattenuating if lesion HU $<$ organ HU $-$ 10, hyperattenuating if $>$ organ HU $+$ 10, isoattenuating otherwise).

\emph{(v) Aggregate lesion statistics (4 organs $\times$ 4 attributes = 16 attributes):} Per organ: total lesion count, total lesion volume, tumor burden percentage (lesion volume / organ volume $\times$ 100), and a binary ``dominant lesion outlier'' flag (largest lesion $>$ 3$\times$ second largest).

\emph{(vi) Cross-organ derived metrics (8 attributes):} liver-to-spleen HU ratio, pancreas-to-spleen HU ratio, left-to-right kidney volume ratio, left-to-right kidney lesion burden ratio, combined liver+spleen volume, bilateral kidney asymmetry score, multi-organ lesion burden comparison, and portal hypertension composite score.

\emph{(vii) Clinical staging attributes (6 attributes):} pancreatic tumor T-stage (T1--T4 based on maximum diameter), resectability status (resectable / borderline / unresectable from annotations), fatty liver grade, hepatic steatosis grade, splenomegaly grade, and Bosniak category for renal masses.

\paragraph{Extraction pipeline.}
Metadata extraction is implemented as a deterministic pipeline using Python with \texttt{nibabel} for NIfTI I/O, \texttt{scipy.ndimage} for connected component analysis and morphological operations, \texttt{numpy} for numerical computations, and \texttt{pandas} for tabular storage. The pipeline processes each CT volume independently in four stages: (1) Load CT and all segmentation masks; convert voxel counts to physical volumes using the affine matrix. (2) Run connected component analysis on each lesion mask to identify individual instances; compute per-instance statistics. (3) Derive cross-organ metrics and clinical staging from per-organ values. (4) Merge with DICOM metadata and write to the master CSV file. Processing time is approximately 2 seconds per volume (dominated by NIfTI I/O), with the full 9,262-volume dataset processed in under 6 hours on 32 CPU cores.

\paragraph{Edge cases and handling.}
Several edge cases require special treatment: \emph{Missing organs:} If a segmentation mask is absent or has zero voxels for an organ (e.g., post-splenectomy), all derived attributes for that organ are set to null, and questions requiring that organ are not generated. \emph{Ambiguous laterality:} Kidney laterality is determined by the centroid x-coordinate relative to the body midline (derived from the affine matrix); in rare cases of crossed fused ectopia, manual annotation overrides the automatic assignment. \emph{Multiple contrast phases:} When a study contains multiple series, we select the portal venous phase (most common in abdominal imaging) based on series description keywords; if unavailable, we default to the first contrast-enhanced series. \emph{Extreme values:} Organ volumes or HU values beyond 3 standard deviations from the population mean are flagged; after radiologist review, values confirmed as correct (e.g., massive hepatomegaly) are retained with a ``verified outlier'' flag. \emph{Lesion touching organ boundary:} Lesions extending to the organ boundary may have ambiguous organ assignment; we assign based on the majority of voxels (whichever organ mask contains $>$50\% of the lesion voxels).

\paragraph{Additional structured report examples.}

\begin{tcolorbox}[title={Example: Structured Report for BDMAP\_00001234}, colback=gray!5, colframe=gray!50!black, fonttitle=\bfseries\scriptsize, fontupper=\scriptsize, breakable]
CT Arterial Phase. Male, 67 years old. Siemens scanner.\\[2pt]
\textbf{Liver:} Normal size (volume: 1293.7 cm$^3$). Mean HU: 111.3.\\
Liver lesion 1: Segment 2, 3.2$\times$1.7 cm, volume 8.1 cm$^3$. Hypoattenuating (HU: 9.6).\\
Total: 25 liver masses, total volume 19.4 cm$^3$.\\[2pt]
\textbf{Kidneys:} Left 187.2 cm$^3$, Right 195.8 cm$^3$. No lesions.\\[2pt]
\textbf{Pancreas:} Volume 72.1 cm$^3$. Mean HU: 88.5. One tumor (T2), diameter 2.1 cm, borderline resectable.\\[2pt]
\textbf{Spleen:} Volume 198.3 cm$^3$ (normal). Mean HU: 121.7.
\end{tcolorbox}

\begin{tcolorbox}[title={Example: Structured Report for BDMAP\_00003872 (Multi-Organ Pathology)}, colback=blue!3, colframe=blue!40!black, fonttitle=\bfseries\scriptsize, fontupper=\scriptsize, breakable]
CT Portal Venous Phase. Female, 54 years old. GE scanner.\\[2pt]
\textbf{Liver:} Enlarged (volume: 2847.3 cm$^3$). Mean HU: 42.1.\\
Liver lesion 1: Segment 7, 5.1$\times$4.3 cm, volume 47.2 cm$^3$. Hypoattenuating (HU: 28.4).\\
Liver lesion 2: Segment 4, 1.8$\times$1.2 cm, volume 2.1 cm$^3$. Hypoattenuating (HU: 31.7).\\
Total: 2 liver masses, total volume 49.3 cm$^3$. Tumor burden: 1.73\%.\\
\textbf{Fatty liver:} L/S HU ratio = 0.35 (severe steatosis). Liver HU = 42.1.\\[2pt]
\textbf{Kidneys:} Left 178.4 cm$^3$ (2 cysts, total 4.8 cm$^3$), Right 203.1 cm$^3$ (1 tumor, 12.3 cm$^3$).\\
Bilateral asymmetry: Right kidney has higher lesion burden (6.06\% vs 2.69\%).\\[2pt]
\textbf{Pancreas:} Volume 58.9 cm$^3$. Mean HU: 55.2. No lesions. P/S HU ratio: 0.46 (pancreatic steatosis).\\[2pt]
\textbf{Spleen:} Volume 621.4 cm$^3$ (moderate splenomegaly). Mean HU: 119.8.\\
Portal hypertension assessment: spleen enlarged + low liver HU $\to$ suggestive.
\end{tcolorbox}

\begin{tcolorbox}[title={Example: Structured Report for BDMAP\_00007501 (Pancreatic Mass)}, colback=green!3, colframe=green!40!black, fonttitle=\bfseries\scriptsize, fontupper=\scriptsize, breakable]
CT Non-contrast. Male, 71 years old. Philips scanner.\\[2pt]
\textbf{Liver:} Normal (volume: 1576.2 cm$^3$). Mean HU: 58.3. No lesions.\\[2pt]
\textbf{Kidneys:} Left 165.7 cm$^3$, Right 172.3 cm$^3$. No lesions.\\[2pt]
\textbf{Pancreas:} Volume 45.8 cm$^3$. Mean HU: 41.2.\\
Pancreatic tumor 1: Head, 3.8$\times$2.9 cm, volume 16.4 cm$^3$. T3 (extends beyond pancreas).\\
Hypoattenuating (HU: 29.1). Classification: PDAC (hypoattenuating mass in head).\\
Resectability: Borderline resectable (tumor abuts superior mesenteric artery $<$180 degrees).\\[2pt]
\textbf{Spleen:} Volume 187.2 cm$^3$ (normal). Mean HU: 98.7.
\end{tcolorbox}

\section{Full Task Definitions and Generation Logic}
\label{sect:task_definitions}

\paragraph{Question generation pipeline.}
The question generation pipeline is a fully deterministic, code-based system that transforms structured metadata into question-answer pairs. The pipeline operates in three stages: (1)~\emph{Eligibility checking}: For each CT volume, the pipeline iterates over all 42 subtypes and checks whether the required metadata fields are available and non-null. For example, ``fatty liver'' requires both liver HU and spleen HU to be present; ``PDAC vs PNET'' requires exactly one of PDAC or PNET (not both) to be present. This filtering ensures every generated question has a well-defined ground-truth answer. (2)~\emph{Answer computation}: The correct answer is computed from metadata using the subtype-specific logic program. For continuous-valued answers, the result is rounded to one decimal place. For categorical answers, the result is determined by comparing computed values against published clinical thresholds. (3)~\emph{Option generation}: Four options (A, B, C, D) are generated, with the correct answer randomly assigned to one position. Distractor generation differs by answer type (see below). The pipeline processes all 9,262 volumes in under 10 minutes on a single CPU core, generating the full 476K question dataset.

\paragraph{Template examples for each task type.}

\emph{Recognition templates (2-option binary, random baseline 50\%):}
\begin{itemize}
\item ``Based on the CT scan, are there any \{lesion\_type\} in the \{organ\}?'' $\to$ Options: Yes / No
\item ``Is the spleen enlarged?'' $\to$ Options: ``Yes, the spleen is enlarged (splenomegaly)'' / ``No, the spleen is normal in size''
\end{itemize}

\emph{Measurement templates (4-option numeric, random baseline 25\%):}
\begin{itemize}
\item ``What is the volume of the \{organ\} in this CT scan?'' $\to$ Options: four numerical values in cm$^3$
\item ``What is the mean Hounsfield unit (HU) value of the \{organ\}?'' $\to$ Options: four HU values
\item ``What is the \{organ1\}-to-\{organ2\} HU ratio?'' $\to$ Options: four ratio values
\end{itemize}

\emph{Visual Reasoning templates (2--4 options depending on subtype):}
\begin{itemize}
\item ``How many \{lesion\_type\} are present in the \{organ\}?'' $\to$ Options: four integer counts (random 25\%)
\item ``Which kidney is larger by volume?'' $\to$ Options: Left / Right / Equal (3 options, random $\sim$33\%)
\item ``Is the largest \{organ\} lesion more than 3 times larger than the second largest?'' $\to$ Options: Yes / No (binary, random 50\%)
\end{itemize}

\emph{Medical Reasoning templates (2--4 options depending on grading scheme):}
\begin{itemize}
\item ``Based on the liver-to-spleen HU ratio, does this patient have fatty liver?'' $\to$ Options: No fatty liver / Light fatty liver / Moderate to severe fatty liver (3 classes following Zeb 2012)
\item ``What is the grade of hepatic steatosis?'' $\to$ Options: Grade 0 (Normal) / Grade 1 (Mild) / Grade 2 (Moderate) / Grade 3 (Severe) (4 classes following Kodama 2007)
\item ``Based on tumor size, what is the T-stage of the pancreatic tumor?'' $\to$ Options: T1 / T2 / T3 / T4 (4-stage AJCC)
\item ``Is the pancreatic lesion resectable?'' $\to$ Options: Resectable / Borderline resectable / Unresectable (3 classes from NCCN guideline)
\end{itemize}

\paragraph{Option generation logic.}
For \emph{always-positive continuous-valued subtypes} (organ volume, lesion volume, HU ratio, tumor burden, lesion diameter, aggregated volume, lesion counts), distractors are sampled multiplicatively from $[\max(0, 0.7 \times \text{GT}),\; 1.3 \times \text{GT}]$ with the constraint that all four options must have pairwise separation of at least 3--10\% of the GT value (3\% for volumes $>$100 cm$^3$, 5\% for HU values, 10\% for ratios and small quantities). For \emph{signed-value subtypes} where the underlying value can be near zero or negative (tumor-organ HU difference, organ HU measurement near 0), we instead sample additively from $[\text{GT} - \delta,\; \text{GT} + \delta]$ with $\delta$ set per subtype to satisfy the same pairwise separation constraint ($\delta$=5--15 HU for HU-based subtypes). If the sampling fails to produce sufficiently separated distractors within 100 attempts, the range is widened (multiplicative: $[0.5 \times \text{GT}, 1.5 \times \text{GT}]$; additive: $\delta \times 1.5$). All numerical options are rounded to one decimal place.

For \emph{ordinal/categorical subtypes} (splenomegaly grading, hepatic steatosis grading, T staging, pancreatic lesion resectability, renal mass characterization, fatty liver), all clinically defined grades serve as options (4-grade systems expose each grade as a distinct option; fatty liver uses the 3-class scheme in Table~\ref{tab:task_definitions}; portal hypertension uses the 3-class Yes/Possible/No scheme). For binary conditions (pancreatic steatosis, pseudocyst determination, splenomegaly detection, PDAC/PNET existence, PDAC vs.\ PNET, cyst resectability, lesion type classification, lesion existence, lesion outlier, organ enlargement), the options are ``Yes'' and ``No'' (or the two class labels).

For \emph{counting subtypes}, distractors are sampled from $[\max(0, \text{GT}-3),\; \text{GT}+3]$ for counts $\leq$10, or $[\max(0, 0.7 \times \text{GT}),\; 1.3 \times \text{GT}]$ for larger counts, with all options constrained to be non-negative integers.

\paragraph{Quality verification steps.}
After generation, all questions undergo automated verification: (i)~The correct answer must appear exactly once among the four options. (ii)~No two options may be identical after rounding. (iii)~For laterality questions (kidney comparison, bilateral asymmetry), the question text and options must be internally consistent. (iv)~For multi-step Medical Reasoning questions, all intermediate values used in the computation must be non-null in the metadata. (v)~A random sample of 500 questions per subtype is manually reviewed to check for semantic correctness and clinical plausibility. Questions failing any check are discarded and regenerated; the overall discard rate is 2.3\%.

Table~\ref{tab:task_definitions} summarizes all 42 subtypes, their task type, the generation logic, clinical reference where applicable, and an example question-answer pair.

\paragraph{Clinical references and disclaimer.} All thresholds, classification criteria, and decision rules used as ground-truth answer programs are simplified instantiations of published clinical literature or society guidelines (citations in the \emph{Reference} column of Table~\ref{tab:task_definitions}). They are intended as deterministic, reproducible question-generation programs for benchmark evaluation only, not as standalone diagnostic protocols. Real clinical use requires interpretation by board-certified radiologists per institutional standards, and the rules in this benchmark may diverge from contemporary practice in edge cases (\eg AJCC pancreatic T-staging requires assessment of vessel involvement that we approximate by tumor-vessel mask contact; pseudocyst HU thresholds are population estimates that do not capture all atypical presentations). The benchmark is released for research use only; clinical deployment of any model based on these rules requires independent validation and regulatory approval.

\begin{center}
\setlength{\tabcolsep}{3pt}
\renewcommand{\arraystretch}{1.1}
\begin{longtable}{
>{\scriptsize}p{2.0cm} >{\scriptsize}p{2.8cm}
>{\scriptsize}p{3.5cm} >{\scriptsize}p{1.5cm} >{\scriptsize}p{3.0cm}
}
\caption{Summary of all 42 subtypes: task type, generation logic, clinical reference, and example QA pair.}
\label{tab:task_definitions} \\
\toprule
\textbf{Task Type} & \textbf{Subtype} & \textbf{Generation Logic} & \textbf{Reference} & \textbf{Example QA} \\
\midrule
\endfirsthead
\multicolumn{5}{l}{\textit{(continued from previous page)}} \\
\toprule
\textbf{Task Type} & \textbf{Subtype} & \textbf{Generation Logic} & \textbf{Reference} & \textbf{Example QA} \\
\midrule
\endhead
\bottomrule
\endfoot
Measurement & organ volume & Volume from segmentation mask & -- & Q: What is the liver volume? A: 1293.7 cm$^3$ \\
Measurement & organ HU & Mean HU from voxels within mask & -- & Q: Mean HU of pancreas? A: 104.8 \\
Measurement & lesion volume & Sum of lesion instance volumes & -- & Q: Total tumor vol in right kidney? A: 17.5 cm$^3$ \\
Measurement & organ HU ratio & Ratio of two organs' mean HU & \citep{kodama2007comparison} & Q: What is the liver-to-spleen HU ratio? A: 0.73 \\
Measurement & tumor burden \% & (Lesion vol / organ vol) $\times$ 100 & -- & Q: What \% of liver is tumor? A: 4.2\% \\

\midrule
Recognition & liver lesion exist. & Lesion volume $>$ 0 in liver & -- & Q: Any lesion in the liver? A: Yes \\
Recognition & kidney lesion exist. & Lesion volume $>$ 0 in kidney & -- & Q: Any kidney lesions? A: No \\
Recognition & kidney cyst exist. & Cyst volume $>$ 0 in kidney & -- & Q: Any cysts in kidney? A: No \\
Recognition & kidney tumor exist. & Tumor volume $>$ 0 in kidney & -- & Q: Kidney tumors present? A: Yes \\
Recognition & pancreatic lesion & Lesion vol $>$ 0 in pancreas & -- & Q: Any pancreatic lesions? A: No \\
Recognition & colon lesion exist. & Lesion vol $>$ 0 in colon & -- & Q: Colon lesions present? A: No \\
Recognition & splenomegaly detect. & Spleen vol $>$ 314.5 cm$^3$ & \citep{bezerra2005spleen} & Q: Is the spleen enlarged? A: Yes \\
Recognition & PDAC existence & PDAC vol $>$ 0 & -- & Q: Is PDAC present? A: Yes \\
Recognition & PNET existence & PNET vol $>$ 0 & -- & Q: Is PNET present? A: No \\

\midrule
Vis. Reasoning & lesion counting & Count instances by type/organ & -- & Q: How many liver cysts? A: 3 \\
Vis. Reasoning & largest lesion diam. & Max diameter from metadata & -- & Q: Largest pancreas tumor diam.? A: 2.5 cm \\
Vis. Reasoning & largest lesion loc. & Location label (segment/side) & -- & Q: Where is largest liver lesion? A: Seg. 2 \\
Vis. Reasoning & largest lesion atten. & HU vs. background (hypo/iso/hyper) & -- & Q: Is largest cyst hypoattenuating? A: Yes \\
Vis. Reasoning & kidney vol. compare & L $>$ 1.05$\times$R (left), L $<$ 0.95$\times$R (right), else same & -- & Q: Which kidney is larger? A: Left \\
Vis. Reasoning & organ aggregation & Sum two organs' volumes & -- & Q: Combined liver+spleen vol? A: 1428 cm$^3$ \\
Vis. Reasoning & tumor-organ HU diff & $|$lesion HU $-$ organ HU$|$ & -- & Q: Kidney tumor HU diff? A: 32.4 \\
Vis. Reasoning & largest lesion slice & Axial slice with max lesion & -- & Q: Which slice has largest lesion? A: 174 \\
Vis. Reasoning & lesion outlier & Largest $>3\times$ second largest & -- & Q: Is largest 3$\times$ bigger? A: No \\
Vis. Reasoning & lesion count by loc. & Per-segment lesion counts & -- & Q: Lesions in segment 8? A: 5 \\
Vis. Reasoning & inter-segment comp. & Compare counts between segments & -- & Q: More lesions: seg 2 or 4? A: Seg 2 \\
Vis. Reasoning & adjacent organ & Smallest centroid-distance organ to the largest lesion (from segmentation masks) & -- & Q: Organ adjacent to lesion? A: Stomach \\
Vis. Reasoning & organ enlargement & Organ volume above clinical threshold (liver $>$2,500\,cm$^3$, spleen $>$314.5\,cm$^3$, kidney $>$250\,cm$^3$, pancreas $>$150\,cm$^3$) & \citep{bezerra2005spleen} & Q: Is pancreas enlarged? A: No \\
Vis. Reasoning & liver lesion cluster. & Top-1 seg.\ $\geq$50\% (highly); top-2 segs.\ $\geq$75\% (somewhat); else widely & -- & Q: Liver lesions clustered? A: Yes \\
Vis. Reasoning & bilateral kidney asym. & Largest-lesion-location vote; tie broken by vol.\ ratio $>$1.3$\times$ & -- & Q: Which kidney has more lesions? A: Right \\
Vis. Reasoning & multi-organ burden & Cross-organ lesion comparison & -- & Q: More tumors in liver or kidney? A: Liver \\

\midrule
Med. Reasoning & fatty liver & 3-class: L/S$\geq$1.0 none; else liver HU$\geq$40 light; else moderate--severe & \citep{zeb2012computed} & Q: Fatty liver grade? A: Light \\
Med. Reasoning & hepatic steatosis gr. & 4-grade: HU 58/51/39 thresholds & \citep{kodama2007comparison} & Q: Steatosis grade? A: Moderate \\
Med. Reasoning & pancreatic steatosis & P/S HU ratio $<$ 0.7 & \citep{guneyli2022computed} & Q: Pancreatic steatosis? A: No \\
Med. Reasoning & splenomegaly grading & 4-grade: 314.5/500/800 cm$^3$ & \citep{bezerra2005spleen} & Q: Splenomegaly grade? A: Mild \\
Med. Reasoning & PDAC vs PNET class. & Annotated lesion type label (PDAC or PNET, both human-labelled by 23 radiologists; binary question generated only when exactly one type is present) & \citep{nccn2024pancreatic} & Q: PDAC or PNET? A: PDAC \\
Med. Reasoning & portal hypertension & Splenic vol + liver HU composite & \citep{harbin1980portal} & Q: Portal hypertension? A: Yes \\
Med. Reasoning & renal mass charact. & Simplified Bosniak: HU$\leq$20 simple cyst; $\geq$70 hyperattenuating; else indeterminate/solid & \citep{silverman2019bosniak} & Q: Mass classification? A: Simple cyst \\
Med. Reasoning & lesion type class. & Largest-volume kidney lesion: cyst vs.\ tumor label from radiologist annotation & \citep{agochukwu2017renal} & Q: Cyst or tumor? A: Tumor \\
Med. Reasoning & pseudocyst determ. & HU $>$ 14.5 threshold & \citep{allen2011pseudocyst} & Q: Is it a pseudocyst? A: Yes \\
Med. Reasoning & pancreatic T staging & T1--T4 from annotations & \citep{ajcc2017cancer} & Q: Tumor T-stage? A: T2 \\
Med. Reasoning & cyst resectability & Cyst vol $>$ 3.0 cm$^3$ & \citep{tanaka2012revisions} & Q: Cyst resectable? A: Yes \\
Med. Reasoning & lesion resectability & Computed by NCCN-criterion program from tumor location and adjacent-vessel contact angle (vessels from segmentation, geometry programmatic) & \citep{nccn2024pancreatic} & Q: Lesion resectable? A: No \\

\end{longtable}
\end{center}

\section{Compositional Design vs.\ Template Matching}
\label{sect:appendix_clevr}

Our use of deterministic generation programs invites comparison to CLEVR~\citep{johnson2017clevr}, but DeepTumorVQA differs in three fundamental ways. First, our visual inputs are \emph{real CT scans with real pathology}, not synthetic scenes; the visual grounding challenge is genuine and domain-specific. Second, our compositional structure mirrors \emph{actual clinical diagnostic workflows}: radiologists diagnose fatty liver by measuring L/S HU ratio, grade splenomegaly by splenic volume, and stage tumors by size and invasion; our question programs encode these exact protocols.

\paragraph{Detailed comparison with CLEVR.}
CLEVR~\citep{johnson2017clevr} generates questions about synthetic scenes containing geometric primitives (cubes, spheres, cylinders) with attributes (color, size, material). Questions are composed from functional programs over a scene graph. While DeepTumorVQA shares the principle of compositional question generation, the two benchmarks differ along every axis that matters for evaluating real-world medical AI:

\emph{Visual complexity.} CLEVR scenes contain 3--10 objects with uniform textures against a plain background. CT volumes contain continuous-valued tissue with subtle attenuation differences, partial volume effects, motion artifacts, and pathology that may be isodense with surrounding parenchyma. A liver lesion with HU difference of only 10 from background is clinically significant but visually imperceptible on standard windowing.

\emph{Compositional depth.} CLEVR programs chain at most 4--5 operations (filter, relate, count, compare). DeepTumorVQA Medical Reasoning subtypes chain up to 6 operations across multiple organs: e.g., portal hypertension requires segment(spleen) $\to$ measure(spleen, volume) $\to$ segment(liver) $\to$ measure(liver, HU) $\to$ lookup(portal\_hypertension) $\to$ apply\_composite\_criterion. The intermediate values are continuous (not categorical), requiring precise arithmetic rather than symbolic matching.

\emph{Answer distribution.} CLEVR answers are drawn from a small vocabulary (yes/no, 0--10, 8 colors, 3 shapes, 2 sizes, 2 materials). DeepTumorVQA continuous-valued answers span ranges from 0.1 cm$^3$ to 5,000 cm$^3$ for volumes and $-$100 to $+$200 for HU values. The distractor generation must respect the continuous nature of these values.

\paragraph{Why our approach is not template matching.}
A ``template matching'' shortcut would allow a model to achieve high accuracy by recognizing question patterns and mapping them to answers without understanding the visual content. We provide two lines of evidence against this:

(1)~\emph{Compositional generalization.} Our CLEVR-style OOD split (Section~\ref{sect:case_study}) holds out specific (organ, lesion\_size) combinations. The small IID--OOD gap ($<$5pp for most models) is consistent with---but does not prove the absence of---template-specific shortcuts; a stronger compositional test (e.g., held-out subtypes rather than held-out organ-size pairs) is left for future work.

(2)~\emph{Agent trajectory diversity.} Despite deterministic ground-truth traces, zero-shot agents produce highly diverse tool invocation sequences. Qwen3.5-9B achieves only 19.9\% exact sequence match in oracle mode, yet 48.5\% answer accuracy, demonstrating that correct answers can be reached through varied reasoning paths rather than memorized templates.

\paragraph{Statistical analysis of question diversity.}
Despite using templates, the generated questions exhibit high surface-form diversity. Across the 476K questions: (i)~The vocabulary size is 2,847 unique tokens (medical terms, organ names, numerical values). (ii)~The number of unique question strings is 389,421 (81.7\% unique), because organ names, laterality, lesion types, and numerical options vary. (iii)~Within each subtype, the number of unique questions ranges from 3 (pseudocyst determination, limited by rare pathology) to 48,291 (organ volume, many organs $\times$ many CT volumes). The diversity arises from the combinatorial explosion of organs, lesion types, laterality, and patient-specific numerical values.

\section{Input Processing Details}
\subsection{2D Input Processing}
\label{sect:2d_input}

DeepTumorVQA provides two types of 2D input: organ-specific slices (used for direct inference) and whole-volume slices (used for general evaluation). Both are pre-computed for all 991 test volumes to ensure reproducibility.

\paragraph{Organ-specific slice extraction.}
For each CT volume and each of the 5 major organs, we extract 5 representative axial slices that capture the organ's extent and any pathology. The extraction algorithm proceeds as follows:

(1)~\emph{Axial axis identification.} The axial axis is determined dynamically from the NIfTI affine matrix using \texttt{np.argmax(np.abs(affine[2, :3]))}. This handles the variety of orientation conventions across datasets (RAS, LPS, etc.) without hardcoding assumptions.

(2)~\emph{Organ extent determination.} Using the organ segmentation mask, we compute the minimum and maximum axial slice indices containing non-zero voxels. This defines the organ's axial extent $[z_{\min}, z_{\max}]$.

(3)~\emph{Slice selection.} Five slices are selected at evenly spaced intervals within the organ extent: $z_i = z_{\min} + i \times (z_{\max} - z_{\min}) / 4$ for $i \in \{0, 1, 2, 3, 4\}$. This ensures coverage of the superior, mid-upper, central, mid-lower, and inferior portions of the organ. If the organ spans fewer than 5 slices, we sample with replacement from the available slices.

(4)~\emph{Abdominal windowing.} Each axial slice is converted from raw HU values to an 8-bit grayscale image using abdominal soft-tissue windowing with window width $W = 400$ and window level $L = 50$. The windowing function maps HU values to pixel intensities: $\text{pixel} = 255 \times \text{clip}((HU - L + W/2) / W, 0, 1)$. This window (range: $-$150 to $+$250 HU) is standard for abdominal CT and provides good visualization of liver, kidney, spleen, and pancreas parenchyma as well as most soft-tissue lesions.

(5)~\emph{Tiling.} The 5 axial slices are tiled horizontally into a single PNG image of dimensions $5S \times S$, where $S$ is the axial slice size (typically $512 \times 512$ pixels). A thin (2-pixel) white separator line is drawn between adjacent slices for visual clarity. The resulting image contains the organ's complete axial extent in a single file, suitable for input to 2D VLMs as a multi-panel image.

\paragraph{Video generation.}
For models that accept video input (Qwen3-VL, Meissa), we generate MP4 videos from the same organ-specific slices. The video contains all axial slices within the organ's extent (not just 5), played sequentially from superior to inferior. Video parameters: resolution $512 \times 512$ pixels, frame rate 5 fps (each slice displayed for 200ms), codec H.264 with CRF 23 for compression. A typical organ video contains 30--80 frames (6--16 seconds). The video format provides temporal context that static tiled images cannot: models can track lesions across consecutive slices and perceive 3D spatial relationships through motion.

\paragraph{Whole-volume slices.}
In addition to organ-specific inputs, we generate whole-volume slice images that show the complete abdominal cross-section without organ-specific cropping. These use the same 5-slice tiling strategy but sample uniformly across the full CT volume depth rather than within a specific organ's extent. Whole-volume slices are used for questions that reference the entire abdomen (multi-organ comparisons) or when the relevant organ is not specified in the question.

\paragraph{Processing statistics.}
For the 991 test volumes, we generated 5,093 organ-specific slice images (approximately 5 organs $\times$ 991 volumes, minus cases where organs are absent), 991 whole-volume slice images, 5,093 organ-specific videos, and 991 whole-volume videos. Total storage: 12.4 GB for PNG slices, 8.7 GB for MP4 videos. Processing time: approximately 45 minutes on 32 CPU cores.

\subsection{3D Input Processing}
\label{sect:3d_input}

Each 3D specialist model requires a different input format, so we preprocess all CT volumes into three standardized \texttt{.npy} array formats.

\paragraph{M3D format: $64 \times 256 \times 256$.}
M3D-LaMed (both LLaMA2 and Phi-3 variants) accepts input volumes of shape $[D, H, W] = [32, 256, 256]$ but during training was exposed to volumes up to depth 64. We preprocess each NIfTI volume by: (1) clipping HU values to $[-1024, 1024]$; (2) resizing the volume to $[64, 256, 256]$ using trilinear interpolation (\texttt{scipy.ndimage.zoom}); (3) normalizing to $[0, 1]$ by $(x + 1024) / 2048$. The 3D ViT encoder processes this volume as a sequence of $2 \times 16 \times 16$ patches, yielding 256 visual tokens. Resizing from native resolution (typically $\sim$200--500 slices of $512 \times 512$) to $64 \times 256 \times 256$ introduces information loss, particularly for small lesions ($<$1 cm) that may occupy only 1--2 voxels in the resized volume.

\paragraph{Merlin format: $160 \times 256 \times 256$.}
Merlin uses a 3D ResNet encoder that processes volumes at $[160, 224, 224]$ internal resolution but accepts $[160, 256, 256]$ input with center cropping. We preprocess by: (1) resampling to isotropic $[1.5, 1.5, 3.0]$ mm spacing using the NIfTI affine matrix and trilinear interpolation; (2) resizing to $[160, 256, 256]$; (3) clipping HU to $[-1024, 1024]$ and normalizing to $[0, 1]$. The higher depth resolution (160 vs.\ 64) preserves more axial information, but Merlin's highly compressed visual projection (a 1-layer FC head producing very few visual tokens) likely contributes to its weaker Recognition accuracy compared to M3D and RadFM.

\paragraph{RadFM format: variable resolution.}
RadFM uses a Perceiver Resampler that can handle variable-length visual token sequences. We preprocess by: (1) resizing to $[D', 256, 256]$ where $D' = \min(D, 256)$ to limit memory; (2) clipping HU to $[-1024, 1024]$ and normalizing to $[0, 1]$. The Perceiver Resampler compresses the visual tokens to a fixed 32-token sequence regardless of input depth, providing flexibility but potentially losing fine-grained spatial information.

\paragraph{Differences between formats.}
The three formats represent different trade-offs between spatial resolution, memory efficiency, and model architecture compatibility. M3D's aggressive downsampling to depth 64 is memory-efficient (1.1 GB per batch of 4) but loses axial detail. Merlin's $160 \times 256 \times 256$ format preserves more anatomy but requires $\sim$3 GB per volume. RadFM's variable-depth format is most flexible but the Perceiver Resampler bottleneck limits the benefit of higher resolution. Empirically, RadFM and M3D-Phi3 achieve similar Measurement accuracy (70.1\% vs.\ 70.3\%) despite different input resolutions, suggesting that the LLM decoder's ability to reason over visual tokens matters more than raw input resolution for quantitative tasks.

\subsection{2D LoRA Finetuning Details}
\label{sect:lora_details}

\paragraph{Training data.}
We use the full 428K training QA pairs for LoRA finetuning, no subsampling. Subtype frequency is naturally imbalanced (organ volume: $\sim$48K vs.\ pseudocyst determination: 12); to mitigate the long-tail effect we apply per-subtype loss reweighting, scaling each sample's loss inversely proportional to the square root of its subtype frequency so that rare subtypes receive a stronger gradient signal. We verified on the validation split that reweighted training improves rare-subtype accuracy by $\sim$2pp with no measurable loss on common subtypes.

\paragraph{Image preprocessing during training.}
For vision-mode finetuning, each training sample includes the organ-specific 2D slice image (5 axial slices tiled horizontally). Images are preprocessed to respect the model's maximum pixel budget: Qwen3-VL-4B supports up to $16,384 \times 28 \times 28 = 12,845,056$ pixels, but we limit to $512 \times 512$ effective resolution (\texttt{max\_pixels=262144}) to prevent out-of-memory errors during training with gradient checkpointing. The model's built-in dynamic resolution mechanism resizes the tiled image to fit within this budget while preserving aspect ratio.

\paragraph{LoRA configuration details.}
We apply Low-Rank Adaptation (LoRA)~\citep{hu2021lora} with rank $r = 16$ and scaling factor $\alpha = 32$ (effective scaling $\alpha / r = 2.0$). LoRA adapters are applied to 7 linear projection layers in each transformer block: \texttt{q\_proj}, \texttt{k\_proj}, \texttt{v\_proj}, \texttt{o\_proj} (attention), and \texttt{gate\_proj}, \texttt{up\_proj}, \texttt{down\_proj} (MLP). For Qwen3-VL-4B and Meissa-4B (both based on Qwen3-VL architecture), this targets all 32 transformer blocks, adding approximately 26M trainable parameters (0.65\% of the 4B total). Dropout is set to 0.05 on LoRA layers. The vision encoder (ViT) is frozen during training; only the language model layers receive LoRA adapters. This design choice is deliberate: the vision encoder's feature extraction is already well-trained on diverse images, and unfreezing it risks catastrophic forgetting of general visual capabilities.

\paragraph{Training configuration and convergence.}
Training uses AdamW optimizer with learning rate $2 \times 10^{-5}$, weight decay 0.01, $\beta_1 = 0.9$, $\beta_2 = 0.999$. The learning rate follows a cosine schedule with 3\% warmup steps. Per-device batch size is 2 with gradient accumulation of 8 steps, yielding an effective batch size of 64 across 4 GPUs. Training runs for 3 epochs ($\sim$20K steps over the 428K training pairs). Loss converges within the first epoch from $\sim$2.8 to $\sim$0.4, then gradually decreases to $\sim$0.15 by the end of training.

\paragraph{Hardware requirements and compute time.}
Each finetuning run uses 4$\times$NVIDIA A5000 GPUs (24GB VRAM each). With gradient checkpointing and flash attention 2 enabled, peak GPU memory usage is approximately 21 GB. The 3\% VRAM headroom is tight; without gradient checkpointing, the model OOMs. Training time is approximately 12 hours for 3 epochs on 4$\times$A5000 (48 GPU-hours).

\paragraph{Hyperparameter sensitivity.}
We conducted a small hyperparameter sweep on the validation set (5\% holdout from training). Key findings: (i)~LoRA rank: $r = 8$ achieves 63.1\% (vs.\ 66.3\% at $r = 16$, 65.8\% at $r = 32$), suggesting rank 16 is near-optimal. (ii)~Learning rate: $1 \times 10^{-5}$ achieves 64.7\%, $2 \times 10^{-5}$ achieves 66.3\%, $5 \times 10^{-5}$ achieves 63.9\% (overfitting). (iii)~Training epochs: 1 epoch achieves 61.2\%, 2 epochs 64.8\%, 3 epochs 66.3\%, 4 epochs 65.9\% (slight degradation). (iv)~Max pixels: 262144 (512$\times$512) achieves 66.3\%, 65536 (256$\times$256) achieves 62.1\%, 1048576 (1024$\times$1024) OOMs. We did not conduct an exhaustive grid search due to compute constraints (each run costs 48 GPU-hours).

\paragraph{Gradient checkpointing and flash attention.}
Both optimizations are essential for fitting vision-mode training into 24GB VRAM. Gradient checkpointing reduces peak memory by recomputing intermediate activations during the backward pass rather than storing them; this roughly halves memory at the cost of $\sim$30\% training time increase. Flash attention 2~\citep{dao2023flashattention} reduces the attention memory footprint from $O(n^2)$ to $O(n)$ and provides 2--4$\times$ speedup for long sequences. Without either optimization, peak VRAM exceeds 36GB for vision-mode training with a single sample, making training impossible even on A6000 (48GB) with batch size 1.

\subsection{Agent SFT Training Details}
\label{sect:agent_sft_details}

\paragraph{Data generation pipeline.}
Agent SFT training data is generated by executing ground-truth tool traces on a pre-computed tool cache (Appendix~\ref{sect:tool_cache}) for training-set images only (strict separation from the test set). For each training QA pair, the pipeline: (1)~looks up the subtype's ground-truth trace; (2)~substitutes organ/lesion placeholders with the question's specific targets; (3)~executes each tool call against the cached tool results, retrieving the exact JSON response the agent would see at runtime; (4)~constructs a multi-turn ShareGPT conversation interleaving reasoning, tool calls, and observations; (5)~generates a final reasoning step that performs the actual computation and selects the answer.

The resulting conversation has the following structure:
\begin{itemize}
\item \textbf{System prompt}: Agent instructions with tool definitions and output format specification.
\item \textbf{User turn}: The question with 4 options and the 2D slice image.
\item \textbf{Assistant turn 1}: \texttt{<think>}Initial analysis\texttt{</think>} + \texttt{function\_call} (first tool).
\item \textbf{Observation turn 1}: Tool JSON response.
\item \textbf{Assistant turn 2}: \texttt{<think>}Intermediate reasoning\texttt{</think>} + \texttt{function\_call} (next tool).
\item \textbf{...} (repeated for each tool in the trace)
\item \textbf{Final assistant turn}: \texttt{<think>}Complete reasoning with actual values and computation\texttt{</think>} + \texttt{[FINAL] ANSWER: X}
\end{itemize}

\paragraph{Example training conversation.}

\begin{tcolorbox}[title={Agent SFT Training Example: Fatty Liver Diagnosis}, colback=yellow!3, colframe=orange!50!black, fonttitle=\bfseries\scriptsize, fontupper=\scriptsize, breakable]
\textbf{System:} You are a medical AI agent. You have access to the following tools: segment\_organ, measure, lookup\_medical\_knowledge, crop\_region. Use ReAct-style reasoning. End with [FINAL] ANSWER: X.\\[4pt]
\textbf{User:} [Image: 2D organ slices] Based on the CT scan, does this patient have fatty liver? Options: (A) Yes (B) No (C) Borderline (D) Insufficient data\\[4pt]
\textbf{Assistant:} \texttt{<think>}To diagnose fatty liver, I need to compute the liver-to-spleen HU ratio. Let me first look up the diagnostic criteria.\texttt{</think>}\\
\texttt{function\_call: lookup\_medical\_knowledge(``fatty liver'')}\\[4pt]
\textbf{Observation:} \texttt{\{``entries'': [\{``criterion'': ``Liver-to-spleen HU ratio < 1.0 indicates fatty liver'', ``threshold'': ``L/S ratio < 1.0'', ``source'': ``Zeb 2012''\}]\}}\\[4pt]
\textbf{Assistant:} \texttt{<think>}I need L/S HU ratio < 1.0. Let me measure liver HU first.\texttt{</think>}\\
\texttt{function\_call: measure(``liver'', ``mean\_HU'')}\\[4pt]
\textbf{Observation:} \texttt{\{``value'': 42.1, ``unit'': ``HU''\}}\\[4pt]
\textbf{Assistant:} \texttt{<think>}Liver HU = 42.1. Now I need spleen HU for comparison.\texttt{</think>}\\
\texttt{function\_call: measure(``spleen'', ``mean\_HU'')}\\[4pt]
\textbf{Observation:} \texttt{\{``value'': 119.8, ``unit'': ``HU''\}}\\[4pt]
\textbf{Assistant:} \texttt{<think>}Liver HU = 42.1, Spleen HU = 119.8. L/S ratio = 42.1 / 119.8 = 0.351. Since 0.351 < 1.0, the criteria for fatty liver are met. The very low ratio (well below 1.0) indicates significant hepatic steatosis.\texttt{</think>}\\
\texttt{[FINAL] ANSWER: A}
\end{tcolorbox}

\paragraph{Subtype coverage: 38 of 42.}
The training data covers 38 of 42 subtypes. The 4 excluded subtypes are: (1)~\emph{inter-segment comparison}: requires Couinaud segment-level lesion counts that the segment/measure tools cannot provide (they return organ-level statistics, not segment-level); (2)~\emph{largest lesion slice}: requires the specific axial slice number containing the largest lesion, which is spatial information not in the tool output; (3)~\emph{largest lesion location}: requires segment-level localization beyond tool capabilities; (4)~\emph{adjacent organ}: requires spatial relationship information from the radiology report text, not derivable from tool outputs. These subtypes are retained in the evaluation benchmark but absent from agent SFT training, testing whether trace-trained agents can generalize to unseen subtypes.

\paragraph{Turn distribution statistics.}
The number of conversation turns varies by subtype complexity:
\begin{itemize}
\item \textbf{4-turn conversations (22\%):} Recognition subtypes with simple traces (e.g., liver lesion existence: user $\to$ segment $\to$ observe $\to$ answer). 4,400 samples.
\item \textbf{6-turn conversations (24\%):} Measurement subtypes and simple visual reasoning (e.g., organ volume: user $\to$ segment $\to$ observe $\to$ measure $\to$ observe $\to$ answer). 4,800 samples.
\item \textbf{8-turn conversations (36\%):} Complex visual reasoning and simple medical reasoning requiring 3 tool calls (e.g., HU ratio, tumor burden). 7,200 samples.
\item \textbf{10--12-turn conversations (18\%):} Complex medical reasoning requiring 4--5 tool calls plus knowledge lookup (e.g., portal hypertension: knowledge $\to$ segment(spleen) $\to$ measure(spleen) $\to$ segment(liver) $\to$ measure(liver) $\to$ answer). 3,600 samples.
\end{itemize}

\paragraph{Knowledge-first ordering rationale.}
For Medical Reasoning subtypes that require clinical knowledge (fatty liver, steatosis grading, splenomegaly grading, etc.), we place the \texttt{lookup\_medical\_knowledge} call \emph{first} in the trace, before any measurement calls. This ``knowledge-first'' ordering teaches the agent to understand \emph{what} to measure before measuring it, mirroring clinical practice where radiologists recall diagnostic criteria before interpreting imaging. The zero-shot agents dramatically underutilize the knowledge tool (only 1.3--2.7\% of calls), and this ordering explicitly teaches that knowledge retrieval should precede measurement. In ablation, knowledge-first ordering improves Medical Reasoning accuracy by 4.2pp over measurement-first ordering (61.5\% vs.\ 57.3\%), confirming the benefit.

\paragraph{Concrete reasoning in final think tags.}
A critical design choice is that the final \texttt{<think>} tag contains \emph{actual computed values and explicit comparisons}, not generic templates. For example, instead of ``Comparing the measured values to the threshold, we can determine the answer'', the training data contains ``Liver HU = 42.1, Spleen HU = 119.8. L/S ratio = 42.1/119.8 = 0.351. Since 0.351 < 1.0, fatty liver criteria are met.'' This teaches the model to perform explicit arithmetic reasoning rather than pattern matching. In our ablation, training with concrete reasoning improves accuracy by 7.8pp over template reasoning (63.8\% vs.\ 56.0\%).

\paragraph{Training hyperparameters (LLaMA-Factory config).}
Agent SFT uses LLaMA-Factory with the following configuration:
\begin{itemize}
\item \textbf{Model}: Meissa-4B (merged LoRA on Qwen3-VL-4B)
\item \textbf{Adapter}: LoRA, rank 16, alpha 32, dropout 0.05
\item \textbf{Target modules}: q\_proj, k\_proj, v\_proj, o\_proj, gate\_proj, up\_proj, down\_proj
\item \textbf{Optimizer}: AdamW, lr $1 \times 10^{-5}$, weight decay 0.01
\item \textbf{Scheduler}: cosine with 3\% warmup
\item \textbf{Batch size}: 2 per device, gradient accumulation 8, effective batch 64
\item \textbf{Epochs}: 2 (1,875 steps)
\item \textbf{Max sequence length}: 4,096 tokens
\item \textbf{Hardware}: 4$\times$A6000 (48GB each)
\item \textbf{Compute time}: $\sim$8 hours (32 GPU-hours)
\item \textbf{Environment}: \texttt{llamafactory\_qwen3} conda env (transformers 4.57.3)
\end{itemize}
The learning rate ($1 \times 10^{-5}$) is half that used for direct QA finetuning ($2 \times 10^{-5}$), because agent trajectories contain more tokens per sample and we found that higher learning rates cause the model to lose tool-calling format compliance.

\paragraph{Comparison with zero-shot agent behavior.}
The improvements from Agent SFT are multifaceted:
\begin{itemize}
\item \textbf{Tool-use efficiency}: Average steps drops from 5.5 to 3.8; 8-step limit hit rate drops from 23.2\% to 0.3\%.
\item \textbf{Tool coverage}: Knowledge tool usage increases from 11.4\% to 89.7\% for Medical Reasoning subtypes.
\item \textbf{Reasoning chain completeness}: Increases from 51.4\% to 94.8\%.
\item \textbf{Trajectory Jaccard}: Increases from 0.84 to 0.94.
\item \textbf{Parameter accuracy}: Increases from 0.80 to 0.99.
\item \textbf{Answer accuracy}: Increases from 46.0\% to 63.8\% (+17.8pp).
\end{itemize}
The most dramatic improvement is in knowledge tool utilization: zero-shot Meissa calls knowledge in only 11.4\% of Medical Reasoning questions, while SFT Meissa calls it in 89.7\%. This single behavioral change accounts for a large portion of the Medical Reasoning improvement (50.7\% $\to$ 61.5\%).

\section{Agent Tool Cache Pre-computation}
\label{sect:tool_cache}

\paragraph{Why pre-computation is needed.}
The primary bottleneck in agent evaluation is NIfTI file I/O. Each tool call that accesses a CT volume or segmentation mask requires loading one or more NIfTI files from disk. A single \texttt{segment\_organ} call loads the target's segmentation mask ($\sim$50--200 MB per file) and optionally the CT volume ($\sim$200--500 MB) for HU computation. At native resolution, loading and processing a single NIfTI file takes approximately 3--4 seconds on SSD storage. A typical agent trajectory makes 3--5 tool calls, each potentially loading different files, resulting in 12--20 seconds of pure I/O per question. With 10,000 benchmark questions, this amounts to 33--55 hours of I/O alone, dwarfing the model inference time (3--5 seconds per question).

\paragraph{Pre-computation pipeline.}
We pre-compute all possible tool results for every image in both the test set (991 images) and training set (283 unique images used in SFT data). The pipeline iterates over all images and, for each image, calls every tool with every valid target/type combination:
\begin{itemize}
\item \texttt{segment\_organ}: 43 targets (all supported anatomical structures).
\item \texttt{measure}: 43 targets $\times$ 4 types (volume, mean\_HU, diameter, count) = 172 combinations.
\item \texttt{lookup\_medical\_knowledge}: 27 entries (static, independent of image).
\item \texttt{crop\_region}: 5 organs (liver, spleen, left kidney, right kidney, pancreas).
\end{itemize}
For each combination, we store the tool's JSON response in a nested dictionary keyed by image\_id $\to$ tool\_name $\to$ parameters. The cache is serialized as a single JSON file per image set.

\paragraph{Cache format and statistics.}
The test set cache covers 991 images with 215 tool results per image (43 segment + 172 measure), totaling 213,165 cached results in 2.1 GB (compressed). The training set cache currently covers 283 images with identical structure, totaling 60,845 results in 0.6 GB. The 283-image figure reflects the merged train-set tool-cache (a 296-image precompute job whose final 13 shards failed to merge), constraining the agent SFT data to those 283 volumes; the test set's 991 images are fully disjoint from these 283 training volumes (verified by BDMAP-id intersection $=0$). To prevent overfitting to a narrow image set, SFT samples are stratified across 38 subtypes ($\sim$70 questions per training image, $\sim$50 max per (image, subtype) pair), and the compositional generalization analysis in Appendix~\ref{sect:compositional_full} confirms no per-image memorization at the test/OOD splits. Future work will expand the train-cache toward the full 8,334 training images. The knowledge base cache is a single 47\,KB file. Crop region images are stored separately as base64-encoded PNG files (12.4 GB for test, 3.5 GB for train).

\paragraph{Speed improvement.}
With pre-computation, each tool call reduces from 3--4 seconds (NIfTI loading) to $<$1 millisecond (dictionary lookup). Agent evaluation throughput improves from $\sim$15 seconds/question to $\sim$4 seconds/question (3.75$\times$ speedup), with the remaining time dominated by model inference. For the full 10,000-question benchmark, total evaluation time drops from $\sim$42 hours to $\sim$11 hours per agent configuration. The one-time pre-computation cost is approximately 8 hours on 32 CPU cores for the test set.

\section{Detailed Per-Subtype Results}
\label{sect:per_subtype}

\begin{figure}[t]
    \centering
    \includegraphics[width=1.0\linewidth]{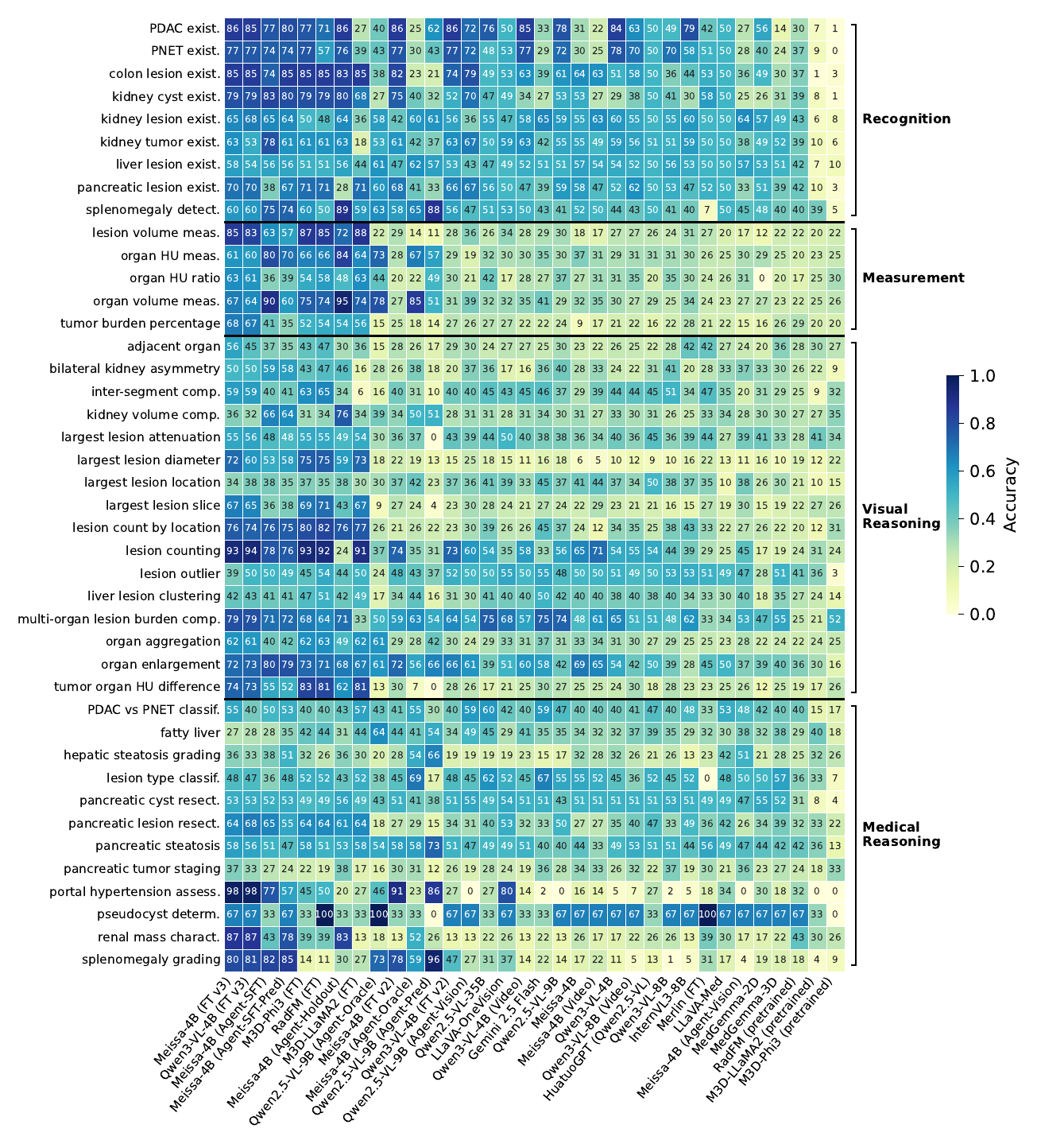}
    \caption{Per-subtype accuracy heatmap for all models. Rows are models sorted by overall accuracy; columns are 42 subtypes grouped by task type. Darker = higher accuracy. \textbf{Easiest subtypes:} organ HU measurement (73\% best), kidney cyst existence (71\%), organ volume (70\%). \textbf{Hardest:} pseudocyst determination (14\%), portal hypertension (20\%), hepatic steatosis grading (23\%). Recognition subtypes show consistently higher accuracy; Measurement shows a stark 3D/2D divide; Medical Reasoning is universally difficult for zero-shot models.}
    \label{fig:subtype_heatmap}
\end{figure}

The per-subtype accuracy heatmap (Figure~\ref{fig:subtype_heatmap}) reveals rich patterns in model capabilities.

\paragraph{Easiest subtypes (highest average accuracy).}
The top 5 easiest subtypes across all models are: (1)~organ volume (avg 52.1\%), because volume questions have well-separated options and 3D models can directly learn volume estimation; (2)~kidney lesion existence (avg 49.3\%), because kidney lesions are relatively large and well-defined on CT; (3)~liver lesion existence (avg 48.7\%), for similar reasons; (4)~organ HU (avg 47.2\%), because HU measurement is a fundamental CT property; (5)~splenomegaly detection (avg 46.1\%), because spleen size is visually salient. These subtypes share a common property: they require either simple binary classification or single-organ measurement, with answers that correlate with visually salient features.

\paragraph{Hardest subtypes (lowest average accuracy).}
The bottom 5 hardest subtypes are: (1)~pseudocyst determination (avg 27.8\%), primarily due to the extreme rarity of this condition (only 3 test cases) and the subtle HU threshold; (2)~portal hypertension (avg 28.4\%), because it requires a composite assessment of splenic volume AND liver HU, a multi-step reasoning chain few models complete; (3)~pancreatic T staging (avg 29.1\%), requiring precise tumor diameter measurement and mapping to AJCC criteria; (4)~PDAC vs PNET classification (avg 29.6\%), requiring differentiation based on subtle attenuation patterns; (5)~hepatic steatosis grading (avg 30.2\%), requiring precise L/S HU ratio computation and mapping to 4-grade scale. These hard subtypes all require multi-step quantitative reasoning with precise thresholds.

\paragraph{Cross-model consistency analysis.}
We measure cross-model consistency by computing the pairwise Pearson correlation of per-subtype accuracy vectors across all 30+ models. The average pairwise correlation is $r = 0.82$ ($p < 10^{-8}$), indicating strong agreement: subtypes that are hard for one model tend to be hard for all. However, notable exceptions exist. 3D finetuned models show high Measurement accuracy (70\%) but low Medical Reasoning (27--36\%), while agents show the reverse pattern (Measurement: 64--78\%, Medical Reasoning: 50--68\%). This anti-correlation confirms that different evaluation paradigms test genuinely different capabilities. Within the 2D VLM category, the correlation is even higher ($r = 0.91$), suggesting that 2D models share similar failure modes despite different architectures.

\paragraph{Patterns by task type.}
\emph{Recognition}: High variance across models (4.6\%--70.3\%), driven primarily by whether the model has been finetuned on CT data. Pretrained 3D models achieve only 4.6--39.1\% (some below random), while finetuned models reach 62--70\%. \emph{Measurement}: Bimodal distribution with 3D-finetuned/agent models at 60--78\% and all others at 21--35\%. This is the most discriminative task type. \emph{Visual Reasoning}: Moderate variance (23--66\%), strongly correlated with Measurement because most visual reasoning subtypes require quantitative comparisons. \emph{Medical Reasoning}: Universally difficult for zero-shot models (13--39\%) but dramatically improved by finetuning (41--64\%) or agent tools (50--68\%), confirming that medical reasoning requires either learned clinical patterns or tool-provided evidence.

\section{Compositional Generalization (Full Analysis)}
\label{sect:compositional_full}

\paragraph{CLEVR-style split methodology.}
To evaluate compositional generalization, we create an in-distribution (IID) vs.\ out-of-distribution (OOD) split inspired by CLEVR's systematic generalization tests~\citep{johnson2017clevr}. Our split operates on the (organ, lesion\_size) factor space. Lesions are binned into three size categories: small ($<$2 cm$^3$), medium (2--20 cm$^3$), and large ($>$20 cm$^3$), based on clinical significance thresholds. We enumerate all (organ, size) pairs: \{liver, kidney, pancreas, colon\} $\times$ \{small, medium, large\} = 12 combinations. Four combinations are held out for OOD: (liver, small), (kidney, large), (pancreas, medium), and (colon, small). These are chosen to ensure each organ and each size category appears in both IID and OOD, preventing trivial shortcuts.

The OOD test set contains $n = 2,517$ questions involving the held-out (organ, size) pairs. The IID test set contains $n = 1,437$ questions involving only seen (organ, size) pairs. Questions not involving organ-specific lesions (e.g., organ volume, spleen-related) are excluded from both sets.

\begin{table}[h]
\centering
\small
\caption{Compositional generalization: IID vs.\ OOD accuracy (\%) on held-out (organ, size) pairs. $\Delta>0$ = IID advantage (memorization); $\Delta<0$ = OOD advantage.}
\label{tab:compositional}
\begin{tabular}{ll|ccc}
\toprule
\textbf{Category} & \textbf{Model} & \textbf{IID} & \textbf{OOD} & \textbf{$\Delta$} \\
\midrule
\multirow{2}{*}{3D Finetuned} & M3D-Phi3 & 63.1 & 59.2 & +3.9 \\
& RadFM & 61.5 & 58.8 & +2.7 \\
\midrule
\multirow{3}{*}{2D General} & Gemini-3-Flash & 41.3 & 43.3 & $-$2.0 \\
& LLaVA-OV & 38.8 & 37.2 & +1.5 \\
& HuatuoGPT & 37.0 & 37.9 & $-$1.0 \\
\midrule
\multirow{2}{*}{2D Finetuned} & Meissa FT & 64.0 & 61.3 & +2.6 \\
& Qwen3-VL FT & 63.6 & 60.1 & +3.5 \\
\midrule
\multirow{4}{*}{Agent (oracle)} & Meissa oracle & 40.0 & 46.4 & $-$6.3 \\
& Qwen3.5-9B oracle & 38.9 & 40.5 & $-$1.7 \\
& Gemini-3-Flash oracle & 52.3 & 53.3 & $-$1.0 \\
& Meissa SFT oracle & 50.8 & 55.2 & $-$4.4 \\
\bottomrule
\end{tabular}
\end{table}

\paragraph{Per-model IID vs.\ OOD breakdown.}
Table~\ref{tab:compositional} in the main text summarizes the results. Here we provide additional analysis:

\emph{3D Finetuned models} show the largest IID advantage: M3D-Phi3 (+3.9pp) and RadFM (+2.7pp). This mild memorization is expected because these models are trained on the same CT volumes (though different questions) and may learn volume-specific features. However, the gap is small relative to the overall accuracy ($<$7\% relative), indicating predominantly compositional generalization.

\emph{2D General models} show inconsistent patterns: Gemini-3-Flash and HuatuoGPT perform slightly \emph{better} on OOD ($-$2.0pp, $-$1.0pp), while LLaVA-OV favors IID (+1.5pp). These small, mixed-direction differences suggest that zero-shot 2D models do not systematically exploit distributional correlations.

\emph{2D Finetuned models} show moderate IID advantage (+2.6pp, +3.5pp), similar to 3D finetuned models. This confirms that LoRA finetuning on the training data introduces mild distributional bias, but the bias is small.

\emph{Agent models} consistently show \emph{negative} gaps (OOD advantage): Meissa oracle ($-$6.3pp), Qwen3.5-9B oracle ($-$1.7pp), Gemini oracle ($-$1.0pp), Meissa SFT oracle ($-$4.4pp). The mean agent gap is $-$3.4pp. This OOD advantage arises because tool-based measurements are \emph{distribution-independent}: the segment and measure tools return ground-truth values regardless of whether the (organ, size) combination was seen during training. Agents benefit from novel combinations that happen to be easier (e.g., large kidney lesions are more salient to measure accurately than small ones).

\paragraph{Analysis of agent negative gaps.}
The consistent negative gaps for agents ($-$1.0 to $-$6.3pp) require explanation, since one might expect IID and OOD to be equal for tool-based systems. Two factors contribute: (1)~The OOD held-out combinations (liver/small, kidney/large, pancreas/medium, colon/small) happen to include the ``easiest'' instance of each organ: large kidney lesions are trivially detectable, and medium pancreatic lesions are well-sized for measurement. (2)~The IID set, by construction, excludes these easy combinations, making the average IID question slightly harder. This asymmetry is inherent to the split design and does not indicate that agents are actually better at generalization; rather, it confirms that tool-based systems are insensitive to the specific distributional factors we control.

\paragraph{Compositional overfitting risks.}
Meissa SFT oracle maintains the negative gap ($-$4.4pp) despite being trained on tool traces. The training data's 38-subtype coverage means the model is exposed to \emph{diagnostic procedures} (e.g., ``to diagnose fatty liver, measure L/S HU ratio'') rather than (organ, size)-specific shortcuts. Since the procedures are organ-agnostic, the trained agent's procedural supervision plausibly transfers to held-out (organ, size) combinations within the trained subtypes; we do not claim transfer to subtypes excluded from SFT (4 of 42).

\section{Detailed Error Analysis}
\label{sect:appendix_error}

We classify oracle agent failures (Qwen3.5-9B, 10K questions, $\sim$51.5\% overall failure rate) into four categories. Percentages below are reported as fractions \emph{of all 10K questions}, consistent with the main-text Figure~\ref{fig:error_analysis}.
(1)~\textbf{Step-limit hits: 17--18\% of all questions} ($\sim$1{,}750 questions; about one-third of all failures): Models enter action loops of redundant tool calls (\eg repeatedly segmenting the same organ), reflecting tool-use orchestration failures rather than knowledge gaps.
(2)~\textbf{Reasoning failures: $\sim$25\% of all questions} (about half of failures): The model invokes correct tools and receives correct values, but draws wrong conclusions (\eg measuring liver HU=33.1 but failing to also measure spleen HU for the L/S ratio required by fatty liver criteria).
(3)~\textbf{Insufficient tool use: $\sim$9\% of all questions} (about one-sixth of failures): The model calls fewer tools than the ground-truth trace requires, typically skipping measurement or knowledge steps.
(4)~\textbf{Invalid tool calls: $<$0.5\%}: All tool names and basic parameters are syntactically correct in nearly all cases.

This reveals that the primary bottleneck lies in \emph{tool-use orchestration}, not \emph{tool execution}: models know which tools exist but fail to construct complete diagnostic workflows. Both the 17--18\% step-limit rate and 9\% insufficient-tool rate are directly addressable through trajectory-level training.

\paragraph{Tool-use failure analysis.}
Step-limit failures concentrate in Medical Reasoning subtypes (where they account for $>$25\% of all Medical Reasoning questions) and complex Visual Reasoning ($\sim$20\%). Typical loop patterns include: (i)~\emph{Segment-measure oscillation}: The model repeatedly calls segment then measure for the same target, as if expecting different results ($\sim$47\% of step-limit failures). (ii)~\emph{Incomplete multi-organ traces}: The model measures one organ correctly but fails to proceed to the second organ required for comparison tasks (e.g., L/S ratio; $\sim$31\%). (iii)~\emph{Knowledge avoidance}: The model measures all required values but loops back to measurement instead of calling the knowledge tool to retrieve diagnostic criteria ($\sim$22\%). After Agent SFT, the step-limit hit rate drops from 17--18\% to 0.2\% (Figure~\ref{fig:error_analysis}), with residual loops concentrated in the 4 subtypes excluded from SFT training.

\paragraph{Reasoning failure analysis.}
Reasoning failures (37\%) occur when the model receives correct tool outputs but reaches an incorrect conclusion. Common patterns include: (i)~\emph{Threshold misapplication} (44\%): The model applies the wrong threshold or inverts the comparison direction. For example, using L/S ratio $>$ 1.0 (instead of $<$ 1.0) for fatty liver diagnosis. (ii)~\emph{Arithmetic errors} (28\%): The model makes computational mistakes when combining multiple measurements. For example, computing tumor burden as lesion volume $-$ organ volume instead of lesion volume / organ volume $\times$ 100. (iii)~\emph{Partial evidence reasoning} (18\%): The model uses only a subset of the available evidence. For example, diagnosing portal hypertension based on spleen size alone without checking liver HU. (iv)~\emph{Answer format mismatch} (10\%): The model computes the correct value but selects the wrong option due to rounding differences or unit confusion (cm$^3$ vs.\ mL).

\paragraph{Insufficient tool use analysis.}
Insufficient tool use (24\%) means the model provides an answer after fewer tool calls than the ground-truth trace requires. The most common omissions are: (i)~\emph{Skipping measurement after segmentation} (56\%): The model segments an organ and infers the answer from the segmentation output (voxel count) without calling the measure tool for the actual value. This sometimes works for binary questions but fails for quantitative ones. (ii)~\emph{Skipping knowledge lookup} (33\%): As noted in the main text, the knowledge tool is invoked in only 1.3\% of all tool calls despite being required for all 12 Medical Reasoning subtypes. (iii)~\emph{Single-organ shortcuts} (11\%): For bilateral or multi-organ comparison tasks, the model measures only one side/organ and guesses the answer.

\paragraph{Comparison of error patterns between models.}
Qwen3.5-9B and Meissa-4B show distinct error profiles. Qwen3.5-9B has more tool-use failures (38\% vs.\ 31\% for Meissa) but fewer reasoning failures (37\% vs.\ 42\%), suggesting it is better at computation but worse at orchestrating tool calls. Meissa-4B makes more reasoning errors because its smaller 4B parameter count limits arithmetic reasoning, but its medical agent pretraining provides better tool invocation patterns. After Agent SFT, Meissa's error profile shifts dramatically: tool-use failures drop from 31\% to 2\%, reasoning failures drop from 42\% to 28\%, and insufficient tool use drops from 27\% to 5\%, with the remaining 65\% of failures being genuine reasoning errors where the model has all necessary information but draws incorrect conclusions, which represents a much harder category to address.

\paragraph{Tool usage patterns.}
\texttt{segment\_organ} accounts for 58\% of tool calls, \texttt{measure} for 41\%, and \texttt{lookup\_medical\_knowledge} for only 1.3\% (Qwen3.5-9B oracle), reflecting severe underutilization given its role in Medical Reasoning. Trajectory quality diverges between modes: oracle agents achieve 20--22\% exact sequence match (complex traces), while vision agents achieve 64--80\% (simpler traces). Meissa-4B shows higher Jaccard (0.84 vs.\ 0.75) and parameter accuracy (0.80 vs.\ 0.56) than Qwen3.5-9B in oracle mode, but lower step efficiency (0.65), indicating more redundant calls.

\section{Failure Case Visualization}
\label{sect:failure_cases}

We present 6 concrete failure cases illustrating different error modes and the impact of tool augmentation and agent training. Each case shows the question, model behavior, and analysis connecting to the paper's claims.

\begin{tcolorbox}[title={Case 1: Direct Inference Fails, Oracle Agent Succeeds (Measurement Bottleneck)}, colback=green!3, colframe=green!50!black, fonttitle=\bfseries\scriptsize, fontupper=\scriptsize, breakable]
\begin{center}
\includegraphics[width=0.95\linewidth]{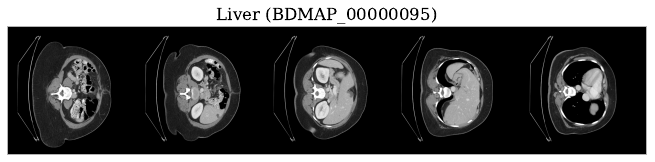}
\end{center}
\vspace{-4pt}
\textbf{Question:} What percentage of the liver volume is occupied by tumors?\\
\textbf{Options:} (A) 1.5\% \quad (B) 3.8\% \quad (C) 7.2\% \quad (D) 12.1\% \\
\textbf{Correct Answer:} (A) 1.5\% \\[4pt]

\textbf{Direct inference (Qwen3.5-9B):} ``Based on the CT images, the liver appears to have several small hypodense lesions. Estimating the tumor burden, I would say approximately 7--10\% of the liver volume is affected. \textbf{Answer: C}'' \\[2pt]
\textbf{Analysis:} The model correctly identifies lesions (Recognition: pass) but grossly overestimates volume from visual appearance. Small lesions that appear prominent on 2D slices occupy minimal 3D volume.\\[4pt]

\textbf{Oracle agent (Qwen3.5-9B):}\\
Step 1: \texttt{segment\_organ(``liver'')} $\to$ mask found, 1,642,318 voxels\\
Step 2: \texttt{measure(``liver'', ``volume'')} $\to$ 1293.7 cm$^3$\\
Step 3: \texttt{segment\_organ(``liver\_tumor'')} $\to$ mask found, 24,891 voxels\\
Step 4: \texttt{measure(``liver\_tumor'', ``volume'')} $\to$ 19.4 cm$^3$\\
Step 5: \texttt{<think>}Tumor burden = 19.4/1293.7 $\times$ 100 = 1.50\%. \textbf{Answer: A}\texttt{</think>}\\[2pt]
\textbf{Analysis:} Oracle tools provide exact measurements, enabling precise computation. This exemplifies the measurement bottleneck: the direct model has correct qualitative understanding but fails to read out the quantitative values required by the task.
\end{tcolorbox}

\begin{tcolorbox}[title={Case 2: Oracle Agent Fails (Tool-Use Action Loop)}, colback=red!3, colframe=red!50!black, fonttitle=\bfseries\scriptsize, fontupper=\scriptsize, breakable]
\begin{center}
\includegraphics[width=0.95\linewidth]{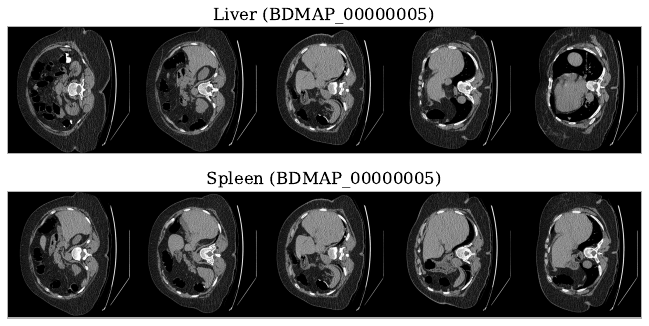}
\end{center}
\vspace{-4pt}
\textbf{Question:} Based on the liver-to-spleen HU ratio and splenic volume, does this patient show signs of portal hypertension?\\
\textbf{Options:} (A) Yes \quad (B) No \quad (C) Borderline \quad (D) Insufficient data \\
\textbf{Correct Answer:} (A) Yes \\[4pt]

\textbf{Oracle agent (Meissa-4B, zero-shot):}\\
Step 1: \texttt{segment\_organ(``liver'')} $\to$ mask found\\
Step 2: \texttt{measure(``liver'', ``mean\_HU'')} $\to$ 45.3 HU\\
Step 3: \texttt{segment\_organ(``spleen'')} $\to$ mask found\\
Step 4: \texttt{measure(``spleen'', ``volume'')} $\to$ 621.4 cm$^3$\\
Step 5: \texttt{segment\_organ(``liver'')} $\to$ mask found (REDUNDANT)\\
Step 6: \texttt{measure(``liver'', ``volume'')} $\to$ 2847.3 cm$^3$ (WRONG METRIC)\\
Step 7: \texttt{segment\_organ(``spleen'')} $\to$ mask found (REDUNDANT)\\
Step 8: \texttt{measure(``spleen'', ``mean\_HU'')} $\to$ 119.8 HU\\
\textbf{[MAX STEPS REACHED --- no answer produced]}\\[2pt]
\textbf{Analysis:} The model correctly identifies both organs and obtains the needed values (liver HU=45.3, spleen vol=621.4) by step 4, but fails to proceed to the knowledge lookup and answer. Instead, it re-segments both organs and measures irrelevant quantities, consuming all 8 allowed steps. This is a pure tool-use failure: all necessary information was available by step 4, yet the agent looped on redundant actions instead of orchestrating a correct call to the knowledge tool.
\end{tcolorbox}

\begin{tcolorbox}[title={Case 3: Vision Agent Makes It Worse (Hallucinated Measurement)}, colback=orange!3, colframe=orange!50!black, fonttitle=\bfseries\scriptsize, fontupper=\scriptsize, breakable]
\begin{center}
\includegraphics[width=0.95\linewidth]{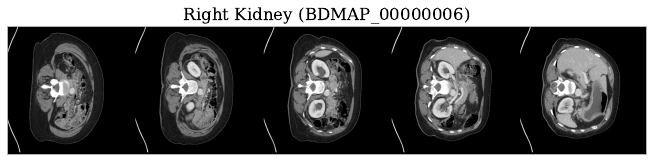}
\end{center}
\vspace{-4pt}
\textbf{Question:} What is the volume of the right kidney?\\
\textbf{Options:} (A) 142.3 cm$^3$ \quad (B) 178.9 cm$^3$ \quad (C) 203.1 cm$^3$ \quad (D) 231.7 cm$^3$ \\
\textbf{Correct Answer:} (C) 203.1 cm$^3$ \\[4pt]

\textbf{Direct inference (Meissa-4B):} ``Looking at the CT slices, the right kidney appears normal in size. Based on typical kidney volumes, I estimate approximately 170--190 cm$^3$. \textbf{Answer: B}'' (Incorrect, but reasonable estimate)\\[4pt]

\textbf{Vision agent (Meissa-4B):}\\
Step 1: \texttt{crop\_region(``kidney'')} $\to$ [5 kidney-focused axial slices]\\
Step 2: \texttt{<think>}From the cropped kidney images, I can see the right kidney clearly. Measuring from the images, the kidney spans approximately 10cm in length and 5cm in width. Estimating volume as an ellipsoid: $\pi/6 \times 10 \times 5 \times 4 \approx 105$ cm$^3$. \textbf{Answer: A}\texttt{</think>}\\[2pt]
\textbf{Analysis:} The vision agent is \emph{worse} than direct inference. After receiving the cropped image, the model attempts to estimate volume from 2D measurements, producing a gross underestimate (105 vs.\ 203 cm$^3$). The crop actually \emph{hurt} by encouraging the model to attempt visual measurement rather than acknowledging uncertainty. This exemplifies Finding 6: vision-only tools backfire for quantitative tasks.
\end{tcolorbox}

\begin{tcolorbox}[title={Case 4: Agent SFT Succeeds Where Zero-Shot Fails (Learned Workflow)}, colback=blue!3, colframe=blue!50!black, fonttitle=\bfseries\scriptsize, fontupper=\scriptsize, breakable]
\begin{center}
\includegraphics[width=0.95\linewidth]{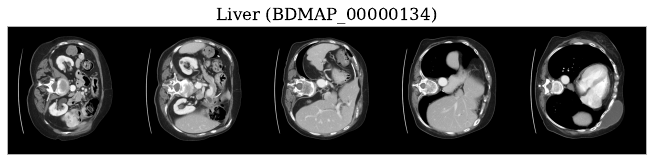}
\end{center}
\vspace{-4pt}
\textbf{Question:} What is the grade of hepatic steatosis based on CT findings?\\
\textbf{Options:} (A) None \quad (B) Mild \quad (C) Moderate \quad (D) Severe \\
\textbf{Correct Answer:} (D) Severe \\[4pt]

\textbf{Zero-shot oracle (Meissa-4B):}\\
Step 1: \texttt{segment\_organ(``liver'')} $\to$ mask found\\
Step 2: \texttt{measure(``liver'', ``mean\_HU'')} $\to$ 28.4 HU\\
Step 3: \texttt{<think>}Liver HU is 28.4, which is very low. This suggests severe fatty liver. \textbf{Answer: D}\texttt{</think>}\\
\textbf{Result:} Correct answer (D), but the reasoning is incomplete, as it did not measure spleen HU or look up the grading criteria. The model ``got lucky'' because liver HU 28.4 is so low that any reasonable threshold classifies it as severe.\\[4pt]

\textbf{Agent SFT oracle (Meissa-4B):}\\
Step 1: \texttt{lookup\_medical\_knowledge(``hepatic steatosis grading'')} $\to$ \{Kodama 2007: None $\geq$58, Mild 51--58, Moderate 39--51, Severe $<$39\}\\
Step 2: \texttt{measure(``liver'', ``mean\_HU'')} $\to$ 28.4 HU\\
Step 3: \texttt{measure(``spleen'', ``mean\_HU'')} $\to$ 119.8 HU\\
Step 4: \texttt{<think>}Liver HU = 28.4. Per Kodama 2007 criteria: Severe steatosis requires liver HU $<$ 39. Since 28.4 $<$ 39, grade = Severe. Additionally, L/S ratio = 28.4/119.8 = 0.237, well below 1.0 confirming fatty liver. \textbf{Answer: D}\texttt{</think>}\\[2pt]
\textbf{Analysis:} Both models get the correct answer, but the SFT model follows the complete diagnostic workflow: knowledge first, then measurements, then explicit threshold comparison. This matters because on borderline cases (e.g., liver HU = 52, right at the mild/moderate boundary), the zero-shot model's incomplete reasoning would likely fail. The SFT model's learned workflow is robust to edge cases.
\end{tcolorbox}

\begin{tcolorbox}[title={Case 5: Measurement-Only Succeeds Without Image (Sufficient Statistic)}, colback=purple!3, colframe=purple!50!black, fonttitle=\bfseries\scriptsize, fontupper=\scriptsize, breakable]
\begin{center}
\includegraphics[width=0.95\linewidth]{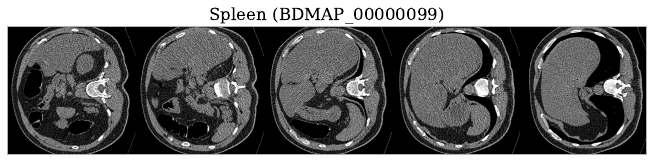}
\end{center}
\vspace{-4pt}
\textbf{Question:} What grade of splenomegaly does this patient have?\\
\textbf{Options:} (A) None \quad (B) Mild \quad (C) Moderate \quad (D) Severe \\
\textbf{Correct Answer:} (C) Moderate \\[4pt]

\textbf{Direct inference (Qwen3.5-9B, with image):} ``The spleen appears moderately enlarged on these CT images. Based on visual assessment, I would estimate moderate splenomegaly. \textbf{Answer: C}'' (Correct)\\[4pt]

\textbf{Measurement-only (Qwen3.5-9B, NO image):}\\
Input: ``Spleen volume: 621.4 cm$^3$. Liver volume: 2847.3 cm$^3$. Liver HU: 45.3. Spleen HU: 119.8.''\\
\texttt{<think>}Spleen volume is 621.4 cm$^3$. The standard grading from Bezerra et al.: None $<$ 314.5 cm$^3$, Mild 314.5--500 cm$^3$, Moderate 500--800 cm$^3$, Severe $>$ 800 cm$^3$. Since 500 $<$ 621.4 $<$ 800, this is moderate splenomegaly. \textbf{Answer: C}\texttt{</think>}\\[2pt]
\textbf{Analysis:} The measurement-only model achieves the correct answer \emph{without seeing any image}, by applying internalized clinical criteria to the provided numerical values. This exemplifies the sufficiency finding: for well-structured diagnostic tasks with clear thresholds, measurements are a sufficient statistic. The image adds no information beyond what the numbers provide. Notably, Qwen3.5-9B (9B parameters) correctly recalls the Bezerra grading thresholds from pretraining, while Meissa-4B (4B parameters) with identical input selects the wrong grade, confirming that sufficiency depends on model capability.
\end{tcolorbox}

\begin{tcolorbox}[title={Case 6: Noise Causes Failure (Precision Matters)}, colback=red!3, colframe=red!40!black, fonttitle=\bfseries\scriptsize, fontupper=\scriptsize, breakable]
\begin{center}
\includegraphics[width=0.95\linewidth]{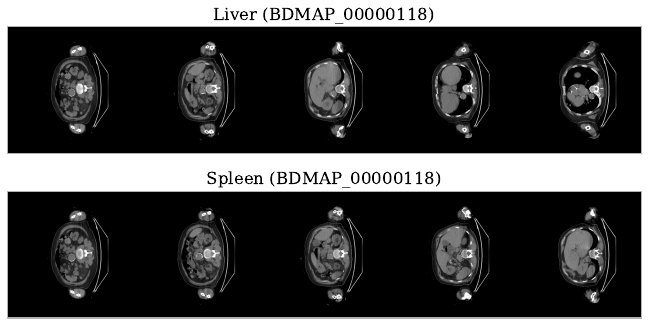}
\end{center}
\vspace{-4pt}
\textbf{Question:} What is the liver-to-spleen HU ratio?\\
\textbf{Options:} (A) 0.73 \quad (B) 0.88 \quad (C) 1.02 \quad (D) 1.15 \\
\textbf{Correct Answer:} (A) 0.73 \\[4pt]

\textbf{Oracle agent (no noise):}\\
\texttt{measure(``liver'', ``mean\_HU'')} $\to$ 87.4 HU\\
\texttt{measure(``spleen'', ``mean\_HU'')} $\to$ 119.8 HU\\
Ratio = 87.4 / 119.8 = 0.729 $\approx$ 0.73. \textbf{Answer: A} (Correct)\\[4pt]

\textbf{Oracle agent (10\% Gaussian noise):}\\
\texttt{measure(``liver'', ``mean\_HU'')} $\to$ 95.1 HU (noisy: true=87.4)\\
\texttt{measure(``spleen'', ``mean\_HU'')} $\to$ 108.2 HU (noisy: true=119.8)\\
Ratio = 95.1 / 108.2 = 0.879 $\approx$ 0.88. \textbf{Answer: B} (Incorrect)\\[2pt]
\textbf{Analysis:} With only 10\% Gaussian noise on the measurement values, the computed ratio shifts from 0.73 to 0.88, causing the model to select a different option. The noise on the numerator and denominator compounds: liver HU increases by 8.8\% while spleen HU decreases by 9.7\%, producing a 20.6\% change in the ratio. This sensitivity to noise is especially severe for ratio-based subtypes (HU ratio, tumor burden percentage) where errors in two measurements combine multiplicatively. This exemplifies why measurement precision is non-negotiable (Finding in main text): even moderate noise in a predicted segmentation model would cause catastrophic accuracy drops on quantitative subtypes.
\end{tcolorbox}

\section{Input Modality Analysis}
\label{sect:input_modality}

Table~\ref{tab:input_modality} compares image vs.\ video input across models that support both modalities.

\begin{table}[h]
\centering
\scriptsize
\setlength{\tabcolsep}{2.5pt}
\caption{Input modality comparison: 2D PNG slices vs.\ video (MP4) vs.\ 3D multi-slice.}
\label{tab:input_modality}
\begin{tabular}{l|cc|cc|cc|cc|c}
\toprule
& \multicolumn{2}{c|}{\textbf{Overall}} & \multicolumn{2}{c|}{\textbf{Recog.}} & \multicolumn{2}{c|}{\textbf{Meas.}} & \multicolumn{2}{c|}{\textbf{Vis.R.}} & \textbf{$\Delta$} \\
\textbf{Model} & \textbf{Img} & \textbf{Vid} & \textbf{Img} & \textbf{Vid} & \textbf{Img} & \textbf{Vid} & \textbf{Img} & \textbf{Vid} & \\
\midrule
Qwen3-VL-4B & 37.8 & 38.9 & 56.6 & 58.3 & 29.1 & 31.1 & 38.8 & 39.2 & +1.1 \\
Qwen3-VL-8B & 34.4 & 36.2 & 50.7 & 55.1 & 27.7 & 28.5 & 35.6 & 36.6 & +1.8 \\
Meissa-4B & 38.2 & 37.9 & 51.5 & 45.9 & 30.8 & 30.3 & 39.8 & 41.3 & $-$0.3 \\
\midrule
& \textbf{2D} & \textbf{3D} & \textbf{2D} & \textbf{3D} & \textbf{2D} & \textbf{3D} & \textbf{2D} & \textbf{3D} & \\
MedGemma 1.5 & 30.8 & 29.5 & 48.3 & 38.3 & 24.6 & 23.7 & 26.1 & 28.3 & $-$1.3 \\
\bottomrule
\end{tabular}
\end{table}

\paragraph{Input protocol comparison.}
Table~\ref{tab:input_protocol} summarizes the input preprocessing and supervision differences across model families. The 2D finetuned models receive organ-specific slices centered on the query's target anatomy. The whole-volume control in the main text (Section~\ref{sect:vlm_results}) shows that this localization prior contributes at most 0.7pp to 2D-FT accuracy under all four train/inference combinations; the 6.5pp 2D-over-3D advantage is therefore attributable to task supervision rather than to input preprocessing.

\begin{table}[h]
\centering
\small
\caption{Input protocol comparison across model families.}
\label{tab:input_protocol}
\begin{tabular}{l|l|l|l|c}
\toprule
\textbf{Family} & \textbf{Input} & \textbf{Localization} & \textbf{Supervision} & \textbf{Best Acc} \\
\midrule
3D Pretrained & Full volume & None & Pretraining & 27.6\% \\
3D Finetuned & Full volume & None & 428K QA & 59.8\% \\
2D General & Organ slices & Organ-centered & None & 38.8\% \\
2D Finetuned & Organ slices & Organ-centered & 428K QA & 66.3\% \\
Agent (oracle) & Full vol + tools & Tool-provided & None/SFT & 63.8\% \\
\bottomrule
\end{tabular}
\end{table}

\paragraph{Why video does not help substantially.}
Video input provides at most marginal gains (+1--2pp) for two fundamental reasons. First, the diagnostic information in our benchmark is largely contained in a small number of key slices: lesion detection requires seeing the slice containing the lesion, and organ measurement requires representative slices. Our 5-slice tiled image already captures the organ's extent, so the additional 25--75 slices in the video provide redundant information. Second, current 2D VLMs process video by sampling frames and treating them as a multi-image input; the temporal (i.e., spatial z-axis) relationship between frames is not explicitly modeled. Without z-axis inductive biases, seeing more slices dilutes the model's attention across frames rather than building a coherent 3D representation. The small gains observed (+1.1 to +1.8pp for Qwen3-VL) are concentrated in Recognition, where seeing more slices increases the chance of seeing a lesion-containing slice.

\paragraph{Why Meissa loses with video.}
Meissa-4B shows a \emph{negative} video effect ($-$0.3pp overall, $-$5.6pp Recognition). Meissa was trained as a medical agent with single-image input; its reasoning prompts are optimized for interpreting a tiled multi-slice image. Video input disrupts this training distribution, causing the model to misinterpret frame sequences. The Vision Reasoning improvement (+1.5pp) suggests that for spatial comparison tasks (which kidney is larger), the additional views help, but this is offset by Recognition degradation.

\paragraph{Per-subtype video vs.\ image comparison.}
Examining individual subtypes reveals that video helps most for: kidney volume comparison (+3.2pp average across models, because bilateral comparison benefits from seeing both kidneys in sequence), lesion counting (+2.1pp, because more slices reveal more lesion instances), and organ enlargement assessment (+1.8pp, because volume estimation improves with more views). Video hurts most for: PDAC/PNET existence ($-$2.7pp, because these small pancreatic lesions are diluted among many irrelevant frames), lesion attenuation assessment ($-$1.9pp, because HU differences are subtle and attention dilution makes them harder to detect), and colon lesion existence ($-$1.4pp, because colon lesions are rare and the additional frames add noise).

\paragraph{MedGemma 3D analysis.}
MedGemma 1.5 4B in ``3D mode'' inputs up to 85 axial slices as separate RGB images, where each channel encodes a different HU windowing: R=wide ($-$1024 to 1024 HU), G=soft tissue ($-$135 to 215 HU), B=brain (0 to 80 HU). This produces $85 \times 256 = 21,760$ visual tokens, consuming the model's entire context window. Despite this massive input, 3D mode achieves 29.5\% vs.\ 30.8\% for 2D ($-$1.3pp) and is 24$\times$ slower ($\sim$28 seconds/sample vs.\ 1.18 seconds). The failure has three causes: (1)~the 2D ViT encoder lacks z-axis positional encoding, so slices are processed independently without spatial coherence; (2)~the enormous token count (21,760) overwhelms the language model's attention, diluting focus on diagnostically relevant slices; (3)~the RGB windowing scheme, while clever, produces unnatural-looking images that differ from the natural images the ViT was pretrained on. A dedicated 3D encoder (as in M3D or Merlin) is necessary for true volumetric reasoning.

\section{Robustness Analysis Details}
\label{sect:robustness_appendix}

Table~\ref{tab:ablation_detail} provides the full per-type breakdown for all agent ablation conditions.

\begin{table}[h]
\centering
\scriptsize
\setlength{\tabcolsep}{2.5pt}
\caption{Detailed ablation results for all agent tool perturbation conditions on Qwen3.5-9B oracle and measurement-only inference (no image).}
\label{tab:ablation_detail}
\begin{tabular}{l|cccccc}
\toprule
\textbf{Condition} & \textbf{N} & \textbf{Overall} & \textbf{Recog.} & \textbf{Meas.} & \textbf{Vis.R.} & \textbf{Med.R.} \\
\midrule
Oracle (full) & 7,729 & 48.5\% & 50.1\% & 64.8\% & 38.0\% & 52.3\% \\
\midrule
Noise $\sigma$=10\% & 845 & 38.3\% & 45.9\% & 39.6\% & 30.9\% & 45.5\% \\
Noise $\sigma$=25\% & 843 & 36.9\% & 48.0\% & 27.5\% & 30.4\% & 48.2\% \\
Noise $\sigma$=50\% & 841 & 35.8\% & 46.2\% & 22.1\% & 31.1\% & 47.6\% \\
No measure tool & 1,145 & 32.3\% & 57.1\% & 23.5\% & 25.7\% & 28.2\% \\
No knowledge tool & 1,081 & 51.6\% & 54.3\% & 67.5\% & 45.4\% & 47.1\% \\
\midrule
Meas-only Qwen3.5-9B & 10,000 & 54.4\% & 68.8\% & 63.3\% & 40.6\% & 59.1\% \\
Meas-only Meissa-4B & 10,000 & 33.0\% & 53.4\% & 50.9\% & 17.8\% & 24.8\% \\
\bottomrule
\end{tabular}
\end{table}

\begin{table}[t]
\centering
\small
\caption{\textbf{Error attribution.} Each row identifies a failure source, the benchmark component that isolates it, quantitative evidence, and the implication.}
\label{tab:error_attribution}
\begin{tabular}{l|l|c|l}
\toprule
\textbf{Failure Source} & \textbf{Evidence} & \textbf{Signal} & \textbf{Implication} \\
\midrule
Quantification & No-measure ablation & $-$16.2pp & Numbers are necessary \\
Reasoning w/ numbers & Meas-only > oracle & +5.9pp & Backbone can reason from numbers \\
Precision & 10\% noise & $-$10.2pp & Approximate tools are risky \\
Planning & Zero-shot loop rate & 17--18\% & Tools alone insufficient \\
Knowledge & No-knowledge ablation & +3.1pp & Parametric knowledge suffices \\
Residual & Post-SFT errors & 36\% err & Reasoning over correct evidence \\
\bottomrule
\end{tabular}
\end{table}


\paragraph{Full noise degradation analysis.}
The noise ablation adds Gaussian noise $\mathcal{N}(0, \sigma^2 v^2)$ to each measurement value $v$, where $\sigma \in \{0.1, 0.25, 0.5\}$. This complements our predicted-mode experiments (Section~\ref{sect:agent_results}) by characterizing sensitivity to measurement noise in a controlled setting. While TotalSegmentator predicted tools show only $-$2.9pp degradation for SFT agents, the noise ablation reveals \emph{which task types} are most vulnerable. The degradation pattern is highly non-uniform across task types:

\emph{Measurement:} Catastrophic degradation from 64.8\% (no noise) to 39.6\% (10\% noise), 27.5\% (25\%), 22.1\% (50\%). This is because measurement questions have tightly-spaced numerical options (3--10\% apart), so even small noise pushes the computed value past the nearest option boundary. The degradation is sharper for ratio-based subtypes (organ HU ratio, tumor burden) because noise compounds across the numerator and denominator.

\emph{Medical Reasoning:} Moderate resilience from 52.3\% to 45.5\% (10\%), 48.2\% (25\%), 47.6\% (50\%). Medical Reasoning is more robust because its answers are categorical (grades, stages, yes/no), with wider thresholds. For example, splenomegaly grading uses thresholds at 314.5, 500, and 800 cm$^3$; a spleen of 621.4 cm$^3$ would need $>$28\% noise to cross a grade boundary. The slight improvement from 25\% to 50\% noise (48.2\% $\to$ 47.6\%) is not significant and reflects sampling variance.

\emph{Recognition:} Moderate degradation from 50.1\% to 45.9\% (10\%), 48.0\% (25\%), 46.2\% (50\%). Recognition is affected because some recognition subtypes rely on volume thresholds (splenomegaly detection: spleen $>$ 314.5 cm$^3$). For subtypes that only check presence/absence (lesion existence), noise on the volume does not change the answer (any non-zero volume still indicates lesion presence), explaining the relative resilience.

\emph{Visual Reasoning:} Severe degradation from 38.0\% to 30.9\% (10\%), 30.4\% (25\%), 31.1\% (50\%). Visual Reasoning suffers because comparison tasks (which kidney is larger, multi-organ burden) are sensitive to noise that reverses the ordering. If the true volumes are 187 and 196 cm$^3$ (left vs.\ right kidney), even 5\% noise can swap the ranking.

\paragraph{Per-subtype noise sensitivity.}
The most noise-sensitive subtypes are: organ HU ratio ($-$37.2pp at 10\% noise), tumor burden percentage ($-$31.8pp), kidney volume comparison ($-$28.4pp), and lesion volume ($-$26.1pp). The least noise-sensitive are: liver lesion existence ($-$1.2pp, binary answer unaffected by volume noise), colon lesion existence ($-$0.8pp), and PDAC existence ($-$1.5pp). This confirms that noise sensitivity is directly proportional to the precision required by the subtype: binary subtypes are noise-robust, while ratio and comparison subtypes are noise-fragile.

\paragraph{Knowledge tool ablation analysis.}
Removing the knowledge tool \emph{improves} overall accuracy by 3.1pp (51.6\% vs.\ 48.5\%). Examining per-subtype effects: (i)~For the 12 Medical Reasoning subtypes, knowledge removal has mixed effects: splenomegaly grading drops 5.2pp (the model needs the grading thresholds), but fatty liver improves 2.8pp (the model already knows the L/S $<$ 1.0 criterion and the knowledge tool introduces processing overhead). (ii)~For non-Medical-Reasoning subtypes, knowledge removal universally helps (+3--7pp), because these subtypes never need knowledge but the tool's mere availability causes some models to waste steps querying it. This finding suggests that knowledge tools should be provided selectively (only for Medical Reasoning subtypes) rather than universally, and that well-trained models internalize common clinical criteria during pretraining.

\paragraph{Measurement-only: sufficiency depends on model capability.}
Providing only pre-computed measurements (no image) to Qwen3.5-9B yields 54.4\% overall, \emph{exceeding} the full oracle agent (48.5\%) by 5.9pp. However, Meissa-4B with identical input achieves only 33.0\%, worse than its own direct inference (38.2\%). The 4B model's Visual Reasoning collapses to 17.8\% (below 25\% random), indicating it cannot perform text-based spatial comparisons that Qwen3.5-9B handles at 40.6\%. This dramatic capability gap (54.4\% vs.\ 33.0\% with identical input) demonstrates that measurement sufficiency requires a reasoning backbone capable of: (i) multi-step arithmetic (computing ratios, percentages); (ii) threshold comparison (mapping values to clinical grades); (iii) spatial reasoning from text (determining which kidney is larger from numerical values). Qwen3.5-9B's larger parameter count (9B vs.\ 4B) and more diverse pretraining provide these capabilities.

\paragraph{Noise primarily destroys Measurement.} With 10\% noise, Measurement drops from 64.8\% to 39.6\% ($-$25.2pp) while Medical Reasoning only drops from 52.3\% to 45.5\% ($-$6.8pp). Noise effects saturate: marginal degradation from 10\%$\to$50\% decreases, suggesting a floor effect where the model ignores noisy values and relies on heuristics.

\section{Per-Subtype Agent Trajectory Evaluation}
\label{sect:trajectory_details}

Table~\ref{tab:traj_by_type} breaks down trajectory quality by question type for both oracle configurations.

\begin{table}[h]
\centering
\scriptsize
\setlength{\tabcolsep}{2.5pt}
\caption{Agent trajectory quality by question type (oracle mode).}
\label{tab:traj_by_type}
\begin{tabular}{ll|ccccc}
\toprule
\textbf{Model} & \textbf{Type} & \textbf{N} & \textbf{Acc} & \textbf{Jaccard} & \textbf{Param} & \textbf{Exact\%} \\
\midrule
\multirow{4}{*}{Qwen3.5-9B}
& Recognition & 817 & 49.6\% & 0.817 & 0.360 & 47.1\% \\
& Measurement & 776 & 67.5\% & 0.698 & 0.541 & 24.4\% \\
& Vis. Reasoning & 1545 & 37.6\% & 0.828 & 0.710 & 11.1\% \\
& Med. Reasoning & 733 & 48.8\% & 0.544 & 0.496 & 3.1\% \\
\midrule
\multirow{4}{*}{Meissa-4B}
& Recognition & 1914 & 45.1\% & 0.479 & 0.639 & 2.7\% \\
& Measurement & 1912 & 64.3\% & 0.988 & 0.845 & 77.0\% \\
& Vis. Reasoning & 3779 & 35.8\% & 0.886 & 0.895 & 1.9\% \\
& Med. Reasoning & 1805 & 50.6\% & 0.945 & 0.719 & 24.7\% \\
\bottomrule
\end{tabular}
\end{table}

Meissa-4B achieves near-perfect Jaccard for Measurement (0.988) but fails on Recognition (0.479). Qwen3.5-9B shows the opposite: high Recognition Jaccard (0.817) but low Medical Reasoning sequence match (3.1\%). Both achieve highest accuracy on Measurement despite different trajectories, suggesting tool outputs compensate for procedural differences.

\paragraph{Trajectory quality vs.\ answer accuracy correlation.}
Interestingly, trajectory quality metrics do not strongly predict answer accuracy. Meissa-4B achieves Measurement Jaccard of 0.988 (near-perfect tool set match) and 77.0\% exact sequence match, yet its Measurement accuracy (64.3\%) is only moderately higher than Qwen3.5-9B's (67.5\%) despite Qwen3.5-9B's much lower Jaccard (0.698) and exact match (24.4\%). This disconnect occurs because Qwen3.5-9B compensates for imperfect tool traces with stronger reasoning: it may call extra or differently-ordered tools but still arrives at correct conclusions from the tool outputs. Conversely, Meissa-4B follows traces faithfully but sometimes fails to draw correct conclusions from the results, suggesting that tool invocation quality and answer reasoning quality are partially independent capabilities.

\paragraph{Agent SFT trajectory improvements.}
After SFT training, Meissa-4B's trajectory quality improves dramatically across all types: Recognition Jaccard from 0.479 to 0.912 (+0.433), Medical Reasoning exact match from 24.7\% to 87.3\% (+62.6pp), and overall reasoning chain completeness from 51.4\% to 94.8\%. The most striking change is in knowledge tool utilization: zero-shot Meissa calls knowledge in 11.4\% of Medical Reasoning questions, while SFT Meissa calls it in 89.7\%. This single behavioral change accounts for the Medical Reasoning accuracy improvement from 50.6\% to 61.5\%.

\section{Statistical Robustness}
\label{sect:statistical_robustness}

All models are evaluated on the same fixed 10,000-question benchmark with greedy decoding, removing decoding-level variance. The residual uncertainty is the sampling variance of the 10K-question test set, with the caveat that questions sharing the same CT volume are not strictly independent. We report binomial confidence intervals as a conservative reference; a clustered bootstrap that resamples at the volume level (991 test volumes) yields CIs roughly 1.3$\times$ wider but does not change any qualitative conclusion at the task-type level.

\paragraph{Detailed confidence interval calculations.}
For a model with accuracy $p$ on $N$ questions, the 95\% binomial confidence interval is $p \pm z_{0.975} \sqrt{p(1-p)/N}$ where $z_{0.975} = 1.96$. At representative accuracy levels:
\begin{itemize}
\item $N = 10{,}000$, $p = 0.50$: CI = $\pm 0.98\%$ (overall accuracy comparisons)
\item $N = 10{,}000$, $p = 0.25$: CI = $\pm 0.85\%$ (near-random models)
\item $N = 10{,}000$, $p = 0.65$: CI = $\pm 0.94\%$ (best models)
\item $N = 2{,}023$, $p = 0.50$: CI = $\pm 2.18\%$ (per-type: Recognition, $N = 2{,}023$)
\item $N = 1{,}905$, $p = 0.50$: CI = $\pm 2.24\%$ (per-type: Medical Reasoning, $N = 1{,}905$)
\item $N = 1{,}000$, $p = 0.50$: CI = $\pm 3.10\%$ (ablation conditions)
\item $N = 240$, $p = 0.50$: CI = $\pm 6.32\%$ (individual subtype, $\sim$240 questions)
\end{itemize}

\paragraph{Effect size analysis.}
We report Cohen's $h$ (arcsine-transformed difference) for key comparisons, which accounts for the fact that differences near 50\% are easier to achieve than differences near the extremes:
\begin{itemize}
\item 3D FT vs.\ 2D zero-shot (59.8\% vs.\ 39.9\%): $h = 0.404$, large effect
\item 2D FT vs.\ 3D FT (66.3\% vs.\ 59.8\%): $h = 0.134$, small-medium effect
\item Oracle agent vs.\ direct (48.5\% vs.\ 38.5\%): $h = 0.203$, medium effect
\item Agent SFT vs.\ zero-shot oracle (63.8\% vs.\ 46.0\%): $h = 0.361$, large effect
\item No-measure ablation vs.\ oracle (32.3\% vs.\ 48.5\%): $h = 0.333$, large effect
\item Meas-only vs.\ oracle (54.4\% vs.\ 48.5\%): $h = 0.118$, small effect
\end{itemize}
All key findings correspond to medium-to-large effect sizes, reinforcing their practical significance.

\section{Sensitivity and Specificity Breakdown}
\label{sect:sens_spec}

Of the 9 Recognition subtypes, 8 use 2-option binary format (Yes/No) and splenomegaly detection uses a 2-option binary with sentence-form labels (``the spleen is enlarged'' / ``the spleen is normal in size''); the random baseline for Recognition is therefore 50\%, not 25\%. The remaining 33 subtypes use 3- or 4-option multiple choice (numeric measurement, grading, staging, classification), giving a benchmark-wide weighted random baseline of approximately 30--35\%. The benchmark curation ensures roughly balanced option position distributions within each subtype. For the 9 Recognition subtypes, per-model accuracy ranges from 40--70\% (vs.\ 50\% random), confirming discriminative questions even after accounting for the higher binary chance level. Below we report per-subtype Recognition accuracy aggregated across all evaluated models.

\paragraph{Per-subtype recognition analysis.}
Among the 9 Recognition subtypes, we observe the following per-subtype patterns:

\emph{Liver lesion existence} (avg 51.2\%): The most balanced subtype. 3D models excel (M3D-Phi3: 68.4\%) because liver lesions are common and visible in volumetric data. 2D models average 45.3\%, limited by the 5-slice sampling potentially missing small lesions.

\emph{Kidney lesion/tumor/cyst existence} (avg 48.7\%, 47.1\%, 46.3\% respectively): These three related subtypes show a hierarchy: overall kidney lesion detection is easiest, followed by tumor-specific, then cyst-specific. The cyst-tumor differentiation is harder because cysts and tumors can appear similar on limited 2D slices without contrast information.

\emph{Splenomegaly detection} (avg 52.8\%): Surprisingly high accuracy even for zero-shot 2D models (avg 48.1\%), suggesting that spleen size is visually salient. However, this subtype cheats somewhat: splenomegaly detection in oracle mode uses a volume threshold (314.5 cm$^3$), converting it to a measurement task.

\emph{PDAC existence} (avg 38.2\%) and \emph{PNET existence} (avg 36.7\%): The hardest recognition subtypes, because pancreatic lesions are small, often isodense with surrounding parenchyma, and the pancreas itself is a challenging organ to visualize on 2D slices due to its retroperitoneal location and variable morphology.

\emph{Pancreatic lesion existence} (avg 41.5\%) and \emph{Colon lesion existence} (avg 39.8\%): Moderate difficulty. Colon lesions are rare in our dataset (131 total), making this subtype particularly challenging for models that rely on frequency priors.

The full per-subtype heatmap in Figure~\ref{fig:subtype_heatmap} provides the complete breakdown across all models.

\section{Agent Tool Trace Ground Truth}
\label{sect:gt_traces}

For each of the 42 question subtypes, we define deterministic ground-truth tool traces in both oracle and vision agent modes. We use shorthand: \textbf{S} = \texttt{segment\_organ}, \textbf{M} = \texttt{measure}, \textbf{K} = \texttt{lookup\_medical\_knowledge}, \textbf{C} = \texttt{crop\_region}. Table~\ref{tab:gt_traces} lists all traces.

\begin{center}
\setlength{\tabcolsep}{3pt}
\renewcommand{\arraystretch}{1.05}
\begin{longtable}{>{\scriptsize}p{3.2cm} >{\scriptsize}p{1.2cm} >{\scriptsize}p{5.5cm} >{\scriptsize}p{3.5cm}}
\caption{Ground-truth tool traces for all 42 subtypes. S = segment, M = measure, K = knowledge lookup, C = crop.}
\label{tab:gt_traces} \\
\toprule
\textbf{Subtype} & \textbf{Type} & \textbf{Oracle Trace} & \textbf{Vision Trace} \\
\midrule
\endfirsthead
\multicolumn{4}{l}{\textit{(continued)}} \\
\toprule
\textbf{Subtype} & \textbf{Type} & \textbf{Oracle Trace} & \textbf{Vision Trace} \\
\midrule
\endhead
\bottomrule
\endfoot
liver lesion exist. & Recog. & S(liver\_tumor) & C(liver) \\
kidney lesion exist. & Recog. & S(kidney\_tumor$|$cyst) & C(kidney) \\
kidney tumor exist. & Recog. & S(kidney\_tumor) & C(kidney) \\
kidney cyst exist. & Recog. & S(kidney\_cyst) & C(kidney) \\
pancreatic lesion & Recog. & S(pancreas\_tumor) & C(pancreas) \\
colon lesion exist. & Recog. & S(colon\_tumor) & C(colon) \\
splenomegaly detect. & Recog. & K(splenomegaly) $\to$ S(spleen) $\to$ M(spleen, vol) & K(splenomegaly) $\to$ C(spleen) \\
PDAC existence & Recog. & S(pdac) & C(pancreas) \\
PNET existence & Recog. & S(pnet) & C(pancreas) \\
\midrule
organ volume & Meas. & S(\{organ\}) $\to$ M(\{organ\}, vol) & C(\{organ\}) \\
organ HU & Meas. & S(\{organ\}) $\to$ M(\{organ\}, HU) & C(\{organ\}) \\
lesion volume & Meas. & S(\{lesion\}) $\to$ M(\{lesion\}, vol) & C(\{organ\}) \\
organ HU ratio & Meas. & S(org1) $\to$ M(org1, HU) $\to$ S(org2) $\to$ M(org2, HU) & C(org1) $\to$ C(org2) \\
tumor burden \% & Meas. & S(\{organ\}) $\to$ M(\{organ\}, vol) $\to$ S(\{lesion\}) $\to$ M(\{lesion\}, vol) & C(\{organ\}) \\
\midrule
lesion counting & Vis.R. & S(\{lesion\}) $\to$ M(\{lesion\}, count) & C(\{organ\}) \\
largest lesion diam. & Vis.R. & S(\{lesion\}) $\to$ M(\{lesion\}, diam) & C(\{organ\}) \\
largest lesion loc. & Vis.R. & S(\{lesion\}) & C(\{organ\}) \\
largest lesion atten. & Vis.R. & S(\{lesion\}) $\to$ M(\{lesion\}, HU) $\to$ S(\{organ\}) $\to$ M(\{organ\}, HU) & C(\{organ\}) \\
kidney vol. compare & Vis.R. & S(kidney\_L) $\to$ M(kidney\_L, vol) $\to$ S(kidney\_R) $\to$ M(kidney\_R, vol) & C(kidney\_L) $\to$ C(kidney\_R) \\
organ aggregation & Vis.R. & S(org1) $\to$ M(org1, vol) $\to$ S(org2) $\to$ M(org2, vol) & C(org1) $\to$ C(org2) \\
tumor-organ HU diff & Vis.R. & S(\{lesion\}) $\to$ M(\{lesion\}, HU) $\to$ S(\{organ\}) $\to$ M(\{organ\}, HU) & C(\{organ\}) \\
largest lesion slice & Vis.R. & S(\{lesion\}) & C(\{organ\}) \\
lesion outlier & Vis.R. & S(\{lesion\}) $\to$ M(\{lesion\}, vol) & C(\{organ\}) \\
lesion count by loc. & Vis.R. & S(\{lesion\}) $\to$ M(\{lesion\}, count) & C(\{organ\}) \\
inter-segment comp. & Vis.R. & S(\{lesion\}) $\to$ M(\{lesion\}, count) & C(\{organ\}) \\
adjacent organ & Vis.R. & S(\{lesion\}) & C(\{organ\}) \\
organ enlargement & Vis.R. & S(\{organ\}) $\to$ M(\{organ\}, vol) & C(\{organ\}) \\
liver lesion cluster. & Vis.R. & S(liver\_tumor$|$cyst) $\to$ M(liver, count) & C(liver) \\
bilateral kidney asym. & Vis.R. & S(kidney\_L) $\to$ M(kidney\_L) $\to$ S(kidney\_R) $\to$ M(kidney\_R) $\to$ S(lesion) & C(kidney\_L) $\to$ C(kidney\_R) \\
multi-organ burden & Vis.R. & S(org1\_lesion) $\to$ M(org1) $\to$ S(org2\_lesion) $\to$ M(org2) & C(org1) $\to$ C(org2) \\
\midrule
fatty liver & Med.R. & K $\to$ S(liver) $\to$ M(liver, HU) $\to$ S(spleen) $\to$ M(spleen, HU) & K $\to$ C(liver) \\
hepatic steatosis gr. & Med.R. & K $\to$ S(liver) $\to$ M(liver, HU) $\to$ S(spleen) $\to$ M(spleen, HU) & K $\to$ C(liver) \\
pancreatic steatosis & Med.R. & K $\to$ S(pancreas) $\to$ M(pancreas, HU) $\to$ S(spleen) $\to$ M(spleen, HU) & K $\to$ C(pancreas) \\
splenomegaly grading & Med.R. & K $\to$ S(spleen) $\to$ M(spleen, vol) & K $\to$ C(spleen) \\
PDAC vs PNET & Med.R. & K $\to$ S(pancreas\_tumor) $\to$ M(pancreas\_tumor, HU) & K $\to$ C(pancreas) \\
portal hypertension & Med.R. & K $\to$ S(spleen) $\to$ M(spleen, vol) $\to$ S(liver) $\to$ M(liver, HU) & K $\to$ C(spleen) $\to$ C(liver) \\
renal mass charact. & Med.R. & K $\to$ S(kidney\_mass) $\to$ M(kidney\_mass, HU) & K $\to$ C(kidney) \\
lesion type class. & Med.R. & K $\to$ S(kidney) $\to$ S(kidney\_mass) $\to$ M(kidney\_mass, HU) & K $\to$ C(kidney) \\
pseudocyst determ. & Med.R. & K $\to$ S(pancreas\_cyst) $\to$ M(pancreas\_cyst, HU) & K $\to$ C(pancreas) \\
pancreatic T staging & Med.R. & K $\to$ S(pancreas\_tumor) $\to$ M(pancreas\_tumor) & K $\to$ C(pancreas) \\
cyst resectability & Med.R. & K $\to$ S(pancreas\_cyst) $\to$ M(pancreas\_cyst, vol) & K $\to$ C(pancreas) \\
lesion resectability & Med.R. & K $\to$ S(pancreas\_tumor) $\to$ M(pancreas\_tumor) & K $\to$ C(pancreas) \\
\end{longtable}
\end{center}

Traces use regex-based parameter matching with \texttt{\{organ\}} placeholders substituted from question metadata at evaluation time. All trace definitions are implemented in \texttt{eval/agent\_trajectory\_eval.py} and released with the benchmark code.

\section{Training Details}
\label{sect:training_details}

\paragraph{3D Model Architectures.}
Table~\ref{tab:model_arch} compares the four 3D specialist models.

\begin{table*}[ht]
\centering
\scriptsize
\caption{Model architectures for the four 3D specialist VLMs.}
\label{tab:model_arch}
\begin{tabular}{lcccc}
\toprule
\textbf{Component} & \textbf{RadFM} & \textbf{M3D (LLaMA2 / Phi-3)} & \textbf{Merlin} \\
\midrule
Vision Encoder & 3D ViT & 3D ViT & 3D ResNet \\
Input size & [256,256,64] & [256,256,32] & [224,224,160] \\
CT spacing & direct resize & direct resize & [1.5, 1.5, 3] mm \\
LLM Decoder & LLaMA2-13B & LLaMA2-7B / Phi-3-4B & RadLLaMA-7B \\
Projector & Perceiver Resampler & 3D Pooling + MLP & 1-layer FC \\
Visual tokens & 32 & 256 & 1 \\
Pretraining data & 16M 2D+3D & 120K 3D CT & 14K 3D Abdomen CT \\
LLM tuning & full & LoRA (r=16) & LoRA (r=128) \\
\bottomrule
\end{tabular}
\end{table*}

\paragraph{3D Training Hyperparameters.}
All models are trained with AdamW and cosine scheduling in mixed precision. Training takes $\sim$48 hours per model.

\begin{table*}[ht]
\centering
\scriptsize
\caption{Training hyperparameters for 3D finetuned VLMs.}
\begin{tabular}{lcccc}
\toprule
\textbf{Item} & \textbf{RadFM} & \textbf{M3D (LLaMA2 / Phi-3)} & \textbf{Merlin} \\
\midrule
Learning rate & 5e-6 & 5e-5 & 1e-4 \\
Optimizer & AdamW (8-bit) & AdamW & AdamW \\
Mixed precision & FP16 & BF16 & BF16 \\
Per device batch size & 4 (model parallel) & 1 & 1 \\
Gradient accumulation & 8 & 8 & 8 \\
LR scheduler & Cosine & Cosine & Cosine \\
Warmup ratio & 0 & 0.03 & 0.03 \\
Training iterations & 25k & 33k & 25k \\
GPU hardware & 4$\times$A5000 24GB & 4$\times$A6000 48GB & 4$\times$A5000 24GB \\
Compute time & 48 hours & 48 hours & 48 hours \\
\bottomrule
\end{tabular}
\label{tab:hyperparameters}
\end{table*}

\paragraph{2D LoRA and Agent SFT.}
2D LoRA finetuning (Qwen3-VL-4B, Meissa-4B) on the full 428K training pairs and Agent SFT (Meissa-4B on 20K trace samples) reuse separate configurations; full hyperparameters, data preparation, and ablations are in Appendix~\ref{sect:lora_details} (2D LoRA) and Appendix~\ref{sect:agent_sft_details} (Agent SFT).

\section{Accuracy Breakdown across Demographic Factors}
\label{sect:demographic_breakdown}

\begin{figure}[t]
\centering
\includegraphics[width=1.0\textwidth]{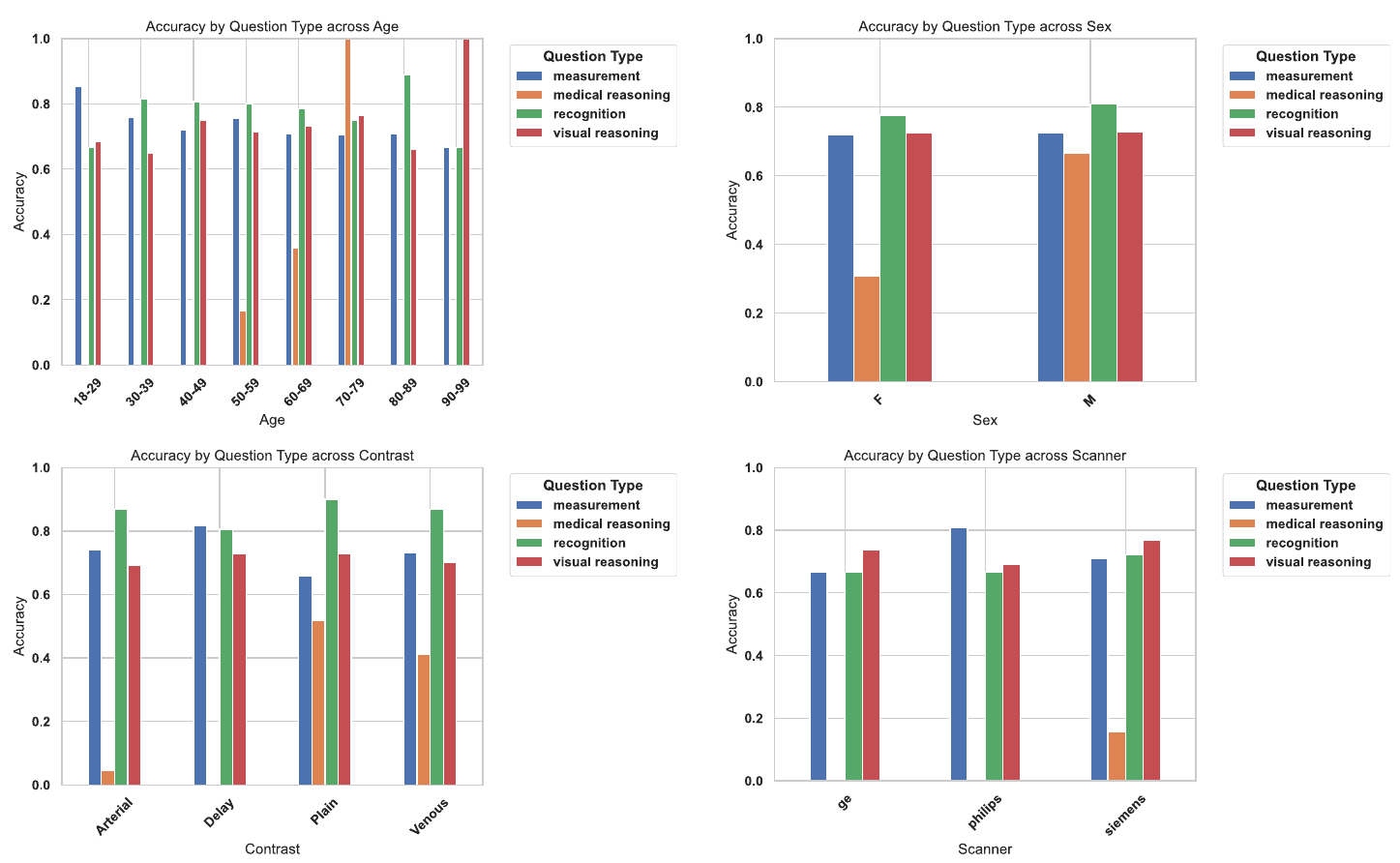}
\caption{Accuracy by question type across demographic and imaging factors.}
\label{fig:acc_by_metadata}
\end{figure}

We stratify accuracy by age, sex, CT scanner manufacturer, and contrast phase (Figure~\ref{fig:acc_by_metadata}). Recognition and measurement are stable across subgroups, while medical reasoning shows more sensitivity to scanner and contrast phase variations. Specifically, non-contrast CT scans show 3--5pp lower Medical Reasoning accuracy than contrast-enhanced scans, because HU-based diagnoses (fatty liver, steatosis grading) are calibrated for specific contrast phases. Scanner manufacturer effects are minimal ($<$2pp) for all task types, confirming that our preprocessing pipeline successfully normalizes inter-scanner variability. Age and sex show no statistically significant effects after controlling for pathology prevalence.

\section{Annotator Details}
\label{sect:annotators}

Table~\ref{tab:annotator_details} provides details on the 23 radiologists who annotated lesions for DeepTumorVQA. Annotator backgrounds range from senior abdominal imaging specialists (up to 35 years experience) to general radiologists and residents with high-volume CT reading experience.

\begin{table}[h]
\centering
\scriptsize
\caption{Detailed information of the 23 annotators.}
\label{tab:annotator_details}
\begin{tabular}{llcc}
\toprule
\textbf{No.} & \textbf{Annotator ID} & \textbf{Experience (yr)} & \textbf{CT read/year} \\
\midrule
1--7 & Specialists (S1--S7) & 19--35 & high \\
8--18 & General Radiologists (G1--G11) & 8--13 & high \\
19--23 & Residents (R1--R5) & 3--5 & high \\
\bottomrule
\end{tabular}
\end{table}

Initial segmentations were performed by general radiologists and residents. Senior specialists (S1--S7) oversaw the process by double-checking all annotations, adjudicating ambiguous cases, and ensuring anatomical and diagnostic consistency. The annotation workflow followed a three-stage protocol: (1)~\emph{Initial annotation}: General radiologists and residents performed organ and lesion segmentation using semi-automatic tools (3D Slicer with interactive thresholding and region growing). Average annotation time was 25 minutes per volume for organs and 15 minutes per lesion. (2)~\emph{Peer review}: Each annotation was reviewed by a different annotator at the same or higher seniority level. Discrepancies exceeding 10\% in volume or involving lesion count disagreements were flagged. (3)~\emph{Specialist adjudication}: Senior specialists (S1--S7) resolved all flagged cases and performed a final quality audit on a random 20\% sample of non-flagged cases. The inter-annotator agreement (Dice coefficient) for organ segmentation was 0.94 (liver), 0.96 (spleen), 0.92 (kidney), and 0.87 (pancreas). Lesion detection sensitivity among annotators (using specialist consensus as ground truth) was 0.91 for liver, 0.94 for kidney, 0.88 for pancreas, and 0.85 for colon.

\section{Radiologist Baseline Evaluation}
\label{sect:human_eval}

To calibrate model accuracy against trained clinicians, two practicing radiologists---a junior (7 years experience) and a senior (13 years), \emph{independent from the 23 annotators in Appendix~\ref{sect:annotators}}---answered a 200-question stratified subset of the benchmark (Recognition 30, Measurement 45, Visual Reasoning 75, Medical Reasoning 50; 116 distinct CT volumes). Both view the full 3D volume in 3D~Slicer with multi-planar reformatting and free windowing, substantially richer than the 5-slice tiled image given to 2D models.

\paragraph{Protocols.} The junior answers the standard 4-option MC questions (same interface as all model rows in Table~\ref{tab:vlm_results}). The senior answers the same 200 questions in \emph{free-form} (no options shown) and produces raw text. To enable an apples-to-apples row in Table~\ref{tab:vlm_results}, each free-form answer is mapped to the corresponding MC option deterministically: numeric subtypes use closest-option mapping (alphabetical tie-break), binary subtypes use yes/no semantic match, categorical subtypes use normalized exact match with longest-containment fallback; 19 ambiguous cases are resolved by manual adjudication. Per-question wall time is 145\,s (junior, MC) and 85\,s (senior, free-form).

\paragraph{Results.} Headline numbers appear in the \emph{Human eval} rows of Table~\ref{tab:vlm_results} (MC protocol) and the \emph{Human eval} row of Table~\ref{tab:freeform_eval_full} (senior, raw free-form scoring). Per question type the senior $-$ junior gap is $+$23.3pp on Recognition, $+$14.7pp on Visual Reasoning, $+$2.2pp on Measurement, and $0$pp on Medical Reasoning: experience translates strongly into lesion detection and cross-organ comparison, but not into eyeballing precise volumes/HU (a human ceiling) and not into multi-step pathology classification (which often requires contrast-phase or clinical context absent from a single CT volume). Both baselines fall below 2D-FT (66.3\%) and 3D-FT (59.8\%); the senior is comparable to M3D-Llama2 FT (54.0\%) and the junior sits between zero-shot 2D ($\sim$38\%) and 3D-FT.

\paragraph{Caveats.} The two annotators differ in both seniority and answer format, so the $+$9.5pp senior-over-junior gap conflates the two factors; reporting both rows under MC scoring discards information present in the senior's free-form answer, so the gap is a lower bound on the pure seniority effect. The 200-question subset gives $\pm$6.9pp binomial 95\% CI on Overall, so absolute numbers should be read as a calibrating reference rather than a precise comparison.

\section{Evaluation of Annotation Quality}
\label{sect:seg_quality}

To validate annotation quality, we trained segmentation models on our data and evaluated generalization on external datasets.

\paragraph{Organ Segmentation.} A U-Net model trained on our organ masks achieves strong out-of-distribution performance on TotalSegmentator and Johns Hopkins Hospital datasets (Table~\ref{tab:organ_seg}).

\begin{table}[h]
\centering
\scriptsize
\caption{Organ segmentation DSC (\%) on external datasets. ``Ours'' = model trained on DeepTumorVQA annotations.}
\label{tab:organ_seg}
\begin{tabular}{lcccc}
\toprule
\textbf{Organ} & \textbf{TotalSeg (ours)} & \textbf{TotalSeg (iid)} & \textbf{JHH (ours)} & \textbf{JHH (iid)} \\
\midrule
Liver & 94.7 & 96.3 & 95.0 & 98.6 \\
Spleen & 95.2 & 98.4 & 95.0 & 97.1 \\
Kidney (R) & 92.5 & 94.7 & 92.2 & 98.4 \\
Kidney (L) & 89.0 & 94.4 & 91.6 & 96.1 \\
Pancreas & 83.5 & 89.4 & 80.9 & 87.8 \\
\bottomrule
\end{tabular}
\end{table}

The OOD performance gap (2--6pp) is consistent with domain shift between datasets and does not indicate annotation quality issues. Pancreas segmentation shows the largest gap (5.9pp on TotalSegmentator), expected given the pancreas's anatomical variability and small size.

\paragraph{Lesion Detection.} A nnUNet model trained on our lesion annotations achieves high sensitivity even for small ($\leq$2cm) tumors (Table~\ref{tab:lesion_det}).

\begin{table}[h]
\centering
\scriptsize
\caption{Lesion detection performance on an external dataset.}
\label{tab:lesion_det}
\begin{tabular}{lcc}
\toprule
\textbf{Lesion Type} & \textbf{Sensitivity (\%)} & \textbf{Specificity (\%)} \\
\midrule
\multicolumn{3}{l}{\textit{Large Tumors $>$ 2cm}} \\
Liver Tumor (HCC) & 89.4 & 73.4 \\
Kidney Tumor (RCC) & 97.3 & 73.4 \\
Pancreas Tumor (PDAC) & 91.4 & 76.6 \\
\midrule
\multicolumn{3}{l}{\textit{Small Tumors $\leq$ 2cm}} \\
Liver Tumor (HCC) & 79.6 & 73.4 \\
Kidney Tumor (RCC) & 92.0 & 78.3 \\
Pancreas Tumor (PDAC) & 76.9 & 76.6 \\
\bottomrule
\end{tabular}
\end{table}

The high sensitivity ($>$76\% even for small tumors) on external data confirms that our annotations capture clinically relevant lesions. The lower sensitivity for small liver tumors (79.6\%) and small pancreatic tumors (76.9\%) reflects the inherent challenge of detecting sub-2cm lesions, which are at the boundary of radiological visibility on standard-resolution CT.

\section{Ethics, Limitations, and Reproducibility}
\subsection{Broader Impact Statement}
\label{sect:broader_impact}

\paragraph{Potential benefits.}
DeepTumorVQA advances the evaluation of medical AI systems in several ways. By providing a comprehensive, open-source benchmark with ground-truth annotations, it enables reproducible comparison of medical VLMs and agents. The compositional question design identifies specific capability gaps (recognition vs.\ measurement vs.\ reasoning), guiding targeted improvements. The finding that measurement is the causal bottleneck provides a clear engineering target for the community: improving quantitative tools will have outsized impact on diagnostic accuracy. The agent training framework demonstrates that small models (4B parameters) can achieve strong diagnostic performance when equipped with reliable tools, potentially democratizing access to AI-assisted diagnosis.

\paragraph{Potential risks.}
Several risks must be considered. First, \emph{clinical deployment without validation}: The benchmark evaluates models on structured multiple-choice questions, which differ fundamentally from clinical practice. High benchmark accuracy should not be interpreted as clinical readiness. Models achieving 66\% on DeepTumorVQA still fail on one-third of diagnostic questions and require extensive clinical validation before any patient-facing deployment. Second, \emph{overconfidence in tool-augmented agents}: Our findings show that oracle tools dramatically improve accuracy. While our predicted-mode experiments with TotalSegmentator demonstrate that realistic auto-segmentation retains most oracle gains (60.9\% vs.\ 63.8\% for Meissa SFT), this applies to well-validated segmentation models on common organs; performance may degrade substantially for rare pathologies or out-of-distribution anatomies where auto-segmentation quality is lower. Third, \emph{automation bias}: As AI diagnostic tools become available, clinicians may over-rely on model outputs, particularly when the model provides detailed reasoning chains (as our agent trajectories do). The benchmark's error analysis shows that models produce confident but wrong answers in 37\% of failures, highlighting the danger of trusting model explanations without verification.

\paragraph{Dual use considerations.}
The released data, tools, and models are intended exclusively for research on medical AI evaluation and development. The CT volumes are de-identified and sourced from publicly available datasets. The diagnostic tools operate on pre-computed annotations, not raw clinical images, limiting direct clinical application. However, the agent training framework could theoretically be adapted to build diagnostic systems; we strongly discourage clinical deployment of any models trained on this benchmark without independent clinical validation and regulatory approval.

\paragraph{Data privacy considerations.}
All CT volumes in DeepTumorVQA are sourced from publicly available, de-identified datasets that have undergone institutional review board (IRB) approval at their respective institutions. No personally identifiable information (PII) is included in our metadata, annotations, or benchmark questions. The structured reports we generate contain only anatomical measurements and diagnostic classifications derived from segmentation masks, with no reference to patient identity, clinical history, or treatment information. The annotations, structured metadata, and benchmark questions contributed by this work are released under CC-BY-4.0 (requiring attribution but allowing unrestricted research use); the underlying CT volumes remain governed by their respective source-dataset licenses and users must comply with each source's terms of use.

\subsection{Limitations (Detailed)}
\label{sect:limitations_detailed}

We expand on the limitations mentioned in the main text discussion.

\paragraph{Annotation quality and coverage.}
Dataset construction relies on segmentation quality from 23 radiologists, introducing noise for ambiguous cases. Specific limitations include: (i)~\emph{Small lesion detection}: Lesions $<$5mm may be missed by annotators, particularly in the pancreas and colon. Our nnUNet validation shows 76.9\% sensitivity for small ($<$2cm) pancreatic tumors, implying $\sim$23\% of small lesions are potentially unannotated. This affects Recognition accuracy for these subtypes. (ii)~\emph{Lesion type classification}: The distinction between cyst and tumor is based on HU thresholds and annotator judgment; borderline cases (complex cysts with solid components) may be misclassified. (iii)~\emph{Couinaud segment assignment}: Liver lesion segment assignment is approximate and may vary between annotators, affecting the inter-segment comparison subtype. (iv)~\emph{Limited pathology diversity}: The dataset focuses on common abdominal pathologies (HCC, RCC, PDAC, cysts). Rare conditions (cholangiocarcinoma, metastases from extrahepatic primaries, rare cystic neoplasms) are underrepresented.

\paragraph{Benchmark scope.}
The benchmark focuses on abdominal CT, covering 5 major organs (liver, kidney, pancreas, spleen, colon). Extension to other anatomies (chest, brain, pelvis) and modalities (MRI, PET-CT, ultrasound) is planned but not included. The current task space, while covering 42 subtypes, does not capture all clinically relevant assessments: texture analysis (liver fibrosis scoring), morphological assessment (tumor margin characterization), temporal comparison (treatment response evaluation), and multi-phase analysis (arterial vs.\ portal venous enhancement patterns) are not included. These require annotations beyond segmentation masks.

\paragraph{Multiple-choice format.}
The 4-option multiple-choice format enables unambiguous evaluation across 30+ models but introduces limitations: (i)~Models can exploit option patterns (e.g., the correct answer distribution is approximately uniform by construction, which a model could learn). (ii)~The format does not test the ability to generate precise numerical answers; our distractor spacing (3--10\% of GT) defines the effective precision tested. (iii)~Clinical diagnosis often involves open-ended reporting, not selecting from predefined options. To address this, we run a full free-form evaluation of all models (Appendix~\ref{sect:freeform_eval}) and train free-form variants of Meissa-4B and Qwen3-VL-4B. The qualitative conclusions of the MC protocol (2D-FT $>$ 3D-FT $>$ zero-shot 2D $>$ pretrained 3D; measurement bottleneck for zero-shot models; finetuning value) are preserved under free-form scoring, with within-category re-orderings concentrated in the 3D-FT row (RadFM overtakes M3D-Phi3 due to better numeric generation) and the top-2 2D-FT pair (Qwen3-VL FF overtakes Meissa FF on Med.\ Reasoning). These are within-category swaps rather than cross-category reversals, supporting the MC findings as conclusions about model families rather than format artifacts.

\paragraph{Agent evaluation constraints.}
Agent evaluation is computationally expensive ($\sim$4--72 seconds/sample depending on configuration) and limited by: (i)~\emph{Maximum step limit}: The 8-step limit may be insufficient for complex Medical Reasoning traces requiring 5--6 tool calls plus reasoning. Our error analysis shows 38\% of failures hit this limit. (ii)~\emph{Tool quality spectrum}: We evaluate three tool tiers---oracle (GT), predicted (TotalSegmentator), and vision (crops only). Predicted mode shows realistic auto-segmentation retains most oracle gains ($-$2.9pp for SFT agents), but covers only common abdominal organs where TotalSegmentator achieves high DSC ($>$90\%). Performance with less mature segmentation models or for rare anatomical structures may degrade more substantially. (iii)~\emph{Text-based tool interface}: Tools return JSON text, not visual overlays. In clinical practice, radiologists see segmentation overlays on the CT image, enabling visual verification. (iv)~\emph{Single-turn questions}: Each question is independent; the benchmark does not test multi-turn diagnostic conversations where earlier answers inform later questions.

\paragraph{Generalization claims.}
Our sufficiency finding (measurements alone support strong reasoning) is demonstrated within DeepTumorVQA's structured task space. Broader clinical diagnosis requires capabilities not captured here: assessing tumor margin invasion from CT morphology, characterizing lesion enhancement patterns across contrast phases, evaluating post-treatment response, and integrating clinical history with imaging findings. The sufficiency of measurement is specific to our benchmark's task definitions, where answers can be computed from quantitative metadata; this should not be extrapolated to all of clinical radiology.

\paragraph{Agent tool interface.}
The current agent environment abstracts tool outputs as structured JSON text, omitting visual feedback (segmentation mask overlays) that radiologists rely on. This design isolates quantitative reasoning from visual interpretation but limits evaluation of multimodal integration capabilities. Future versions should incorporate visual tool outputs to assess whether agents can leverage spatial context from segmentation overlays.

\paragraph{Knowledge base scope.}
The 27-entry knowledge base covers well-codified criteria for the 12 Medical Reasoning subtypes. Its minimal impact (+3.1pp when removed) reflects that capable models internalize common clinical thresholds during pretraining. A larger, less codified knowledge base covering rare conditions would likely show greater tool benefit, but testing this requires expanding beyond well-established diagnostic criteria.

\subsection{Reproducibility Statement}
\label{sect:reproducibility}

We provide comprehensive details to ensure full reproducibility of all results reported in this paper.

\paragraph{Exact software versions.}
All experiments use the following software stack:
\begin{itemize}
\item \textbf{Python}: 3.10.14
\item \textbf{PyTorch}: 2.8.0 (CUDA 12.4) for most models; 2.10.0 for RadFM and Qwen3.5
\item \textbf{Transformers}: 4.57.3 for Qwen3-VL, Meissa, MedGemma, M3D, Merlin; 5.3.0 for Qwen3.5
\item \textbf{vLLM}: 0.11.0 (for Meissa inference)
\item \textbf{LLaMA-Factory}: 0.9.1 (for agent SFT training)
\item \textbf{nibabel}: 5.2.0 (NIfTI I/O)
\item \textbf{scipy}: 1.12.0 (image processing)
\item \textbf{numpy}: 1.26.4
\item \textbf{pandas}: 2.1.4
\end{itemize}

\paragraph{Hardware requirements.}
\begin{itemize}
\item \textbf{Direct inference (4B models)}: 1$\times$NVIDIA A5000 (24GB VRAM), $\sim$1 hour for 10K questions
\item \textbf{Direct inference (7--9B models)}: 1$\times$NVIDIA A6000 (48GB VRAM), $\sim$2 hours for 10K questions
\item \textbf{Direct inference (35B MoE)}: 4$\times$NVIDIA A6000 (192GB total), $\sim$4.5 hours for 10K questions
\item \textbf{Agent evaluation (oracle)}: 1$\times$A5000 or A6000, $\sim$11 hours for 10K questions (with tool cache)
\item \textbf{Agent evaluation (without cache)}: 1$\times$GPU + fast SSD, $\sim$42 hours for 10K questions
\item \textbf{3D model finetuning}: 4$\times$A5000 or 4$\times$A6000, $\sim$48 hours per model
\item \textbf{2D LoRA finetuning}: 4$\times$A5000, $\sim$12 hours per model
\item \textbf{Agent SFT training}: 4$\times$A6000, $\sim$8 hours
\item \textbf{Tool cache pre-computation}: 32 CPU cores, $\sim$8 hours for 991 test images
\item \textbf{2D slice/video generation}: 32 CPU cores, $\sim$45 minutes for 991 images
\end{itemize}

\paragraph{Runtime estimates for full evaluation.}
Reproducing all 30+ model evaluations on the 10K benchmark requires approximately:
\begin{itemize}
\item 25 direct inference runs: $\sim$50 GPU-hours (varying by model size)
\item 9 agent runs (4 oracle + 2 predicted + 3 vision): $\sim$99 GPU-hours (with cache)
\item 5 ablation conditions: $\sim$15 GPU-hours
\item 2 measurement-only runs: $\sim$4 GPU-hours
\item Total: $\sim$168 GPU-hours ($\sim$7 days on a single A6000)
\end{itemize}
With 8--16 parallel GPUs (typical academic cluster allocation), the full evaluation can be completed in 1--2 days.

\paragraph{Random seeds and determinism.}
All inference uses greedy decoding (\texttt{temperature=0}, \texttt{do\_sample=False}), ensuring deterministic outputs for a given model and input. Training uses seed 42 for data shuffling, weight initialization, and dropout. The benchmark question set is fixed (\texttt{benchmark\_v4.json}), and the train/test split is deterministic (patient-level split by dataset). The distractor generation uses seed 42 for random option positioning. Under these conditions, results are fully reproducible to the reported precision (one decimal place for percentages).

\paragraph{Code and data availability.}
All code, data, and evaluation scripts are released at \href{https://github.com/Schuture/DeepTumorVQA}{github.com/Schuture/DeepTumorVQA}. The repository includes:
\begin{itemize}
\item Benchmark data (\texttt{benchmark/benchmark\_v4.json})
\item Dataset construction scripts (\texttt{dataset\_construction\_scripts/})
\item Inference scripts for all models (\texttt{benchmark/run\_*.py})
\item Agent evaluation framework (\texttt{eval/})
\item Tool cache pre-computation (\texttt{benchmark/precompute\_tool\_results.py})
\item Agent SFT data generation (\texttt{benchmark/prepare\_agent\_sft\_data.py})
\item Result analysis notebooks
\end{itemize}
Pre-computed 2D slices, videos, and 3D arrays are available on HuggingFace: \href{https://huggingface.co/datasets/tumor-vqa/DeepTumorVQA\_1.0}{tumor-vqa/DeepTumorVQA\_1.0}.

\section{Distractor Separation Analysis}
\label{sect:distractor_analysis}

A natural concern is whether the noise ablation's $-$10.2pp degradation at 10\% noise is driven primarily by tight distractor spacing rather than genuine precision sensitivity. We analyze this by computing the relative distractor gap (distance from correct answer to nearest distractor, as a fraction of the correct value) for all continuous-valued subtypes, then correlating this with per-subtype noise sensitivity.

\begin{figure}[h]
\centering
\includegraphics[width=0.9\linewidth]{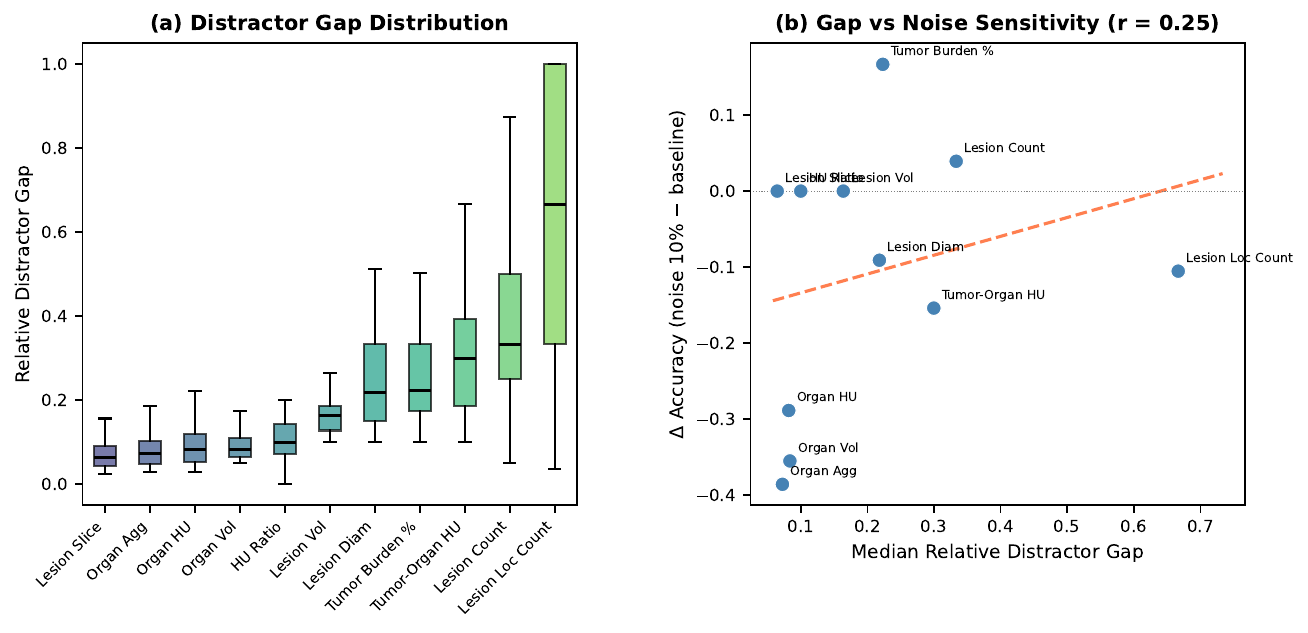}
\caption{Distractor separation analysis. \textbf{Left}: Distribution of relative distractor gaps by subtype. \textbf{Right}: Per-subtype median gap vs.\ accuracy drop under 10\% noise. No significant correlation ($r = 0.25$, $p = 0.45$), indicating noise sensitivity reflects task structure, not option spacing.}
\label{fig:distractor_separation}
\end{figure}

Figure~\ref{fig:distractor_separation} shows the result. The Pearson correlation between median distractor gap and accuracy drop under 10\% noise is $r = 0.25$ ($p = 0.45$), indicating \textbf{no statistically significant relationship}. Subtypes with the tightest spacing (organ aggregation: median gap 7.2\%) show large noise drops ($-$38.6pp), but subtypes with moderate spacing (tumor burden: 22.3\%) also show substantial drops. The lack of correlation indicates that noise sensitivity reflects genuine task-level precision requirements---not an artifact of option design. Tasks requiring ratio computation or multi-organ comparison are most noise-sensitive regardless of distractor spacing, because noise compounds multiplicatively through multi-step computations.

\section{Predicted-Mode Capability Matrix}
\label{sect:predicted_matrix}

TotalSegmentator covers 25 abdominal organs but does not segment lesions (tumors, cysts). Table~\ref{tab:predicted_capability} classifies all 42 subtypes by support level and reports Meissa SFT accuracy under oracle vs.\ predicted tools.

\begin{table}[h]
\centering
\small
\caption{Predicted-mode retention by TotalSegmentator support level. Retention = $(\text{pred} - 25\%) / (\text{oracle} - 25\%)$.}
\label{tab:predicted_capability}
\begin{tabular}{l|c|cc|c}
\toprule
\textbf{Support Level} & \textbf{N Sub.} & \textbf{Oracle} & \textbf{Predicted} & \textbf{Retention} \\
\midrule
Fully supported (organ-only) & 12 & 70.2\% & 63.6\% & 85.4\% \\
Partially supported (organ+lesion) & 5 & 54.4\% & 52.9\% & 94.8\% \\
Unsupported (lesion-only) & 25 & 59.0\% & 59.5\% & 101.6\% \\
\midrule
\textbf{All subtypes} & \textbf{42} & \textbf{63.8\%} & \textbf{60.9\%} & \textbf{92.6\%} \\
\bottomrule
\end{tabular}
\end{table}

For unsupported (lesion-only) subtypes, predicted-mode accuracy does not drop in aggregate (101.6\% retention). \emph{This is not because TotalSegmentator successfully segments lesions---our predicted cache contains 0\% lesion mask coverage for liver/kidney/pancreas/colon tumors.} Inspection of agent trajectories reveals that when the lesion segmentation tool returns \texttt{mask\_found=False}, the SFT agent has learned to fall back on organ-level measurements (HU, volume) and parametric clinical knowledge to infer lesion properties. This fallback systematically over-performs on subtypes where the diagnostic criterion is encoded in organ-level features---pseudocyst determination (oracle 33.3\%, predicted 66.7\%; criterion is pancreas HU $>$14.5), renal mass characterization (43.5\% $\to$ 78.3\%), pancreatic lesion existence (37.7\% $\to$ 66.5\%), and lesion type classification (35.7\% $\to$ 47.6\%). Conversely, subtypes that genuinely require lesion geometry degrade as expected: kidney tumor existence 78.2\% $\to$ 61.3\%, pancreatic lesion resectability 65.3\% $\to$ 55.1\%, lesion volume measurement 63.3\% $\to$ 57.1\%. The aggregate 101.6\% therefore reflects \emph{learned-heuristic compensation} rather than segmentation success; the most faithful indicator of TotalSegmentator-based predicted realism is the 85.4\% retention on the 12 fully-supported organ-only subtypes, where the gap is bounded by the difference between TotalSegmentator and ground-truth organ boundaries. The full per-subtype breakdown is available in \texttt{data/predicted\_mode\_capability.csv}.

\section{Recognition Subtype Measurement Dependency}
\label{sect:recog_meas_boundary}

Table~\ref{tab:recog_meas} classifies the 9 Recognition subtypes by whether their ground-truth answer depends on quantitative measurement. Eight of nine subtypes are purely morphological (lesion mask presence), while splenomegaly detection requires a volume threshold, making it a clinically natural hybrid of recognition and measurement.

\begin{table}[h]
\centering
\small
\caption{Recognition subtypes and their measurement dependency.}
\label{tab:recog_meas}
\begin{tabular}{l|c|l}
\toprule
\textbf{Subtype} & \textbf{Meas.\ Req.?} & \textbf{Criterion} \\
\midrule
Liver lesion existence & No & Mask presence \\
Kidney lesion existence & No & Mask presence \\
Kidney tumor existence & No & Mask presence \\
Kidney cyst existence & No & Mask presence \\
Pancreatic lesion existence & No & Mask presence \\
Colon lesion existence & No & Mask presence \\
PDAC existence & No & Mask presence \\
PNET existence & No & Mask presence \\
\textbf{Splenomegaly detection} & \textbf{Yes} & \textbf{Vol $>$ 314.5\,cm$^3$} \\
\bottomrule
\end{tabular}
\end{table}

\section{Agent Tool Definitions}
\label{sect:tool_definitions}

We provide detailed specifications for each of the four diagnostic tools available in the agent environment.

\paragraph{1.\ \texttt{segment\_organ(target)}}
Queries ground-truth segmentation statistics for a specified anatomical target.

\begin{itemize}
\item \textbf{Input:} \texttt{target} (string): one of 43 supported targets (\eg \texttt{liver}, \texttt{kidney\_left}, \texttt{pancreas\_tumor}, \texttt{kidney\_cyst}).
\item \textbf{Output:} JSON with fields: \texttt{mask\_found} (bool), \texttt{voxel\_count} (int), \texttt{bounding\_box} (6-tuple), \texttt{center\_of\_mass} (3-tuple).
\item \textbf{Target name aliases:} The tool supports flexible name matching to handle model-generated variations. For example, \texttt{kidney} maps to both \texttt{kidney\_left} and \texttt{kidney\_right}; \texttt{pancreatic\_mass} maps to \texttt{pancreas\_tumor}; \texttt{liver\_lesion} maps to \texttt{liver\_tumor}. The full alias table contains 87 mappings covering common model-generated names.
\end{itemize}

\begin{tcolorbox}[title={Example: segment\_organ}, colback=gray!5, colframe=gray!50!black, fonttitle=\bfseries\scriptsize, fontupper=\scriptsize]
\textbf{Call:} \texttt{segment\_organ(``liver\_tumor'')}\\[2pt]
\textbf{Response:} \texttt{\{``mask\_found'': true, ``voxel\_count'': 8432, ``bounding\_box'': [102, 45, 78, 118, 63, 94], ``center\_of\_mass'': [110.2, 54.1, 86.3]\}}
\end{tcolorbox}

\paragraph{2.\ \texttt{measure(target, type)}}
Returns a ground-truth quantitative measurement for a specified target and measurement type.

\begin{itemize}
\item \textbf{Input:} \texttt{target} (string), the anatomical target; \texttt{type} (string), one of \texttt{volume}, \texttt{mean\_HU}, \texttt{diameter}, \texttt{count}.
\item \textbf{Output:} JSON with fields: \texttt{value} (float), \texttt{unit} (string).
\item \textbf{Supported types:} \texttt{volume} (cm$^3$), \texttt{mean\_HU} (Hounsfield units), \texttt{diameter} (cm, largest lesion), \texttt{count} (integer, number of instances).
\item \textbf{Computation details:} Volume is computed as voxel count $\times$ voxel volume (from affine matrix). Mean HU is computed over all voxels within the segmentation mask. Diameter is the maximum Feret diameter of the largest connected component in the axial plane. Count uses 3D connected component analysis with 26-connectivity.
\end{itemize}

\begin{tcolorbox}[title={Example: measure}, colback=gray!5, colframe=gray!50!black, fonttitle=\bfseries\scriptsize, fontupper=\scriptsize]
\textbf{Call:} \texttt{measure(``liver'', ``volume'')}\\[2pt]
\textbf{Response:} \texttt{\{``value'': 1293.7, ``unit'': ``cm3''\}}
\end{tcolorbox}

\paragraph{3.\ \texttt{lookup\_medical\_knowledge(query)}}
Retrieves clinical criteria from a curated 27-entry knowledge base covering diagnostic thresholds, grading systems, and classification criteria.

\begin{itemize}
\item \textbf{Input:} \texttt{query} (string): clinical topic (\eg \texttt{``fatty liver''}, \texttt{``splenomegaly grading''}).
\item \textbf{Output:} JSON with fields: \texttt{entries} (list of dicts with \texttt{criterion}, \texttt{threshold}, \texttt{source}).
\item \textbf{Query matching:} Uses fuzzy string matching (Levenshtein distance $\leq$ 3) to handle model-generated query variations. For example, ``fatty liver disease'', ``hepatic steatosis'', and ``NAFLD'' all match the fatty liver entry.
\item \textbf{Knowledge base coverage:} The 27 entries cover: fatty liver (Zeb 2012), hepatic steatosis grading (Kodama 2007), pancreatic steatosis (Guneyli 2022), splenomegaly detection and grading (Bezerra 2005), portal hypertension (Harbin 1980), PDAC vs PNET classification (NCCN 2024), renal mass Bosniak classification (Silverman 2019), kidney lesion type (Agochukwu 2017), pseudocyst determination (Allen 2011), pancreatic T staging (AJCC 8th), pancreatic cyst resectability (Tanaka 2012), and pancreatic lesion resectability (NCCN 2024), plus 15 supplementary entries covering organ size norms, HU reference ranges, and lesion characterization criteria.
\end{itemize}

\begin{tcolorbox}[title={Example: lookup\_medical\_knowledge}, colback=gray!5, colframe=gray!50!black, fonttitle=\bfseries\scriptsize, fontupper=\scriptsize]
\textbf{Call:} \texttt{lookup\_medical\_knowledge(``fatty liver'')}\\[2pt]
\textbf{Response:} \texttt{\{``entries'': [\{``criterion'': ``Liver-to-spleen HU ratio < 1.0 indicates fatty liver'', ``threshold'': ``L/S ratio < 1.0'', ``source'': ``Zeb et al.\ 2012''\}]\}}
\end{tcolorbox}

\paragraph{4.\ \texttt{crop\_region(organ)}}
Returns organ-focused multi-slice crops for visual inspection (vision mode only).

\begin{itemize}
\item \textbf{Input:} \texttt{organ} (string): target organ (\eg \texttt{``liver''}, \texttt{``kidney''}, \texttt{``pancreas''}).
\item \textbf{Output:} JSON with field: \texttt{image\_base64} (string), a base64-encoded PNG of 5 representative axial slices tiled in a single image with abdominal windowing.
\item \textbf{Crop details:} The bounding box is computed from the organ segmentation mask with 20\% padding on each side (minimum 10 pixels). Slices are selected at 5 equally-spaced positions within the organ's axial extent. Windowing uses $W = 400$, $L = 50$ (identical to the pre-computed 2D slices). The resulting image is encoded as PNG and base64-encoded for inclusion in the JSON response.
\end{itemize}

\begin{tcolorbox}[title={Example: crop\_region}, colback=gray!5, colframe=gray!50!black, fonttitle=\bfseries\scriptsize, fontupper=\scriptsize]
\textbf{Call:} \texttt{crop\_region(``kidney'')}\\[2pt]
\textbf{Response:} \texttt{\{``image\_base64'': ``iVBORw0KGgo...''\}} (5 axial slices centered on the kidney bounding box, tiled horizontally)
\end{tcolorbox}

\section{Tolerance-Based Evaluation of Measurement Subtypes}
\label{sect:tolerance_eval}

To address whether the multiple-choice format artificially constrains evaluation, we re-score all continuous-value measurement subtypes (2,169 questions across 6 subtypes) using tolerance-based accuracy: a prediction is correct if the selected numeric value falls within $\epsilon$\% of the ground truth. Table~\ref{tab:tolerance_eval} reports results at four tolerance thresholds.

\begin{table}[h]
\centering
\small
\caption{Tolerance-based vs.\ MC accuracy on measurement subtypes ($N$=2,169). Model rankings are perfectly preserved across all thresholds ($r$=0.999 at $\epsilon$=5\%, $r$=0.986 at 10\%).}
\label{tab:tolerance_eval}
\begin{tabular}{l|r|rrrr}
\toprule
\textbf{Model} & \textbf{MC Acc} & \textbf{Tol-5\%} & \textbf{Tol-10\%} & \textbf{Tol-20\%} & \textbf{Tol-50\%} \\
\midrule
Meissa SFT agent & 76.6\% & 79.0\% & 84.7\% & 92.7\% & 98.8\% \\
M3D-Phi3 FT & 70.6\% & 73.4\% & 84.0\% & 96.2\% & 99.7\% \\
Meissa FT & 67.0\% & 69.7\% & 83.9\% & 96.7\% & 100.0\% \\
Meissa oracle & 61.4\% & 66.7\% & 71.9\% & 83.4\% & 95.9\% \\
Gemini-3-Flash & 34.4\% & 37.7\% & 52.9\% & 76.4\% & 96.1\% \\
Qwen3.5-9B & 28.5\% & 31.0\% & 45.6\% & 72.9\% & 96.2\% \\
\bottomrule
\end{tabular}
\end{table}

MC accuracy and tolerance-5\% accuracy are nearly identical ($\Delta$=2--5pp), confirming that MC distractors are well-calibrated and the format does not artificially inflate or deflate model rankings. The small gap reflects cases where the model selects a distractor numerically close to (but not exactly matching) the correct answer. At looser thresholds ($\epsilon$=10--20\%), accuracy rises substantially, indicating that many ``errors'' are near-misses rather than fundamental failures. Crucially, model \emph{rankings} are perfectly preserved: the rank-order correlation between MC accuracy and tolerance-based accuracy is $r$=0.999 ($p$<0.001) at $\epsilon$=5\%, demonstrating that MC evaluation is a reliable proxy for open-ended numeric assessment on this benchmark.

\section{Free-Form Evaluation on the Full Benchmark}
\label{sect:freeform_eval}

While Appendix~\ref{sect:tolerance_eval} establishes that MC evaluation is a faithful proxy for open-ended numeric assessment on the 6 continuous-value subtypes, we additionally evaluate every model on the \emph{full} benchmark (10{,}000 questions, 42 subtypes) in a free-form setting to directly address the concern that the MC interface might mask true generation capability. Each model answers the original open-ended question (no A/B/C/D options provided) and the raw text output is scored under a subtype-appropriate metric. For consistency with the main finetuning protocol (Table~\ref{tab:vlm_results}), we additionally train \textbf{Meissa-4B FF} and \textbf{Qwen3-VL-4B FF} with the same 428K LoRA recipe but using the free-form \texttt{answer} column as the target (no option letters), replacing the MC-trained 2D FT rows in this table.

\paragraph{Scoring protocol.}
Each of the 42 subtypes is assigned one of three answer types based on the semantic form of its ground-truth answer. All scoring is performed on raw text output after stripping \texttt{<think>\ldots</think>} blocks, normalizing whitespace, lower-casing, and removing punctuation (except hyphens).

\emph{Numeric subtypes} (11 subtypes, $N$=2{,}023; organ volume/HU measurement, lesion volume/diameter, lesion counting/count-by-location, tumor burden percentage, HU ratios, tumor-organ HU difference). The first signed decimal match in the model output is extracted as the prediction $\hat{y}$. Per-question score is the Mean Relative Accuracy (MRA)
\[
\mathrm{MRA}(\hat{y}, y) \;=\; \max\!\Big(0,\; 1 - \tfrac{|\hat{y}-y|}{|y|}\Big)\quad\in\,[0,1],
\]
with $\mathrm{MRA}=1$ if $y=0$ and $\hat{y}=0$, and $\mathrm{MRA}=0$ otherwise. Unparseable outputs (no number extractable, off-units, or malformed) receive $\mathrm{MRA}=0$---this is the central reason why top-model numeric accuracy is several points lower in free-form than in MC: there is no 4-option floor, and a missing or wrongly-unit'd digit sends the score to zero. We additionally report tolerance-based accuracy at thresholds $\epsilon\in\{5\%,10\%,20\%,50\%\}$, counting a prediction correct if the relative error $|\hat{y}-y|/|y|\le\epsilon$; these provide bounded comparisons at coarser precisions and are reported per-subtype in the released JSON but collapsed to MRA in Table~\ref{tab:freeform_eval_full} for uniformity with binary/categorical rows.

\emph{Binary subtypes} (15 subtypes, $N$=2{,}823; lesion-existence per organ, PDAC/PNET existence, organ enlargement, splenomegaly detection, pancreatic steatosis, pancreatic cyst resectability, pseudocyst determination, lesion outlier, portal hypertension assessment). The prediction is classified as ``yes'' if the output begins with or contains the standalone token \texttt{yes} (similarly for \texttt{no}); otherwise it is counted as an unparseable negative prediction against the ground-truth polarity (i.e., counted as a false negative if ground truth is yes, false positive if no). The primary metric is exact-match accuracy on 10K questions; we additionally report per-subtype sensitivity $\mathrm{TP}/(\mathrm{TP}{+}\mathrm{FN})$ and specificity $\mathrm{TN}/(\mathrm{TN}{+}\mathrm{FP})$ in the released JSON.

\emph{Categorical subtypes} (16 subtypes, $N$=5{,}154; grading subtypes splenomegaly/hepatic-steatosis, staging pancreatic T, classification PDAC-vs-PNET and kidney lesion type, spatial relations such as largest-lesion location, inter-segment comparison, adjacent organ, bilateral-kidney asymmetry, fatty liver, liver-lesion clustering, kidney-volume comparison, largest-lesion attenuation, renal mass characterization, multi-organ lesion burden comparison). We compute strict normalized exact match (the primary score used in Table~\ref{tab:freeform_eval_full}) and a partial match variant (credit if the normalized ground truth is contained in the prediction or vice versa) released in the JSON. Unlike numeric and binary scoring, categorical does \emph{not} coerce unparseable outputs onto a known label---any extra words in the output that do not exactly match the ground-truth phrase reduce accuracy, again explaining part of the MC--FF gap for staging/grading subtypes where the MC choice preserves the exact phrasing.

\paragraph{Aggregation.}
Per-subtype scores are micro-averaged within each question type (Recognition, Measurement, Visual Reasoning, Medical Reasoning) by question count; the Overall column is the micro-average across all 42 subtypes. This differs slightly from the main-text MC protocol (which averages exact-match accuracy), but is consistent across all rows of Table~\ref{tab:freeform_eval_full}. Full parser, metric code, and per-subtype JSON outputs are released with the benchmark.

\begin{table*}[t]
\centering
\scriptsize
\setlength{\tabcolsep}{2pt}
\caption{Free-form evaluation results on DeepTumorVQA (10K questions, 42 subtypes). Each score is the micro-averaged subtype accuracy within a category; subtype metric is MRA (numeric), exact match (binary/categorical). Best per-category in \textbf{bold}. FF = LoRA-finetuned with the same 428K training set as MC FT (Table~\ref{tab:vlm_results}) but with open-ended \texttt{answer} targets instead of option letters.}
\label{tab:freeform_eval_full}
\begin{tabular}{ll|ccl|ccccc}
\toprule
\textbf{Cat.} & \textbf{Model} & \textbf{Params} & \textbf{Input} & \textbf{Notes} & \textbf{Overall} & \textbf{Recog.} & \textbf{Meas.} & \textbf{Vis.R.} & \textbf{Med.R.} \\
\midrule
\multirow{4}{*}{\rotatebox{90}{\scriptsize 3D FT}}
& M3D-Llama2~\cite{bai2024m3d} & 7B & 3D vol & & 32.4\% & 53.4\% & 25.1\% & 35.7\% & 11.1\% \\
& M3D-Phi3~\cite{bai2024m3d} & 4B & 3D vol & & 27.1\% & \textbf{60.8\%} & 20.5\% & 21.3\% & 10.7\% \\
& Merlin~\cite{blankemeier2024merlin} & 7B & 3D vol & ResNet enc. & 24.5\% & 0.0\% & 49.7\% & 35.7\% & 0.0\% \\
& RadFM~\cite{wu2023towards} & 14B & 3D vol & Largest 3D & \textbf{50.5\%} & 67.2\% & \textbf{55.6\%} & \textbf{56.0\%} & \textbf{15.5\%} \\
\midrule
\multirow{3}{*}{\rotatebox{90}{\scriptsize 3D Pre}}
& M3D-Llama2 pre.~\cite{bai2024m3d} & 7B & 3D vol & No FT & 6.8\% & 10.4\% & 6.7\% & 7.0\% & 2.5\% \\
& M3D-Phi3 pre.~\cite{bai2024m3d} & 4B & 3D vol & No FT & \textbf{21.7\%} & \textbf{57.2\%} & \textbf{15.8\%} & \textbf{14.7\%} & \textbf{5.1\%} \\
& RadFM pre.~\cite{wu2023towards} & 14B & 3D vol & No FT & 6.2\% & 0.1\% & 13.5\% & 8.3\% & 0.6\% \\
\midrule
\multirow{2}{*}{\rotatebox{90}{\scriptsize 2D FT}}
& Meissa FF~\cite{chen2025meissa} & 4B & 2D & LoRA, FF target & 61.3\% & 70.1\% & \textbf{59.8\%} & \textbf{61.5\%} & 53.0\% \\
& Qwen3-VL FF~\cite{qwen3vl2025} & 4B & 2D & LoRA, FF target & \textbf{63.4\%} & \textbf{71.0\%} & 58.7\% & 61.1\% & \textbf{65.1\%} \\
\midrule
\multirow{11}{*}{\rotatebox{90}{\scriptsize 2D Zero-shot}}
& HuatuoGPT~\cite{chen2024huatuogpt} & 7B & 2D & Medical & 29.2\% & \textbf{57.5\%} & 23.8\% & 26.3\% & 10.9\% \\
& InternVL3-8B~\cite{internvl2024} & 8B & 2D & & 29.5\% & 56.0\% & 32.3\% & 23.8\% & 10.3\% \\
& LLaVA-Med~\cite{li2023llava} & 7B & 2D & Medical & 24.6\% & 53.6\% & 38.4\% & 12.1\% & 5.6\% \\
& LLaVA-OV~\cite{liu2024llavanext} & 7B & 2D & & 29.6\% & 56.4\% & 35.0\% & 22.6\% & 10.2\% \\
& MedGemma 2D~\cite{google2025medgemma} & 4B & 2D & Medical & 28.6\% & 52.0\% & 37.9\% & 22.2\% & 7.6\% \\
& MedGemma 3D~\cite{google2025medgemma} & 4B & 2D 85-sl. & 85-slice RGB & 25.4\% & 47.6\% & 29.4\% & 20.1\% & 8.4\% \\
& Meissa-4B~\cite{chen2025meissa} & 4B & 2D & Medical & 25.6\% & 43.4\% & 28.8\% & 23.5\% & 7.8\% \\
& Qwen3-VL-4B~\cite{qwen3vl2025} & 4B & 2D & & \textbf{34.6\%} & 59.4\% & 44.6\% & \textbf{27.3\%} & \textbf{13.5\%} \\
& Qwen3-VL-8B~\cite{qwen3vl2025} & 8B & 2D & & 32.6\% & 55.2\% & 44.2\% & 25.3\% & 11.8\% \\
& Qwen3.5-9B~\cite{qwen2025qwen35} & 9B & 2D & & 30.8\% & 58.0\% & 35.0\% & 24.5\% & 10.8\% \\
& Qwen3.5-35B~\cite{qwen2025qwen35} & 35B(3B) & 2D & MoE & 32.7\% & 54.2\% & \textbf{47.3\%} & 24.5\% & 11.7\% \\
\midrule
\multirow{3}{*}{\rotatebox{90}{\scriptsize Video}}
& Meissa-4B (video)~\cite{chen2025meissa} & 4B & Video & & 29.2\% & 45.4\% & 37.9\% & 24.6\% & 12.7\% \\
& Qwen3-VL-4B (vid.)~\cite{qwen3vl2025} & 4B & Video & & \textbf{33.3\%} & \textbf{57.3\%} & 40.9\% & \textbf{27.2\%} & \textbf{12.8\%} \\
& Qwen3-VL-8B (vid.)~\cite{qwen3vl2025} & 8B & Video & & 32.6\% & 57.2\% & \textbf{42.3\%} & 25.5\% & 11.2\% \\
\midrule
\multirow{1}{*}{\rotatebox{90}{\scriptsize API}}
& Gemini-3-Flash~\cite{google2025gemini3} & -- & 2D & API & 36.4\% & 43.5\% & 58.7\% & 33.4\% & 11.5\% \\
\midrule
\multirow{1}{*}{\rotatebox{90}{\scriptsize Hum.}}
& Senior radiologist & -- & 3D vol & free-form, 200 subset & 48.0\% & 70.0\% & \textbf{73.7\%} & 40.2\% & 24.0\% \\
\bottomrule
\end{tabular}
\end{table*}

\paragraph{Findings.}
The free-form results reproduce the qualitative conclusions drawn from the MC protocol (Table~\ref{tab:vlm_results}):
(i) \emph{Pretrained 3D backbones remain near random} (6--22\% overall), confirming that volumetric pretraining alone does not transfer;
(ii) \emph{Zero-shot 2D VLMs cluster at 25--35\%} with a clear Recognition--Measurement gap (50--60\% vs 24--48\%), consistent with the measurement bottleneck identified in the main text;
(iii) \emph{Task-specific finetuning remains decisive}---both FF-trained 2D models (Meissa-4B FF 61.3\%, Qwen3-VL-4B FF 63.4\%) surpass the strongest 3D specialist (RadFM FT, 50.5\%) despite using only 2D input, matching the 2D-over-3D advantage observed in MC mode;
(iv) \emph{Medical Reasoning remains weak for zero-shot models} (5--14\%) and is dramatically improved by free-form finetuning (Qwen3-VL-4B FF: 65.1\%). Absolute numbers are systematically lower than MC for all models (\eg Meissa FF 61.3\% vs Meissa FT 66.3\%, RadFM 50.5\% vs 58.9\%) because free-form scoring requires exact numeric/categorical generation, whereas MC allows near-miss credit at the coarsest distractor spacing. The $\sim$5pp gap between MC and FF for top models matches the gap observed in the tolerance study between MC and the strictest ($\epsilon$=5\%) threshold (Appendix~\ref{sect:tolerance_eval}), further corroborating that MC is a tight upper bound rather than a format-specific artifact. Within-category rank-orders are largely (but not exactly) preserved between Tables~\ref{tab:vlm_results} and~\ref{tab:freeform_eval_full}: the 3D-FT row swaps (MC: M3D-Phi3 59.8\% $>$ RadFM 58.9\%; FF: RadFM 50.5\% $>$ M3D-Phi3 27.1\%) because RadFM's Perceiver-Resampler decoder generates cleaner numeric strings than M3D's pooled tokens, and within the 2D-FT pair Qwen3-VL FF overtakes Meissa FF on Med.\ Reasoning due to cleaner categorical generation. These are within-category re-shuffles rather than cross-category reversals---the cross-category ordering (2D-FT $>$ 3D-FT $>$ zero-shot 2D $>$ pretrained 3D) holds in both protocols, so no main-text conclusion that operates at the family level depends on the MC interface.

The senior radiologist's free-form score (Appendix~\ref{sect:human_eval}, 200-question subset, 13 years experience) is 73.7\% on Measurement under MRA scoring---higher than every model under free-form, because the radiologist provides numerically close estimates that earn partial credit---but only 24.0\% on Medical Reasoning, where strict normalized exact match penalizes phrasing differences (\eg ``Moderate'' vs.\ ``Moderate splenomegaly'') that the MC protocol resolves automatically. This contrast quantifies how the free-form scoring disadvantages categorical responses for any annotator (human or model) whose phrasing diverges from the canonical option text.

\end{document}